\documentclass[referee,sn-vancouver,Numbered]{sn-jnl}

\renewenvironment{table}[1][]%
{\tableorg[#1]%
\tablebodyfont%
\renewcommand\footnotetext[2][]{{\removelastskip\vskip3pt%
\let\tablebodyfont\tablefootnotefont%
\hskip 0pt \if!##1!\else{\smash{$^{##1}$}}\fi##2\par}}%
}{\endtableorg}

\usepackage[utf8]{inputenc}
\usepackage{graphicx}
\usepackage[ruled]{algorithm2e}
\usepackage{verbatim}
\usepackage{amsmath,amssymb,amsfonts}
\usepackage{amsthm}
\usepackage{mathrsfs}
\usepackage[title]{appendix}
\usepackage{xcolor}
\usepackage{textcomp}
\usepackage{manyfoot}
\usepackage{booktabs}
\usepackage{listings}
\usepackage{rotating}
\usepackage{array}
\usepackage{multirow}
\usepackage{acro}
\usepackage{tabularx}
\usepackage{float}
\usepackage{lscape}
\usepackage{colortbl} % For cell colors
\usepackage{adjustbox}
\usepackage{tabularray}
\usepackage{longtable}
\usepackage{makecell}
\usepackage{svg}
\usepackage{hyperref}
\usepackage{hhline}
\usepackage{makecell}

\usepackage{ragged2e}

\makeatletter
\long\def\@tablecaption#1#2{%
  \setbox\tabcapbox\vbox{%
    \tablecaptionfont
    \justifying
    \setlength{\parindent}{0pt}
    {\bfseries #1}\hskip2mm #2\vphantom{y}\par
  }%
  \box\tabcapbox
}
\makeatother

\newcommand{\fref}[1]{\footnote{\href{#1}{\nolinkurl{#1}}}}

\title{CardioLM: A multilingual suite of small language models for the cardiology domain}

\author[1, 2]{\fnm{Manuela-Daniela} \sur{Danu}}\email{manuela.voinea@siemens-healthineers.com}
\author[3]{\fnm{Miguel} \sur{Rodríguez-Ortega}}\email{mrodrig8@bsc.es}
\author[3]{\fnm{Alberto} \sur{Becerra-Tome}}\email{abecerr1@bsc.es}
\author[3]{\fnm{Pablo} \sur{Arancibia}}\email{pablo.arancibia.bsc@gmail.com}
\author[3]{\fnm{Fernando} \sur{Gallego-Donoso}}\email{fernando.gallego@bsc.es}
\author[4]{\fnm{Lorenzo} \sur{Pratesi}}\email{lorenzo.pratesi@translated.com}
\author[5]{\fnm{Xi} \sur{Bai}}\email{xi.bai@ucl.ac.uk}
\author[1, 2]{\fnm{Andrei-Damian}  \sur{Dinescu}}\email{damian.dinescu@siemens-healthineers.com}
\author[3]{\fnm{Salvador} \sur{Lima-López}}\email{salvador.limalopez@bsc.es}
\author[6,7]{\fnm{Andrea} \sur{Guala}}\email{andrea.guala@vhir.org}
\author[8]{\fnm{Saskia} \sur{Haitjema}}\email{s.haitjema@umcutrecht.nl}
\author[]{\fnm{Jonas}  \sur{Mack}}\email{jns.mck@gmail.com}
\author[9]{\fnm{Michele}  \sur{Miranda}}\email{miranda@di.uniroma1.it}
\author[]{\fnm{Wesam} \sur{Al Nabki}}\email{wesam.alnabkial@bsc.es}
\author[]{\fnm{Jan} \sur{Rodríguez-Miret}}\email{jan.rodriguez@bsc.es}
\author[1]{\fnm{Vasile George} \sur{Marica}}\email{marica.vasile.george@gmail.com}

\author[10]{\fnm{Vladimír} \sur{Kincl}}\email{vladimir.kincl@fnusa.cz}
\author[10]{\fnm{Kateřina} \sur{Dostálová}}\email{katerina.dostalova@fnusa.cz}
\author[10, 11]{\fnm{Michael} \sur{Andrej}}\email{514853@muni.cz}
\author[12, 13]{\fnm{Alicia} \sur{Uijl}}\email{a.uijl@amsterdamumc.nl}
\author[8]{\fnm{Corinne} \sur{Allaart}}\email{C.G.Allaart@umcutrecht.nl}
\author[12, 13]{\fnm{Machteld} \sur{Boonstra}}\email{m.j.boonstra@amsterdamumc.nl}
\author[8]{\fnm{Malin} \sur{Overmars}}\email{L.M.Overmars-2@umcutrecht.nl}
\author[5]{\fnm{Anoop} \sur{Shah}}\email{a.shah@ucl.ac.uk}
\author[5]{\fnm{Giovanna} \sur{Ceroni}}\email{g.ceroni@ucl.ac.uk}
\author[5]{\fnm{Jonathan} \sur{Kossoff}}\email{j.kossoff@nhs.net}
\author[14]{\fnm{Domenico} \sur{D'Amario}}\email{domenico.damario@uniupo.it}
\author[14]{\fnm{Renzo} \sur{Laborante}}\email{dottorrenzolaborante@gmail.com}
\author[14]{\fnm{Antonio} \sur{Iaconelli}}\email{iaconelli.antonio88@gmail.com}
\author[14]{\fnm{Isabella} \sur{Fumarulo}}\email{isabella.fumarulo@guest.policlinicogemelli.it}
\author[15]{\fnm{Jacopo} \sur{Lenkowicz}}\email{jacopo.lenkowicz@policlinicogemelli.it}
\author[15]{\fnm{Livia} \sur{Lilli}}\email{livia.lilli@policlinicogemelli.it}
\author[16]{\fnm{Emma} \sur{Weiss}}\email{emmaweissmd@gmail.com}
\author[16]{\fnm{Alex} \sur{Scafa}}\email{alexscafa@yahoo.com}
\author[17]{\fnm{Felix} \sur{Lindberg}}\email{felix.lindberg@ki.se}
\author[17]{\fnm{Milou} \sur{Gamage}}\email{milou.gamage@outlook.com}
\author[3]{\fnm{Eulàlia} \sur{Farré-Maduell}}\email{eulalia.farre@bsc.es}
\author[4]{\fnm{Sébastien} \sur{Bratières}}\email{sebastien@translated.com}
\author[3]{\fnm{Martin} \sur{Krallinger}}\email{martin.krallinger@bsc.es}
\author[8]{\fnm{Bram} \sur{van Es}}\email{B.vanEs-3@umcutrecht.nl}

\affil[1]{\orgname{Siemens SRL}, \orgaddress{\city{Brasov}, \country{Romania}}}
\affil[2]{\orgdiv{Automation and Information Technology}, \orgname{Transilvania University of Brasov}, \orgaddress{\city{Brasov}, \country{Romania}}}
\affil[3]{\orgdiv{NLP4BIA}, \orgname{Barcelona Supercomputing Center}, \orgaddress{\city{Barcelona}, \country{Spain}}} % 4
\affil[4]{\orgname{Translated}, \orgaddress{\city{Rome}, \country{Italy}}} % 2
\affil[5]{\orgname{University College London}, \orgaddress{\city{London}, \country{United Kingdom}}}
\affil[6]{ \orgname{Vall d'Hebron Institute of Research}, \orgaddress{\city{Barcelona}, \country{Spain}}} % 5
\affil[7]{ \orgdiv{CIBER‑CV}, \orgname{Instituto de Salud Carlos III}, \orgaddress{\city{Madrid}, \country{Spain}}} % 6
\affil[8]{\orgname{University Medical Center Utrecht}, \orgaddress{\city{Utrecht}, \country{The Netherlands}}}
\affil[9]{\orgname{Sapienza University of Rome}, \orgaddress{\city{Rome}, \country{Italy}}} % 2

\affil[10]{\orgdiv{International Clinical Research Center, St. Anne 's University Hospital, Faculty of Medicine}, \orgname{Masaryk University}, \orgaddress{\city{Brno}, \country{Czech Republic}}}
\affil[11]{\orgdiv{Department of Internal Medicine/Cardioangiology}, \orgname{University Hospital Hradec}, \orgaddress{\city{Králové}, \country{Czech Republic}}}
\affil[12]{\orgdiv{Department of Cardiology, Amsterdam University Medical Centers}, \orgname{University of Amsterdam}, \orgaddress{\city{Amsterdam}, \country{the Netherlands}}}
\affil[13]{\orgname{Amsterdam Cardiovascular Sciences Research Institute}, \orgaddress{\city{Amsterdam}, \country{the Netherlands}}}
\affil[14]{\orgname{Fondazione Policlinico Universitario Agostino Gemelli IRCCS}, \orgaddress{\city{Rome}, \country{Italy}}}
\affil[15]{\orgdiv{Data Science Unit}, \orgname{Fondazione Policlinico Universitario Agostino Gemelli IRCCS}, \orgaddress{\city{Rome}, \country{Italy}}}
\affil[16]{\orgname{Bucharest Emergency Clinical Hospital}, \orgaddress{\city{Bucharest}, \country{Romania}}}
\affil[17]{\orgdiv{Department of Clinical Science and Education, Södersjukhuset}, \orgname{Karolinska Institutet}, \orgaddress{\city{Stockholm}, \country{Sweden}}}

\DeclareAcronym{nlp}{
  short=NLP,
  long=natural language processing,
}
\DeclareAcronym{ehr}{
  short=EHR,
  long=electronic health record,
}
\DeclareAcronym{ner}{
  short=NER,
  long=named entity recognition,
}
\DeclareAcronym{all}{
  short=ALL,
  long=approximate list lookup,
}
\DeclareAcronym{cnn}{
  short=CNN,
  long=convolutional neural network,
}
\DeclareAcronym{umcu}{
  short=UMCU,
  long=University Medical Center Utrecht,
}
\DeclareAcronym{bow}{
  short=BOW,
  long=bag-of-words,
}
\DeclareAcronym{icd}{
  short=ICD,
  long=International Classification of Disease,
}
\DeclareAcronym{crf}{
  short=CRF,
  long=conditional random fields,
}
\DeclareAcronym{rnn}{
  short=RNN,
  long=recurrent neural network,
}
\DeclareAcronym{svm}{
  short=SVM,
  long=support vector machine,
}
\DeclareAcronym{lstm}{
  short=LSTM,
  long=long short-term memory,
}
\DeclareAcronym{bilstm}{
  short=biLSTM,
  long=bilateral LSTM,
}
\DeclareAcronym{tf-idf}{
  short=TF-IDF,
  long=term frequency-inverse document frequency,
}
\DeclareAcronym{gru}{
  short=GRU,
  long=gated recurrent unit,
}
\DeclareAcronym{qrnn}{
  short=QRNN,
  long=quasi-recurrent neural network,
}
\DeclareAcronym{bigru}{
  short=biGRU,
  long=bidirectional GRU,
}

\begin{document}
\raggedbottom

\abstract{
\textbf{Background:} Cardiovascular care generates large volumes of clinical documentation, most of it recorded as free text in the language of the treating institution. Clinical natural language processing (NLP) resources remain concentrated in English, cardiology is rarely treated as a distinct target domain, and no existing benchmark is simultaneously cardiology-specific, multilingual, and broad in entity coverage. This limits cross-border reuse of cardiovascular data in Europe.
\textbf{Methods:} We present two contributions. The first is CardioCCC, a corpus of 508 cardiology case reports originally published in Spanish and extended to Czech, Dutch, English, Italian, Romanian, and Swedish through machine translation, annotation projection, and clinician review, yielding over 400,000 validated annotations of diseases, symptoms, procedures, and medications. The second is CardioLM, a suite of small language models covering the same seven languages. It comprises CardioBERTa, a set of language-specific encoders obtained by continued pretraining of existing backbones on cardiology and biomedical corpora, and CardioNER, a set of named entity recognition (NER) models fine\mbox{-}tuned on CardioCCC. CardioNER includes monolingual models built on the CardioBERTa encoders and on their corresponding base models, as well as multilingual XLM\mbox{-}RoBERTa\mbox{-}large models fine\mbox{-}tuned jointly on all seven languages. We measured the contribution of cardiology-specific pretraining both intrinsically, by pseudo-perplexity evaluation of the encoders, and extrinsically, by comparing the NER models fine-tuned from the CardioBERTa encoders and from their base models.
% We measured the contribution of cardiology-specific pretraining both intrinsically, by pseudo-perplexity evaluation of the encoders, and extrinsically, by comparing the paired fine-tuned NER models. 
We further compared the multilingual models with the monolingual ones to assess the effect of joint multilingual training. Finally, CardioNER was validated by clinicians on synthetic hospital-style records and anonymised discharge letters.
\textbf{Results:} Cardiology-specific pretraining substantially reduced pseudo-perplexity in most languages, by up to a factor of eight, yet changed strict span-level F1 by at most 0.04, with benefits concentrated in languages whose backbones lacked biomedical coverage. The jointly trained multilingual XLM\mbox{-}RoBERTa\mbox{-}large models achieved the best strict F1 in nearly every language and entity type, exceeding the strongest monolingual system by up to 0.13 F1. A large share of residual errors concerned boundary detection: relaxed matching recovered 0.10\mbox{--}0.23 F1 for disease, procedure, and symptom, but only 0.01\mbox{--}0.07 for medication. Clinicians rated CardioNER well above a rule- and dictionary-based baseline, with the mean rating across entity types rising from 26\% to 73\%, and judged its output useful or very useful throughout.
\textbf{Conclusions:} Specialty-specific pretraining is not the decisive factor for downstream entity recognition, whereas multilingual joint training yields the strongest results in nearly all settings and a simpler deployment path for multi\mbox{-}country clinical information extraction.}


\maketitle

\section{Introduction}
Cardiovascular diseases (CVDs) remain a leading cause of morbidity and mortality in Europe and place a substantial burden on healthcare systems \cite{timmis2026european}. As part of routine care, European clinical centers generate large volumes of cardiovascular data through diagnostic procedures, treatments, and patient monitoring. These data represent a highly valuable resource to realize evidence-based care and data-driven innovation in cardiology \cite{hemingway2018big}. Their reuse across institutions and countries, however, remains limited by substantial technical, legal and organisational reasons \cite{vukovic2022enablers}. Clinical records are stored in heterogeneous formats, follow institution-specific documentation practices, and vary in language, coding systems, and degree of structure \cite{neveol2018clinical, OECD2025CardioHealth}. This fragmentation hinders the integration and joint analysis of cardiovascular data across sites, thereby reducing their value for research and healthcare innovation \cite{OECD2025CardioHealth}.

Electronic health records (EHRs) have become one of the main sources of clinical information in modern healthcare systems, capturing rich and detailed accounts of patients' medical histories, diagnoses, symptoms, procedures, treatments, and clinical findings \cite{cowie2017electronic}. However, a large share of this information is not recorded in structured fields but embedded in unstructured free-text narratives such as discharge summaries, progress notes, referral letters, and imaging reports. These narratives are characterized by specialized terminology, institution-specific conventions, dense abbreviations, frequent misspellings, measurement-heavy passages, and extensive synonymy \cite{tayefi2021challenges}. Consequently, clinically relevant information remains difficult to access at scale without dedicated NLP methods, limiting its use in downstream applications such as cohort identification \cite{shivade2014review}, epidemiological analysis \cite{gianfrancesco2021narrative}, clinical decision support \cite{eguia2024clinical}, and multicenter cardiovascular research \cite{cunningham2024natural}. 

Named entity recognition (NER) is a fundamental step toward making this information accessible \cite{kundeti2016clinical}. NER automatically identifies and classifies mentions of clinically relevant concepts, thereby converting unstructured narratives into structured representations \cite{li2022neural}. The extracted entities can subsequently be linked to controlled medical terminologies and ontologies \cite{french2023overview}, including SNOMED CT \cite{snomed2026}, ICD-11 \cite{who2019icd11}, and the Unified Medical Language System (UMLS) \cite{bodenreider2004unified}, to support semantic interoperability and cross-site data reuse. Particularly at European level, however, effective clinical NER must address two challenges simultaneously: the highly specialized language of clinical documentation and the linguistic diversity of national healthcare systems \cite{neveol2018clinical}.

Transformer-based language models have become the dominant approach to clinical text processing \cite{vaswani2017attention}. Through large-scale self-supervised pretraining followed by efficient task-specific fine\mbox{-}tuning, these models learn reusable linguistic representations that can be adapted to downstream applications \cite{devlin2019bert}. Their success in general-domain NLP has motivated domain-adaptive pretraining on specialized corpora to better align model representations with the vocabulary and distribution of a target domain \cite{gururangan2020don}. In biomedical and clinical NLP, representative models such as BioBERT \cite{lee2020biobert}, \mbox{SciBERT} \cite{beltagy2019scibert}, PubMedBERT \cite{gu2021domain}, and ClinicalBERT \cite{alsentzer2019publicly} have expanded terminology coverage and outperformed general-domain baselines on tasks including NER, relation extraction (RE), and concept normalisation. 

Despite this progress, existing approaches remain limited in multilingual clinical settings \cite{shaitarova2023exploring}. Most biomedical and clinical language models are trained predominantly on English corpora, while general-domain multilingual models, such as multilingual BERT (mBERT) \cite{devlin2019bert} and XLM-RoBERTa \cite{conneau2020unsupervised}, provide broad linguistic coverage but lack explicit clinical specialisation. Both types of models therefore remain sensitive to the distributional shift between their pretraining data and target clinical text \cite{gururangan2020don}, whose vocabulary, style, and structure vary across languages, medical specialties, institutions, and documentation practices \cite{sohn2018clinical}. Clinical narratives may also combine the local language with Latin-derived or English medical expressions, while incorporating locally specific abbreviations, brand names, and regional jargon \cite{neveol2018clinical}. Such variation reduces model robustness and limits reliable generalisation across clinical settings \cite{laparra2020rethinking}. The problem is further exacerbated by the limited availability of sufficiently large annotated clinical corpora \cite{neveol2018clinical, shaitarova2023exploring}, in turn affected by annotation inconsistency across clinical datasets \cite{yada2020towards}. As a result, even languages well represented in general-domain NLP may remain under-represented in clinical NLP. These limitations are especially pronounced in clinically under-resourced European languages such as Czech, Dutch, Swedish, and Romanian. 

Previous research has attempted to bridge this gap by combining clinical domain adaptation with multilingual transfer strategies \cite{jorgensen2021mdapt}. Common approaches include continued pretraining on biomedical or clinical corpora, typically using masked language modelling or span-level objectives \cite{gururangan2020don, jorgensen2021mdapt}, zero-shot transfer from a pivot language \cite{gaschi2023exploring}, machine-translation pipelines \cite{rodriguez2024exploring, seinen2024annotation}, language-specific adaptation through dedicated adapters or task heads \cite{pfeiffer2020mad}, and ontology-guided pretraining that injects knowledge from resources such as UMLS \cite{bodenreider2004unified} and SNOMED CT \cite{liu2021learning}. While these strategies typically improve either domain alignment or cross-lingual transfer, few address both systematically \cite{jorgensen2021mdapt, geng2023kbioxlm}. 

These limitations carry particular weight in cardiology, which is rarely treated as a distinct target domain, and for which cross-lingual generalisation is seldom evaluated systematically beyond high-resource settings \cite{lima2024overview, reading2022systematic}. Within this landscape, downstream cardiology tasks, such as NER, are hampered by scarce and heterogeneous datasets and annotation inconsistency \cite{lima2024overview, reading2022systematic}, ambiguous entity boundaries and nested mentions \cite{becker2025extending}, abbreviation expansion \cite{tang2026end}, and variation in the mapping of textual mentions to controlled terminologies \cite{dimitriadis2025overview}. This underscores the need for multilingual, cardiology-aware models that combine specialty knowledge with robust multilingual representations, evaluated under a standardized framework across languages and clinical tasks \cite{lima2024overview}.

The strategy we investigate is domain-adaptive pretraining: the self-supervised training of a general-domain pretrained encoder on in-domain text, realigning its lexical and contextual representations with cardiology-specific terminology without requiring labeled data \cite{gururangan2020don}. We apply it separately to each language, and the adapted encoders are then fine-tuned for token classification to identify diseases, symptoms, procedures, and medications. This raises our first research question: does continued pretraining on biomedical and cardiology-related corpora improve multilingual cardiology NER? Because the resulting encoders are language-specific, a second question follows: how do they compare with a single general-domain multilingual encoder fine\mbox{-}tuned jointly on all languages, with no cardiology-specific pretraining stage --- an approach that would considerably simplify deployment across
countries? 

We focus on encoder-based BERT architectures \cite{devlin2019bert}, typically containing 100–400 million parameters, rather than on multi-billion-parameter generative large language models (LLMs) that currently dominate the broader NLP landscape, prioritizing reproducible, fine-tunable encoders that can be deployed with relatively modest computational requirements. Their fixed tokenisation and stable fine-tuning behavior facilitate controlled comparisons of domain adaptation and multilingual transfer. In this paper, we therefore assess multiple BERT-based backbones within a common domain-adaptation, fine-tuning, and evaluation pipeline for multilingual cardiology NER.

This work was conducted within the DataTools4Heart (DT4H) European Union (EU) project (Grant Agreement 101057849), targeting seven European languages: Czech, Dutch, English, Italian, Romanian, Spanish, and Swedish. We make two main contributions. The first is CardioCCC, a corpus of 508 cardiology case reports originally published in Spanish and released for the MultiCardioNER shared task \cite{lima2024overview} with disease annotations in Spanish and medication annotations in Spanish, English, and Italian. We extend it to seven languages and four entity types --- diseases, symptoms, procedures, and medications --- through machine translation, annotation projection, and clinician review, yielding over 400,000 validated annotations~\cite{farre_maduell_2026_23060841}. To the best of our knowledge, this makes CardioCCC the first cardiology corpus that is simultaneously multilingual and broad in entity coverage. The second contribution is CardioLM, a suite of small language models covering the same seven languages, comprising \mbox{CardioBERTa}\fref{https://huggingface.co/collections/DT4H/cardioberta-family}, a set of language-specific encoders obtained through continued pretraining on biomedical and cardiology-related corpora, and CardioNER\fref{https://huggingface.co/collections/DT4H/cardioner}, a set of monolingual and multilingual models fine-tuned for cardiology NER on \mbox{CardioCCC}. The CardioCCC corpus, the domain-adapted encoders, and the fine-tuned NER models are all openly released.

This work extends the systems introduced in our BioASQ MultiCardioNER shared\mbox{-}task paper \cite{danu2025multilingual} in four main directions: (1) increasing language coverage from three to seven languages; (2) expanding the entity inventory from two to four clinical categories; (3) introducing a cardiology-specific domain-adaptation stage preceding NER fine-tuning; and (4) improving the NER training and evaluation pipeline.

We evaluated the resulting models along three complementary axes: intrinsically, by measuring how cardiology-specific pretraining affects domain fit, quantified through pseudo-perplexity; extrinsically, by fine-tuning CardioNER from both the adapted encoders and their original backbones, so that the effect of domain adaptation is isolated under an otherwise identical pipeline; and clinically, through comparison with a rule- and dictionary-based baseline on synthetic hospital-style records and anonymised discharge letters evaluated by practising clinicians.

A central finding is that intrinsic and downstream improvements diverge. Although cardiology-specific pretraining substantially improves domain fit in six of the seven languages, these gains translate into little or no improvement in strict span-level recognition, which changes by at most 0.04 F1 and mainly for languages whose original backbones lacked prior biomedical exposure. By contrast, joint multilingual training, which involves no cardiology-specific pretraining at all, yields consistent gains across languages and entity types. These results suggest that specialty-specific pretraining is less consequential for NER than multilingual knowledge sharing.
 
\section{Related work}

Clinical NLP has been developed predominantly around English resources, leaving information in non-English clinical records comparatively difficult to access~\cite{shaitarova2023exploring}.
The European Health Data Space (EHDS) Regulation~\cite{EHDS2025}, which entered into force in 2025, establishes a common framework for the cross-border exchange and secondary use of EHR data across EU member states, strengthening the need for NLP pipelines capable of processing clinical documentation across European languages and healthcare systems. This section surveys the technical foundations relevant to meeting this need, including Transformer-based language models and domain adaptation for biomedical and clinical NLP, multilingual and cross-lingual clinical NLP, and named entity recognition as a core clinical information extraction task. We review each in turn before synthesising the specific gaps addressed by our paper.

\subsection{Transformer-based language models and domain adaptation for biomedical and clinical NLP}
\label{transformers}

Transformer-based encoders pretrained using masked language modelling have become foundational to biomedical and clinical NLP. Early research focused primarily on adapting general-domain models to biomedical literature:  BioBERT~\cite{lee2020biobert} continued the pretraining of BERT on PubMed abstracts and full-text articles, improving performance across biomedical NER, relation extraction, and question answering. Clinical documentation, however, poses challenges less prominent in biomedical literature, including dense abbreviations, non-standard spelling, frequent negation, typos, and writing conventions that vary across clinical specialties and institutions. Models trained primarily on scientific publications may, therefore, not transfer effectively to clinical narratives as they may not capture the terminology, structure, and style of routine clinical practice. ClinicalBERT~\cite{alsentzer2019publicly}, pretrained on de-identified discharge summaries from the MIMIC-III \cite{PhysioNet-mimiciii-1.4}, demonstrated the value of modelling clinical language directly and established BERT-based clinical encoders as an important foundation for downstream clinical NLP tasks.

What these models share is domain adaptation, whose scope and limits have been studied more systematically in the general NLP literature. General-purpose pretrained language models can be adapted to a specific domain through two principal strategies: continued pretraining (CPT) on in-domain data, or pretraining from scratch on a domain-specific corpus, both via a self-supervised learning objective such as masked language modelling or next-sentence-prediction. The choice between them depends on the availability of in-domain text, the computational resources at hand, and the degree of divergence between the target domain and the general-domain pretraining distribution.

A foundational systematic study of CPT is that of Gururangan et al.~\cite{gururangan2020don}, who introduced domain-adaptive pretraining (DAPT) and showed that a second phase of pretraining on unlabelled in-domain data yields consistent gains across a range of tasks, under both high- and low-resource settings, with benefits greatest when the target domain diverges substantially from the pretraining distribution. A well\mbox{-}documented limit of CPT, however, is catastrophic forgetting: as the model is updated on narrow in-domain data, it can lose general linguistic capabilities that matter for out-of-domain generalisation and for handling the diversity of clinical text. Proposed mitigations include model merging, which interpolates the weights of the original and CPT\mbox{-}adapted models as a post-hoc form of regularisation~\cite{Alexandrov2024}, and domain data mixing, which interleaves general or broad-domain corpora with the target data so that earlier representations are preserved through experience replay~\cite{ibrahim2024simple}.

Pretraining from scratch instead learns both the model parameters and vocabulary directly from domain-specific text. PubMedBERT~\cite{gu2021domain} applied this approach to English biomedical text, showing that it can outperform CPT from a general-domain initialisation when sufficient in-domain data is available. This gain, however, comes at the cost of
large corpora and substantial compute, whereas CPT reuses an existing model's parameters and vocabulary at far lower cost.

Beyond downstream task-performance improvements, domain adaptation also aligns with the data-governance constraints of clinical NLP. Because it trains a model directly on in-domain text, whether by continued pretraining or from scratch, it can be carried out entirely within an institution's own computing environment. Private records therefore never leave the
institution, making in-house adaptation a structurally GDPR-compatible route to domain-knowledge transfer that satisfies data-minimisation
requirements~\cite{griot2025implementation, Verkijk2022}.

The encoders discussed so far represent one branch of the transformer family; the other, generative decoder-based large language models (LLMs), has recently achieved strong results on a range of medical language-understanding tasks, including structured information extraction~\cite{Singhal2022,Agrawal2022,Hu2024ner}. However, their suitability depends strongly on the operational setting and output requirements. For large-scale retrospective annotation of clinical records, Transformer-based token-classification models offer several practical advantages over LLMs. First, autoregressive models generate output sequentially, one token at a time, whereas encoder-based models predict token-level labels in parallel within a single forward pass. Although batching, quantisation, and hardware acceleration can reduce costs, sequential decoding remains more computationally expensive at scale, especially for long outputs or repeated prompting. Second, clinical information-extraction pipelines often require entities to be returned as verbatim source spans with stable character offsets to support reliable linking to source documents. Generative architectures are less directly aligned with this extractive requirement because they may paraphrase or normalise mentions, while prompt sensitivity and post-processing can introduce omissions, misalignment, and inconsistent boundaries. Encoder-based token classification better supports exact-span extraction by assigning labels directly to input tokens, though subword tokenisation and subsequent character-level span reconstruction still require careful handling. Third, deploying generative models in clinical settings is constrained by strict data-protection regulations and demanding security and infrastructure requirements. Because European hospitals are often required to process patient data within their own institutional computing environments, running billion-parameter models on premises can require specialised GPU infrastructure, substantial memory, and considerable engineering effort to ensure secure, reliable, and scalable inference. Encoder-based transformers, by contrast, have lower computational and operational demands and can be integrated more readily into established clinical NLP pipelines running on standard hardware~\cite{Hu2025clinical}. Lastly, we shortly mention diffusion-based language models~\cite{Nie2026}, an alternative non-causal generative model that allows for the parallel generation of tokens, which is more performant for inference compared to causal models but is less amenable for classification tasks, and is still, comparatively speaking, lacking support from the open source community.

These considerations do not preclude the use of generative models for clinical information extraction; rather, they justify our choice of encoder-based models, for which on-premises deployment, computational efficiency, reproducibility, and exact span recovery are central requirements.

\subsection{Multilingual and cross-lingual clinical NLP}
\label{subsec:multilingual}

Clinical documentation in European healthcare systems is produced in many languages and shaped by national, institutional, and specialty-specific conventions. Despite this diversity, biomedical and clinical NLP resources are still primarily focused on the English language, leaving many European languages with limited pretrained models, annotated corpora, and evaluation benchmarks. Existing work has therefore pursued two complementary directions: the development of language- and domain-specific models, and the use of multilingual and cross-lingual methods to transfer knowledge across languages.

Monolingual encoders provide an important basis for language-specific NLP. Examples include RobeCzech~\cite{straka2021robeczech} for Czech, RobBERT~\cite{Delobelle2020} and RobBERT-2023~\cite{Delobelle2024} for Dutch, DrBERT for French~\cite{Labrak2023}, AlBERTo for Italian~\cite{Polignano2019}, Romanian BERT~\cite{Dumitrescu2020}, BETO~\cite{canete2023spanish} for Spanish, and KB\mbox{-}BERT for Swedish~\cite{malmsten2020playing}. By dedicating their vocabularies and model capacity to a single language, such models can provide stronger language-specific representations than broadly multilingual encoders. However, they are generally pretrained on Wikipedia, news corpora, social media, or general web crawl data, and therefore remain poorly matched to the terminology, abbreviations, and discourse conventions of clinical documentation.

Domain-specific pretraining reduces this mismatch. For Spanish, Carrino et al.~\cite{Carrino2022} introduced language models trained from scratch on separate biomedical and clinical corpora. For Italian, BioBIT and MedBIT~\cite{buonocore2023localizing} were developed using resources including machine-translated PubMed abstracts, Italian medical textbooks, and health-related web content. Although such biomedical pretraining improves the representation of medical terminology, scientific publications and curated medical resources generally consist of grammatical, carefully edited prose and therefore differ from routine clinical notes, which are often telegraphic, abbreviation-heavy, and shaped by local practices. Even common clinical concepts may have different language-specific abbreviations: left ventricular ejection fraction, for example, is commonly abbreviated as ``FEVI'' in Spanish, ``FEVS'' in Italian and Romanian, ``EFLK'' in Czech, and ``LVEF'' in Dutch and English documentation. Clinical-domain pretraining addresses this register difference more directly. MedRoBERTa.nl~\cite{Verkijk2022}, developed through continued pretraining on Dutch electronic health records, has been applied to lifestyle classification~\cite{Muizelaar2024} and negation detection~\cite{VanEs2023}. Similarly, Lamproudis et al.~\cite{lamproudis2021developing} adapted the Swedish KB-BERT model through continued pretraining on a large corpus of clinical text. Within cardiology, CardioBERTpt~\cite{Schneider2023} was developed by adapting a Portuguese encoder to specialty-specific clinical text and demonstrated improved clinical NER performance compared with more general biomedical models. Together, these studies suggest a continuum of increasing specialisation, from general-language and biomedical models to clinical and specialty-specific encoders. Nevertheless, the availability of such resources remains uneven, with comparable biomedical or clinical models still limited for several European languages, including Czech and Romanian.

Multilingual models offer a more scalable alternative to developing separate language-specific models, particularly in low-resource settings. Multilingual BERT (mBERT)~\cite{devlin2019bert}, pretrained on Wikipedia in 104 languages, demonstrated strong zero\mbox{-}shot transfer across a range of languages and NLP tasks \cite{wu2019beto}. XLM-RoBERTa~\cite{conneau2020unsupervised} followed the same multilingual paradigm but used a RoBERTa-style architecture and substantially larger-scale pretraining on CommonCrawl data from 100 languages, outperforming mBERT on several cross-lingual benchmarks. However, both models are predominantly trained on general-domain text, and their capacity is shared across many languages. Their ability to represent language-specific medical terminology and clinical writing conventions may therefore be weaker than that of specialised monolingual or domain-adapted models.

Cross-lingual transfer seeks to exploit multilingual representations when target\mbox{-}language annotations are limited. In zero-shot transfer, a multilingual model is fine-tuned on a high-resource language and applied directly to another language~\cite{WuDredze2019}. Performance may nevertheless be affected by differences in terminology, syntax, abbreviation practices, and documentation style. Machine translation and annotation projection can instead transfer labelled data across languages~\cite{gaschi2023exploring}, although their reliability depends on translation quality and the accurate alignment of task-relevant spans and labels. Parameter-efficient methods, including adapter modules~\cite{Pfeiffer2020} and low-rank adaptation techniques~\cite{Remy2024_2}, adapt multilingual models by training a small set of additional parameters while keeping most of the pretrained model fixed, thereby reducing computational and storage costs when target-language data are limited. 

Overall, language-specific pretraining, clinical domain adaptation, multilingual modelling, and cross-lingual transfer provide complementary approaches to multilingual clinical NLP. Monolingual and domain-adapted models improve language and clinical specificity, while multilingual methods extend coverage to lower-resource languages. Progress nevertheless remains constrained by the limited availability of representative clinical corpora, language-specific clinical encoders, and consistently annotated evaluation resources across European languages.

\subsection{Clinical named entity recognition}
Named entity recognition (NER) is a foundational task in clinical NLP, enabling medically relevant concepts to be extracted from unstructured clinical narratives and represented in a structured form. Robust NER systems are essential for achieving reliable performance in downstream applications, including entity linking, clinical coding, pharmacovigilance, cohort identification, and decision support~\cite{navarro2023}. Early approaches relied primarily on rule-based systems and conventional sequence-labelling methods, while more recent work has been dominated by Transformer-based architectures, which generally achieve stronger performance on clinical concept-extraction benchmarks~\cite{Yang2020_3}.  

Progress in clinical NER has been driven largely by shared tasks and community benchmarks. However, few initiatives have simultaneously provided multilingual coverage, a broad entity inventory, and an explicit focus on cardiology. The CLEF eHealth 2021 SpRadIE task, for instance, addressed the recognition of seven entity types and hedge cues in Spanish radiology reports~\cite{suominen2021overview}. Other Spanish-language shared tasks targeted general biomedical and clinical semantic categories, including drugs and chemical compounds in PharmaCoNER~\cite{gonzalez2019pharmaconer}, diseases in DisTEMIST~\cite{miranda2022distemist}, symptoms, signs, and clinical findings in SympTEMIST~\cite{lima2023symptemist}, and clinical procedures in MedProcNER~\cite{lima2023medprocner}. Although some of the concepts covered by these tasks are relevant to cardiology, they were not cardiology-specific, and their annotations were not restricted to cardiovascular conditions or care. Furthermore, all of the aforementioned NER initiatives were monolingual and based on Spanish-language texts, while the latter four tasks each focused on a single broad semantic category. 

Multilingual clinical information extraction was explored in the CLEF eHealth 2018 ICD-10 coding challenge, which used death certificates in French, Hungarian, and Italian~\cite{suominen2018overview}. However, this task focused on clinical coding rather than span-level NER. Taken together, none of these initiatives provided a multilingual cardiology benchmark covering multiple entity types across the clinical care pathway. 
This gap was partially addressed by the MultiCardioNER shared task~\cite{lima2024overview}, organised within the BioASQ Lab at CLEF~2024. The task released the CardioCCC corpus, a collection of cardiology clinical case reports with disease annotations provided in Spanish and medication annotations provided in Spanish, English, and Italian. Accordingly, the benchmark comprised a Spanish disease-recognition task and a multilingual medication-recognition task covering all three languages. Several systems submitted to MultiCardioNER~\cite{Danu2024, Aksenova2024} demonstrated the suitability and effectiveness of BERT-based token classification approaches for cardiology NER.

Despite its contribution, MultiCardioNER remains limited in both semantic and linguistic scope. First, an entity inventory restricted to diseases and medications captures only part of the information required for detailed cardiovascular phenotyping. A comprehensive representation of the clinical care pathway should encompass patient presentation (symptoms), diagnosis (diseases), intervention (procedures), and clinical management (medications). Second, the benchmark covers three European languages that are comparatively better resourced than many languages used in clinical practice, leaving numerous clinically under-resourced languages without annotated cardiology corpora or suitable evaluation benchmarks. These limitations motivate the development of multilingual cardiology NER resources that combine broader semantic coverage with the inclusion of under-resourced clinical languages.

\subsection{Research gaps}
\label{subsec:gaps}
Two interrelated barriers currently limit multilingual cardiology NLP across European languages. First, cardiology-specific pretraining remains almost entirely unexplored: to the best of our knowledge, CardioBERTpt~\cite{Schneider2023} for Portuguese is the only published encoder pretrained on cardiology text, and no such model exists for any of the seven languages we target. Where cardiology NER has been addressed at all, it has relied on general-domain or broad clinical encoders fine-tuned on cardiology data, rather than on encoders pretrained on cardiovascular text. Second, no standardised benchmark is at once cardiology-specific, multilingual, and broad in its entity inventory. Existing cardiology resources are limited in linguistic scope and, as in MultiCardioNER, provide multilingual annotation for only a single entity type, while the broader entity coverage available for some languages comes from separate, single-category tasks that are not specific to cardiology. This precludes systematic measurement of cross-lingual performance across concepts relevant to cardiovascular phenotyping.

This paper addresses both barriers: CardioBERTa closes the encoder gap through cardiology-specific continued pretraining in each target language, while the seven\mbox{-}language CardioCCC extension closes the benchmark gap by broadening coverage along both the linguistic and the entity dimensions identified here, with the accompanying CardioNER models providing baselines across all seven languages.

\section{Methodology}

\begin{figure}[b]
  \centering
  \includegraphics[width=0.99\linewidth]{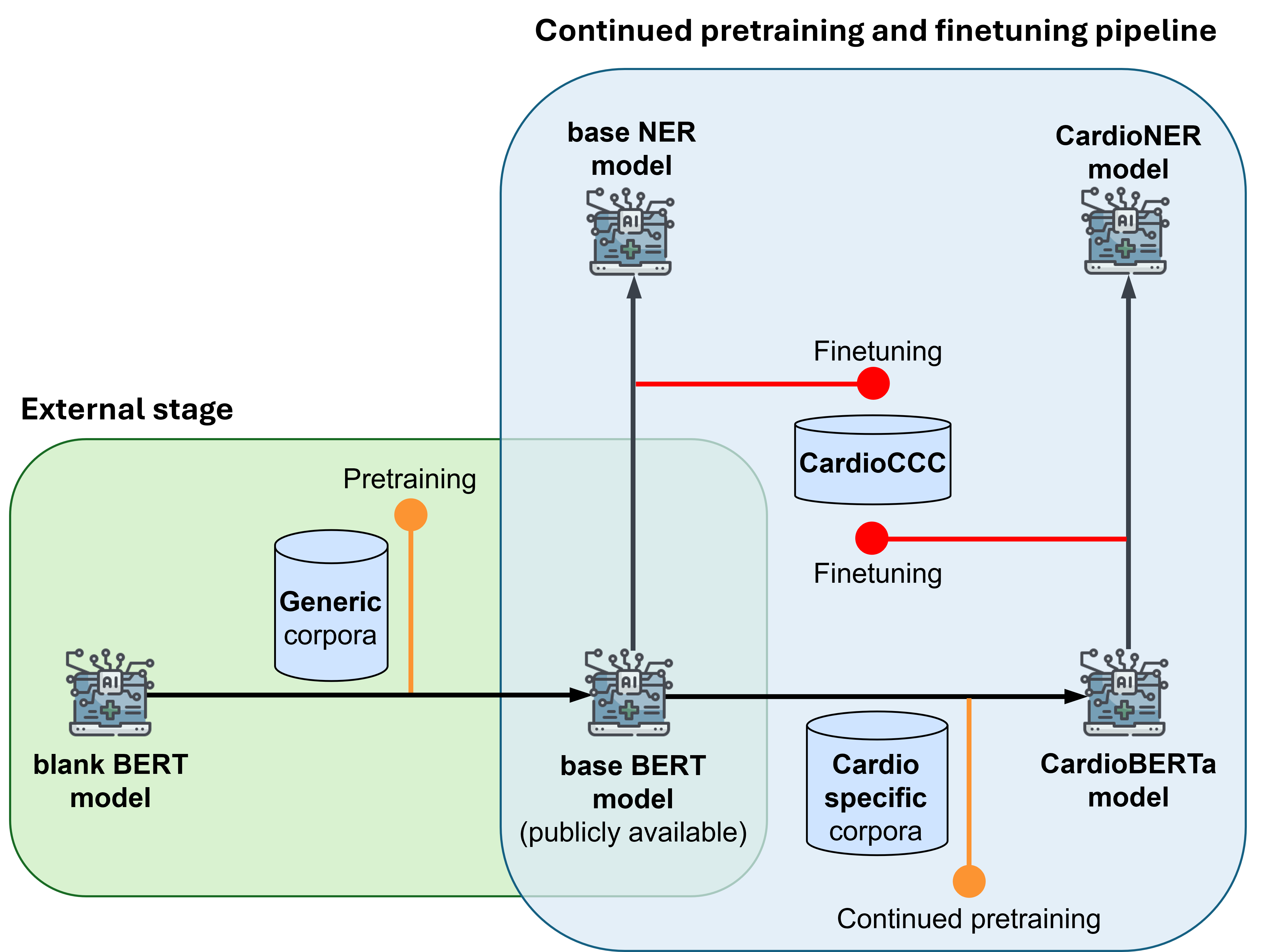}
  \caption{End-to-end pipeline from cardiology-domain continued pretraining to NER system development.}
  \label{fig:ner_pretraining_schema}
\end{figure}

Our approach follows an end-to-end pipeline that transforms unstructured multilingual cardiology text into structured clinical concepts. Starting from publicly available general or biomedical-domain transformer encoders, we (i) adapt them to the cardiology domain through continued pretraining on multilingual cardiology corpora and (ii) fine-tune the adapted encoders for span-level clinical entity recognition to identify four types of clinical entities: diseases, medications, procedures, and symptoms (Figure~\ref{fig:ner_pretraining_schema}). These two stages produce the two components of CardioLM — \mbox{CardioBERTa} and CardioNER — which are developed for seven European languages. To isolate the contribution of domain adaptation, we additionally fine-tuned the original, publicly available checkpoints directly on the annotated NER corpus, yielding baseline NER models trained without cardiology-specific pretraining. As a third configuration, we fine-tuned a single multilingual XLM-RoBERTa-large model
jointly on all seven languages, without any cardiology-specific pretraining stage, as a comparator to the language-specific models.

The remainder of this section details the data used in this study, followed by the proposed model development pipeline and the procedures used to evaluate it. It first describes the corpora assembled for continued pretraining, the ParaCliTe corpus of artificial cardiology documents, the CardioCCC annotated corpus, and the document set used for clinical validation. It then presents the selection of the base language models, the domain-adaptive continued pretraining procedure, the subsequent development and fine-tuning of the NER systems, and the NER inference setup used to generate entity predictions on unseen clinical text. Finally, it introduces the evaluation metrics: pseudo-perplexity for the language models, and span-level precision, recall, and F1 under strict and relaxed matching for NER.

\subsection{Data}

\subsubsection{Continued pretraining data}
\label{dapt-data}

A central consideration in constructing the continued pretraining corpora was the limited and uneven availability of cardiology-specific text across languages. Although relatively large collections of cardiovascular literature and clinical material are available for some languages, such resources are substantially scarcer for Czech, Romanian, and Swedish. Moreover, cardiology content is often embedded within broader biomedical or clinical repositories rather than provided as a standalone corpus. Relying exclusively on cardiology-specific data would therefore have resulted in large differences in corpus size and domain coverage across languages and, for the three languages above in particular, insufficient material for robust domain adaptation. To mitigate this imbalance while preserving cardiology as the primary target domain, the corpora were assembled independently for each language and constructed hierarchically: cardiology-specific material was prioritised, broader biomedical and clinical text was added to increase corpus scale and medical coverage, and machine-translated or general-domain material was incorporated where additional data were required. The resulting collections originated from several complementary source types, including anonymised clinical records and patient histories, curated biomedical and clinical datasets, academic literature, and machine-translated biomedical content, and were organised into three domain
categories: \emph{cardiology-specific}, \emph{biomedical/clinical}, and \emph{general-domain} text. 

Machine-translated resources played a particularly important role in compensating for the scarcity of native clinical and cardiovascular text, especially in Czech, Romanian, and Swedish, while also contributing to the other non-English corpora. Approximately two billion words of biomedical text per non-English language were derived from English PubMed abstracts translated using ModernMT \cite{germann2016modern}. This increased exposure to biomedical and cardiovascular terminology while reducing the disparity in available training data across languages.

Because cardiology-specific text rarely exists as a standalone dataset, relevant documents were frequently selected from larger repositories using domain-oriented filtering. For journal collections such as SciELO, cardiology-related articles were identified using regular expressions and keyword matching over journal names, descriptions, and publisher metadata, with a small filtering set of cardiology-related terms --- \emph{cardio}, \emph{cardiology}, \emph{heart}, \emph{coronary}, and \emph{vascular}. A similar terminology-based strategy was applied to broader clinical-note collections, including MIMIC, to identify notes containing cardiovascular content. For PubMed, cardiovascular documents were selected using MeSH descriptors associated with cardiovascular diseases together with their narrower terms. These procedures made it possible to construct cardiology-enriched subsets from resources that were not originally organised by clinical specialty.

All collected text underwent a common preprocessing pipeline before being incorporated into continued pretraining. First, character and encoding normalisation was applied to remove malformed or inconsistent symbols, particularly those introduced during PDF extraction. Duplicate documents were then removed using string-based hashing. For collections in which extraction artefacts could prevent duplicate documents from producing identical hashes, an additional fuzzy deduplication step based on locality-sensitive hashing was applied. 

Table~\ref{tab:cpt_data} reports the resulting size of the continued pretraining corpus for each language and domain, expressed in millions of words. Across the seven languages, the collections contain approximately 24.9 billion words, of which 3.12 billion words (12.5\%) are cardiology-specific, 21.22 billion (85.2\%) belong to the biomedical\mbox{/}clinical domain, and 0.57 billion (2.3\%) correspond to general-domain text. The relative contribution of these categories varies considerably between languages, reflecting the availability of suitable native-language resources as well as the volume of machine-translated content used to supplement them.

\begin{table}[t]
\centering
\caption{Size of the continued pretraining corpus per language, expressed in millions of words.}\label{tab:cpt_data}
\begin{tabular}{lcccc}
\toprule
Language & Cardiology & Biomedical & General & Total \\
\midrule
Czech    &    150 & 2{,}330 & 403 &  2{,}883 \\
Dutch    & 1{,}400 & 4{,}073 &   0 &  5{,}473 \\
English  &    771 &       8 & 163 &     942 \\
Italian  &    266 & 4{,}562 &   0 &  4{,}828 \\
Romanian &    184 & 4{,}211 &   0 &  4{,}395 \\
Spanish  &    207 & 3{,}578 &   0 &  3{,}785 \\
Swedish  &    143 & 2{,}460 &   0 &  2{,}603 \\
\midrule
 Total   &  3{,}121 &  21{,}222 &  566 &  24{,}909 \\
\bottomrule
\end{tabular}
\end{table}

A detailed overview of the resources used to construct these corpora is provided in Appendix~\ref{appendix_cpt} (Table~\ref{tab:corpus-overview}). This table catalogues the collected source corpora, whereas Table~\ref{tab:cpt_data} reports the amount of text ultimately used for domain adaptation after translation, where applicable, filtering, and preprocessing. The two tables are therefore not expected to yield identical totals.

\subsubsection{ParaCliTe corpus}
Assessing how well domain-specific language models capture clinical language requires evaluation text that is both clinically realistic and free of the access restrictions that normally encumber real patient records. To this end, we rely on ParaCliTe~\cite{pratesi2026multilingualartificialdatasetsaidriven}, a parallel corpus of artificial cardiology documents developed within the Horizon Europe project DataTools4Heart, which serves as the evaluation corpus for the pseudo\mbox{-}perplexity analysis described in Section~\ref{subsec:ppl}. ParaCliTe comprises 94 fictional adult cardiology cases, each authored by a clinician in one of seven European languages (Czech, Dutch, English, Italian, Romanian, Spanish, and Swedish) and subsequently rendered into the remaining six through a rigorous multi-stage professional translation and validation workflow, for a total of 658 documents. Because the cases describe fictional patients and contain no personal data, the corpus is openly released, making our evaluation fully reproducible.

\begin{table}[t]
\centering
\caption{Characteristics of the ParaCliTe dataset.}
\label{tab:paraclite_stats}
\resizebox{\textwidth}{!}{
\begin{tabular}{lrrrr}
\toprule
\multicolumn{5}{l}{\textit{Dataset characteristics}} \\
\quad Total word count (source; translations; total) & \multicolumn{4}{r}{51,079; 343,368; 394,447} \\
\quad Total number of documents & \multicolumn{4}{r}{658} \\
\addlinespace
\multicolumn{5}{l}{\textit{Demographics}} \\
\quad Age (average; range) & \multicolumn{4}{r}{58.2; 20--110} \\
\quad Female sex (n; \%) & \multicolumn{4}{r}{343 (52\%)} \\
\addlinespace
\multicolumn{5}{l}{\textit{Document types}} \\
\quad Outpatient report (n; \%) & \multicolumn{4}{r}{196 (30\%)} \\
\quad Inpatient report (n; \%) & \multicolumn{4}{r}{140 (21\%)} \\
\quad Discharge report (n; \%) & \multicolumn{4}{r}{322 (49\%)} \\
\quad Referral (n; \%) & \multicolumn{4}{r}{0 (0\%)} \\
\addlinespace
\multicolumn{5}{l}{\textit{Language distribution}} \\
 & \textbf{Source docs} & \textbf{Source words} & \textbf{Total words} & \textbf{\% of total} \\
\quad English  & 17 & 7,567  & 56,098 & 14.2\% \\
\quad Spanish  & 8  & 7,177  & 59,928 & 15.2\% \\
\quad Italian  & 11 & 7,433  & 59,933 & 15.2\% \\
\quad Dutch    & 19 & 7,347  & 54,241 & 13.8\% \\
\quad Romanian & 8  & 7,085  & 59,959 & 15.2\% \\
\quad Czech    & 8  & 7,362  & 53,707 & 13.6\% \\
\quad Swedish  & 23 & 7,108  & 50,581 & 12.8\% \\
\addlinespace
\multicolumn{5}{l}{\textit{Heart diseases or conditions}} \\
\quad Heart failure (n; \%) & \multicolumn{4}{r}{399 (61\%)} \\
\quad Ischemic heart disease (n; \%) & \multicolumn{4}{r}{287 (44\%)} \\
\quad Valve diseases (n; \%) & \multicolumn{4}{r}{238 (36\%)} \\
\quad Arrhythmias (n; \%) & \multicolumn{4}{r}{168 (26\%)} \\
\quad Atrial fibrillation (n; \%) & \multicolumn{4}{r}{154 (23\%)} \\
\quad Cardiac arrest (n; \%) & \multicolumn{4}{r}{21 (3\%)} \\
\bottomrule
\end{tabular}}
\end{table}
 
Several properties make ParaCliTe particularly well suited to our study. First, it is a genuinely parallel resource, released in two aligned views — a segment (sentence) level view and a document level view — covering all seven languages: the same clinical content is available in each language with a comparable information volume (approximately 7,100–7,600 words per language — mean 7,297 — totalling 51,079 words). Second, and central to our analysis, each case is tagged with the language in which it was originally authored prior to translation. This lets us contrast text natively written in a given target language against text translated into that language from a different source, and thereby test whether translated clinical text is systematically harder for a model to predict. The corpus is demographically balanced (49 documents describe female patients and 45 male patients, with a mean patient age of 58 ± 18 years) and spans multiple encounter types and cardiac conditions, reducing the risk that perplexity estimates are driven by a narrow slice of clinical language. The distribution of the 94 cases across the seven source languages is reported in Table~\ref{tab:paraclite_stats}.

\subsubsection{CardioCCC corpus}

The Cardiology Clinical Case Corpus (CardioCCC)~\cite{farre_maduell_2026_23060841} is a parallel corpus of 508 cardiology clinical case reports annotated across four semantic categories --- diseases, symptoms and findings, clinical procedures, and medications --- in Czech, Dutch, English, Italian, Romanian, Spanish, and Swedish. The texts were originally published in Spanish in open-access cardiology journals and were selected for their stylistic similarity to real EHRs. The six other language versions were produced through machine translation, annotation projection, and clinician review. Part of the corpus was previously released for the MultiCardioNER shared task at BioASQ 2024~\cite{lima2024overview}, comprising disease annotations in Spanish and medication annotations in Spanish, English, and Italian. The present work contributes all remaining annotations.

\begin{figure}[t]
    \centering
    \includegraphics[width=\linewidth]{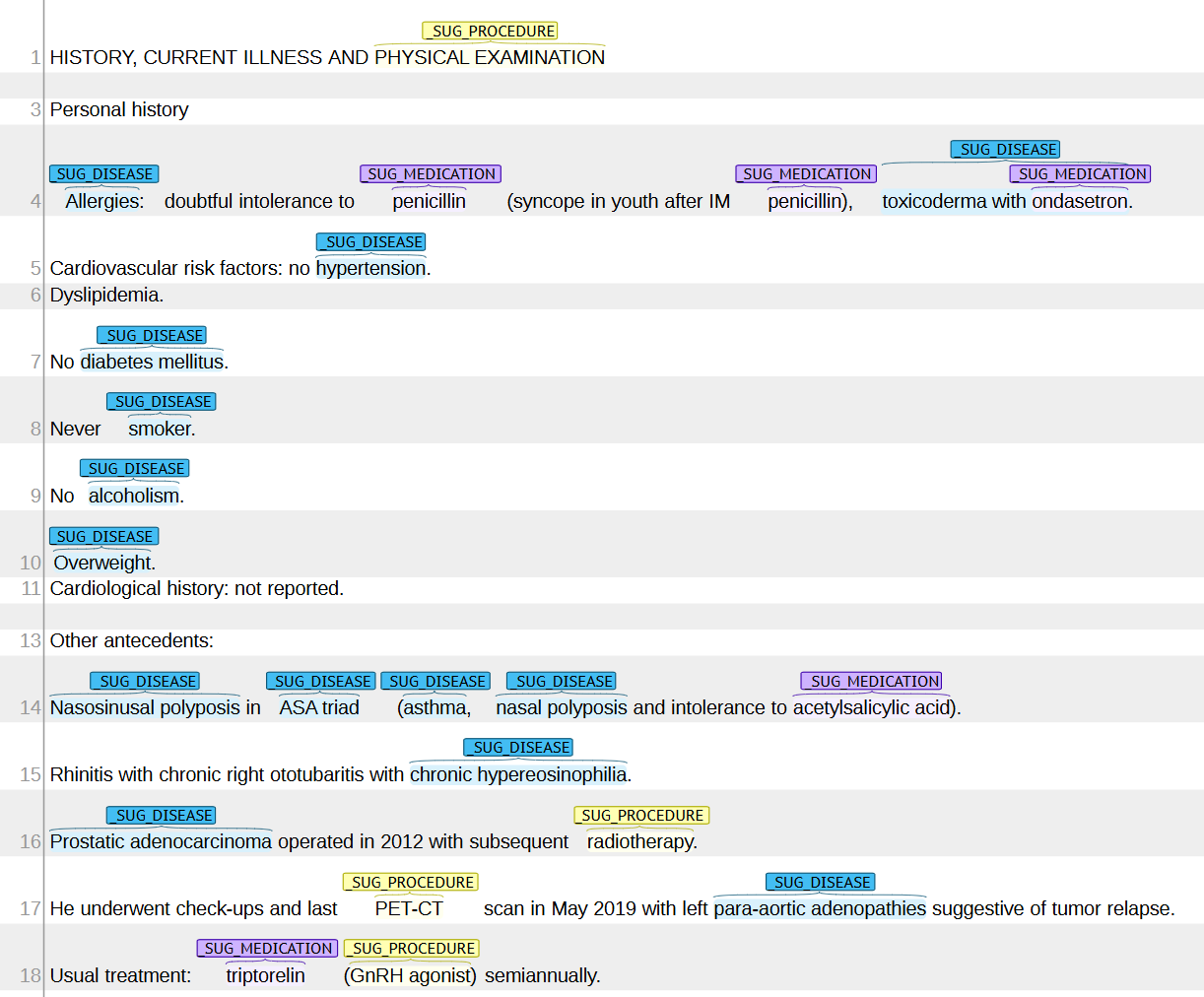}
    \caption{Illustrative English CardioCCC sample showing projected entity annotations for diseases, symptoms and clinical findings, procedures, and medications prior to manual clinical review.}
    \label{fig:cardioccc_english_sample}
\end{figure}

All four categories were first annotated over the Spanish documents by senior clinical annotators experienced in similar tasks, following established guidelines for the corresponding entity types. These guidelines, originally developed for published Spanish benchmark corpora such as DisTEMIST (diseases) \cite{miranda2022distemist}, MedProcNER (procedures) \cite{lima2023medprocner}, SympTEMIST (symptoms) \cite{lima2023symptemist}, and DrugTEMIST (medications) \cite{lima2024overview}, were lightly adapted to cover cardiology-specific content not addressed in the original versions. Once this Spanish Gold Standard was complete, both the case reports and their annotations were automatically translated into each of the other six languages. The translated annotations were then aligned with the corresponding spans in the translated texts using dictionary-based lookup, yielding equivalent annotated corpora in all target languages (Figure~\ref{fig:cardioccc_english_sample} shows an English example). The automatic translations were created using the proprietary tool DeepL\footnote{\url{https://www.deepl.com}}. DeepL was chosen because, at the time of creating the dataset, it was one of the few machine translation systems that allowed for all needed language combinations at once, which was preferred to using different systems for different language combinations.

Native-speaking clinicians for each language then manually reviewed the automatically generated Silver Standard. Using the side-by-side view of the BRAT annotation tool \cite{Stenetorp2012}, with the Spanish Gold Standard version displayed alongside the machine\mbox{-}translated text, clinicians corrected the projected annotations to align as closely as possible with the Spanish original. Ambiguous boundaries, terminology choices, and category assignments raised during review were discussed across the language teams, and the resulting decisions were incorporated into the shared guidelines and applied retrospectively to previously reviewed documents. All in all, a total of 23 different annotators across the 7 languages were involved in the dataset's annotation and projection process.

Beyond validating the mentions themselves, clinicians also flagged mistranslated or non-idiomatic terms and suggested better alternatives. These suggestions form the basis of a second version of the corpus, in which flagged mentions were replaced with the clinician-proposed alternatives. Because these replacements were applied independently for each semantic category, the four annotation layers in the second version of the corpus no longer share an identical underlying text. This prevents its use with our multilabel systems, which predict all four categories over a single input document and therefore require a common text across all annotation layers. To ensure consistency across experiments, we therefore use the first version of the corpus for all systems, including the multiclass models, so that every model is trained and evaluated on the same data.

This machine translation and annotation projection approach offers two key advantages: (i) it accelerates the annotation process by simplifying annotator training and removing the need for language-specific guidelines; and (ii) it produces a dataset that is directly comparable across languages, since the same documents are annotated under identical criteria in every language, making cross-lingual performance comparisons more reliable. 

Table~\ref{tab:cardioccc_stats} reports the total annotation counts of CardioCCC by language and
entity type. Each language includes more than 55,000 mentions across the four categories. Symptoms are the most frequent category in every language and diseases the second most frequent, whereas medications account for only 29,930 mentions, roughly a quarter as many as any other category. In total, the seven-language dataset comprises 400,506 annotations, representing a substantial resource for biomedical NLP in general and for cardiology-focused information extraction in particular. This contribution is especially valuable for lower-resource languages such as Czech, Romanian, and Swedish.

\begin{table}[t]
    \centering
    \caption{Statistics of the multilingual, annotated CardioCCC dataset.}
    \resizebox{\textwidth}{!}{
    \begin{tabular}{l|cccc|c}
    \hline
        Language & DISEASE & MEDICATION & PROCEDURE & SYMPTOM & TOTAL \\  \hline
        \hline
        Czech & 17,788 & 4,290 & 16,477 & 18,729 & 57,284 \\
        Dutch & 17,646 & 4,317 & 16,195 & 18,590 & 56,748 \\
        English & 17,097 & 4,231 & 15,708 & 18,694 & 55,730 \\
        Italian & 18,175 & 4,385 & 16,521 & 19,467 & 58,548 \\
        Romanian & 17,546 & 4,239 & 16,379 & 18,068 & 56,232 \\
        Spanish & 18,232 & 4,227 & 17,073 & 19,983 & 59,515 \\
        Swedish & 17,588 & 4,241 & 16,131 & 18,489 & 56,449 \\
        \hline
        TOTAL & 124,072 & 29,930 & 114,484 & 132,020 & 400,506 \\
        \hline
    \end{tabular}}
    \label{tab:cardioccc_stats}
\end{table}

\subsubsection{Clinical validation data}
The clinical validation dataset was designed to expose the entity extraction systems to complementary forms of cardiology documentation rather than a single source of clinical text. It comprises 20 documents, of which 14 were selected from \mbox{ParaCliTe~\cite{pratesi2026multilingualartificialdatasetsaidriven}} and six from ELCardioCC~\cite{dimitriadis2025overview}. The same document set was evaluated in all seven languages, enabling direct comparison of clinician ratings across languages. The ELCardioCC documents were machine translated into the target languages using Salamandra-7B-Instruct \cite{gonzalez2025salamandra}.

\begin{figure}[t]
    \centering
    \includegraphics[width=\linewidth]{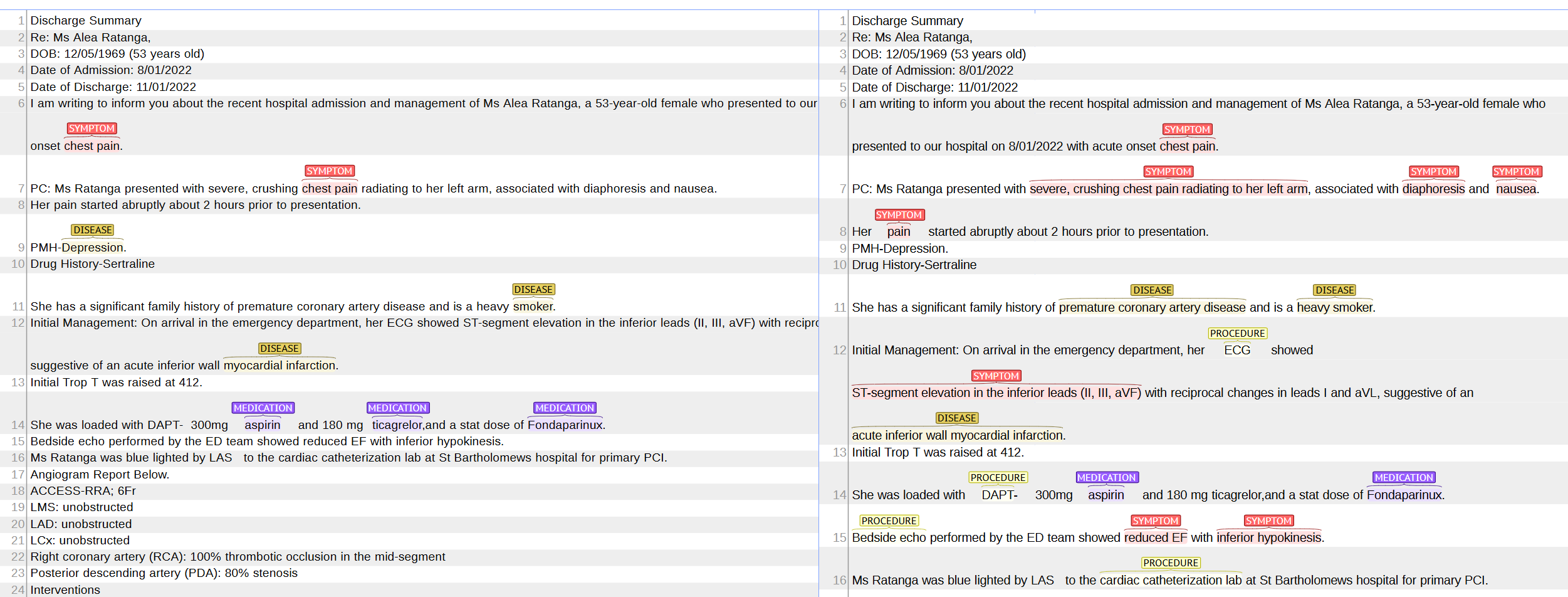}
    \caption{Illustrative comparison of the lexical baseline (left) and the Transformer-based CardioNER system (right) on an English ParaCliTe report.}
    \label{fig:clinical_validation_example}
\end{figure}

Combining documents from two distinct sources preserves variation in both provenance and linguistic generation process. ParaCliTe~\cite{pratesi2026multilingualartificialdatasetsaidriven} provides controlled, privacy\mbox{-}preserving clinical narratives while retaining hospital-oriented reporting conventions and variation across cardiac conditions and encounter types, whereas ELCardioCC~\cite{dimitriadis2025overview} introduces texts derived from real-world discharge documentation. All documents contain mentions of the four semantic categories addressed throughout the study --- diseases, symptoms and findings, clinical procedures, and \mbox{medications ---} but they carry no reference annotations. Accordingly, each document was processed by two entity extraction systems whose outputs were subsequently compared by clinicians: a rule- and dictionary-based baseline and the Transformer-based CardioNER model for the corresponding language. Figure~\ref{fig:clinical_validation_example} illustrates this comparison for an English ParaCliTe report. The output of the lexical baseline is shown on the left, while the predictions generated by the CardioNER system are displayed on the right, allowing the extracted entities from both approaches to be inspected side by side across the four target clinical categories.

To ensure transparency and facilitate reproducibility of the clinical assessment, the complete validation protocol, together with the evaluation files and questionnaires provided to the clinical experts, has been made publicly available\footnote{\url{https://doi.org/10.5281/zenodo.20394665}}.

\subsection{Proposed approaches}

\subsubsection{Domain-adaptive continued pretraining}
\label{subsec:dapt}

To create robust semantic and contextual representations for the cardiology domain across the seven European languages, we selected an existing encoder for each language as the initialisation backbone, based on the availability of suitable checkpoints. Where a biomedical or clinical model was available, we used it (e.g., MedRoBERTa.nl for Dutch and BioBIT for Italian); otherwise, we selected a general-purpose model, preferring a monolingual encoder when available (e.g., RobeCzech for Czech and \mbox{KB-BERT} for Swedish) and a multilingual encoder otherwise (e.g., XLM-RoBERTa-base for Romanian).

As summarised in Table~\ref{tab:base-models-overview}, the selected backbones range from 109M to 270M trainable parameters and include BERT, RoBERTa, and XLM-RoBERTa variants, all belonging to the broader family of bidirectional transformer encoders. Their prior exposure to biomedical and clinical language varies substantially across languages, motivating a common cardiology-specific adaptation stage.

\begin{table}[b]
    \centering
    \caption{Overview of the base models selected for language-specific continued pretraining.}
    \resizebox{\textwidth}{!}{
    \begin{tabular}{l|llllc}
    \hline
        Language & Base-model name & Domain & Architecture & \makecell[l]{Pretraining task} & \makecell[c]{Model size\\(\# trainable\\parameters)} \\ \hline \hline
        \multirow{2}{*}{Czech} & ufal/robeczech-base & \multirow{2}{*}{generic} & \multirow{2}{*}{RoBERTa} & \multirow{2}{*}{MLM} & \multirow{2}{*}{125M} \\
        & \cite{straka2021robeczech} & & & & \\
        \hline
        \multirow{2}{*}{Dutch} & CLTL/MedRoBERTa.nl & \multirow{2}{*}{biomedical} & \multirow{2}{*}{RoBERTa} & \multirow{2}{*}{MLM} & \multirow{2}{*}{125M} \\
        & \cite{Verkijk2022} & & & & \\
        \hline
        \multirow{2}{*}{English} & allenai/biomed\_roberta\_base & \multirow{2}{*}{biomedical} & \multirow{2}{*}{RoBERTa} & \multirow{2}{*}{MLM} & \multirow{2}{*}{125M} \\
        & \cite{gururangan2020don} & & & & \\
        \hline
        \multirow{2}{*}{Italian} & IVN-RIN/bioBIT & \multirow{2}{*}{biomedical} & \multirow{2}{*}{BERT} & \multirow{2}{*}{MLM, NSP} & \multirow{2}{*}{109M} \\
        & \cite{buonocore2023localizing} & & & & \\
        \hline
        \multirow{2}{*}{Romanian} & FacebookAI/xlm-roberta-base & \multirow{2}{*}{generic} & \multirow{2}{*}{XLM-RoBERTa} & \multirow{2}{*}{MLM} & \multirow{2}{*}{270M} \\
        & \cite{conneau2020unsupervised} & & & & \\
        \hline
        \multirow{2}{*}{Spanish} & \makecell[l]{PlanTL-GOB-ES/roberta-base-\\biomedical-clinical-es} & \multirow{2}{*}{biomedical} & \multirow{2}{*}{RoBERTa} & \multirow{2}{*}{MLM} & \multirow{2}{*}{125M} \\
        & \cite{carrino2021biomedical} & & & & \\
        \hline
        \multirow{2}{*}{Swedish} & KB/bert-base-swedish-cased & \multirow{2}{*}{generic} & \multirow{2}{*}{BERT} & \multirow{2}{*}{MLM, NSP} & \multirow{2}{*}{110M} \\
        & \cite{malmsten2020playing} & & & & \\
    \hline
    \end{tabular}
    }
    \label{tab:base-models-overview}
\end{table}

For domain-adaptive continued pretraining, we used a Masked-Language \mbox{Modelling} (MLM) objective for all languages. Through this self-supervised objective, the models learn to reconstruct masked input tokens from their surrounding bidirectional context, progressively adapting their lexical and contextual representations to cardiology\mbox{-}specific terminology and clinical writing patterns. CPT was performed independently for each target language on the corresponding monolingual cardiology and biomedical corpus described in Section~\ref{dapt-data}, with corpus size and composition varying across languages according to the availability of suitable clinical and biomedical data. Where possible, part of the data used to pretrain the original backbone was mixed with the newly incorporated in-domain corpus, thereby reducing the risk of losing previously acquired representations during further specialisation \cite{Alexandrov2024}. However, this was not always feasible, as some original pretraining datasets were unavailable due to copyright, licensing, or patient-privacy restrictions.

All models except the Dutch one were adapted in a single stage. The Dutch CardioBERTa model required a two-stage adaptation procedure, because part of its in\mbox{-}domain data consisted of institutional clinical texts that could not leave the local data infrastructure. In the first stage, CPT was conducted off-premise on publicly available clinical corpora; in the second, the resulting model was further adapted on\mbox{-}premise on the institutional clinical texts, with a subset of the public corpus from the first stage interleaved to limit catastrophic forgetting. 

\begin{figure}[!t]
  \centering
  \includegraphics[width=\linewidth]{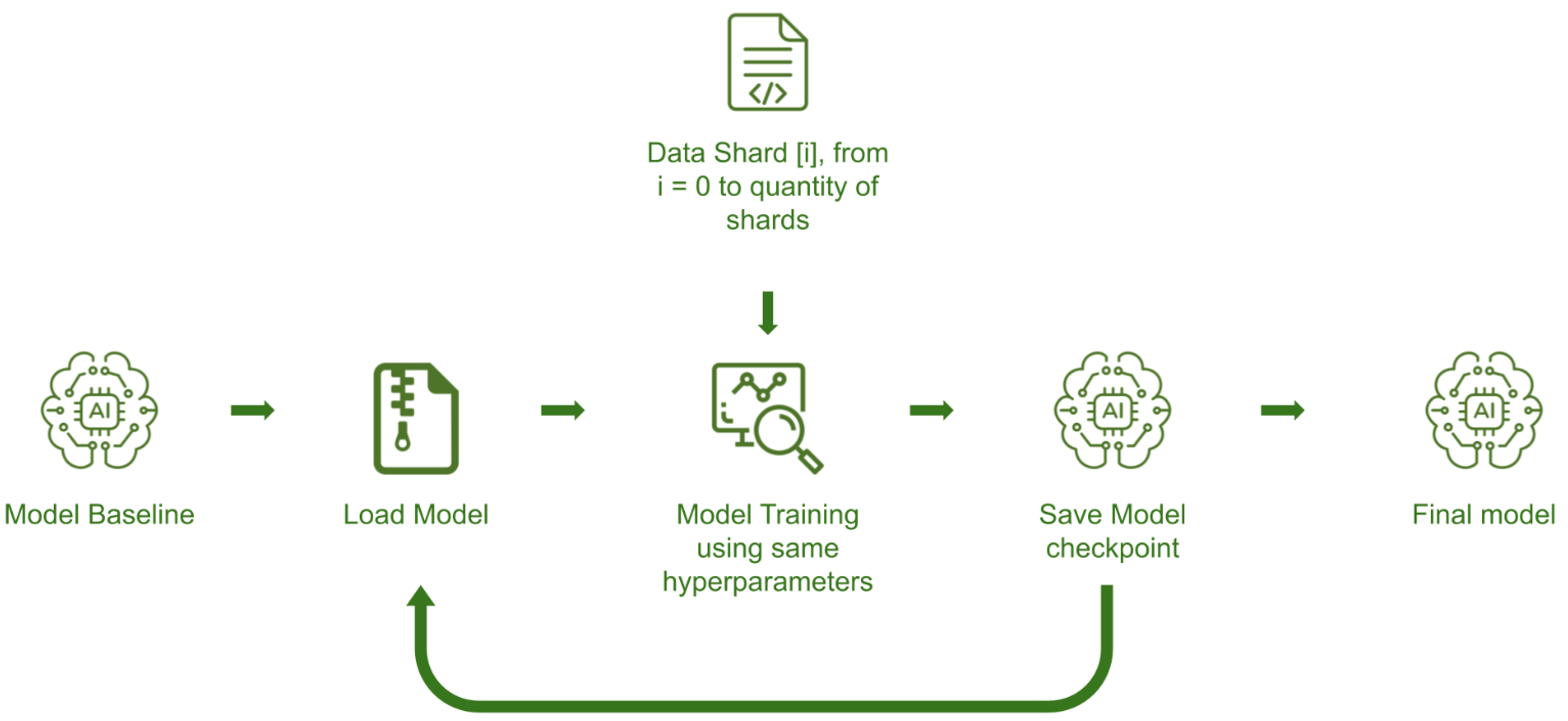}
  \caption{Shard-wise continued pretraining: the checkpoint produced on each corpus shard initialises training on the next, until all shards have been consumed.}
  \label{fig:cpt-shards}
\end{figure}

Training within each adaptation stage was performed sequentially over corpus shards to accommodate large-scale datasets while maintaining reproducibility and manageable runtime constraints. Training proceeded sequentially over these shards, as shown in Figure~\ref{fig:cpt-shards}: the base model was trained on the first shard for five epochs, the resulting checkpoint was used to initialise training on the next shard, and the process was repeated over all data shards until the final CardioBERTa model for that language was obtained.

The continued pretraining workflow was executed on different hardware. The Czech, English, Italian, Romanian, Spanish, and Swedish models were trained on the MareNostrum 5 supercomputing infrastructure, using distributed PyTorch training on eight nodes with four NVIDIA H100 GPUs each, for a total of 32 GPUs. The Dutch model was trained on the Google TPU research cloud for the publicly available data, and on a single NVIDIA RTX 4000 Ada for the on-premise medical records. Hyperparameters were largely standardized across languages. Documents were chunked with a sliding window of 512 tokens and a stride of 256. Training used a masking probability of $15\%$, a batch size of 16 per GPU, and the AdamW optimiser with a weight decay of $1\times10^{-3}$ and a peak learning rate of $1\times10^{-4}$, following a linear warm-up over the first 5{,}000 steps and linear decay thereafter.

The resulting CardioBERTa models are publicly available on Hugging Face\fref{https://huggingface.co/collections/DT4H/cardioberta-family}.

\subsubsection{Named entity recognition}
We framed clinical NER as a token-classification task spanning four clinically relevant entity types: diseases, symptoms, procedures, and medications. Each token was labeled using the standard BIO tagging scheme, which assigns, per entity type, one of three positional tags: B (beginning of an entity), I (inside an entity), or O (outside any entity). This scheme is widely adopted due to its explicit boundary marking, seamless alignment with sequence labeling architectures, and its interoperability across NLP tools and systems.  On top of each
transformer encoder, we added a token-classification head, consisting of a linear layer that maps each token's contextual representation to the BIO label space. The model outputs per-token probability distributions from which entity labels are predicted. 

We evaluated three model configurations. The first fine-tuned the cardiology\mbox{-}adapted language models described in Section \ref{subsec:dapt} on CardioCCC. The second fine-tuned their original backbones on the same data to quantify the contribution of cardiology-specific domain adaptation to system performance. Both configurations are monolingual, with separate models trained and evaluated for each of the seven languages. The third configuration is multilingual, using XLM\mbox{-}RoBERTa\mbox{-}large \cite{conneau2020unsupervised} as a shared backbone, fine-tuned on the combined data from all seven languages. This backbone received no cardiology-specific pretraining. 

Each configuration was trained under two label formulations that differ in the token-level output layer. In the multiclass setting, each entity type is handled by a separate model whose classifier applies a softmax over its BIO labels, assigning a single label per token. In the multilabel setting, a single model jointly predicts all entity types through independent per-type sigmoid outputs, allowing multiple entity labels per token. This experimental design enabled us to assess the effects of cardiology\mbox{-}specific domain adaptation, multilingual training, and label formulation on clinical NER performance.
 
\begin{figure}[t]
  \centering
  \includegraphics[width=\linewidth]{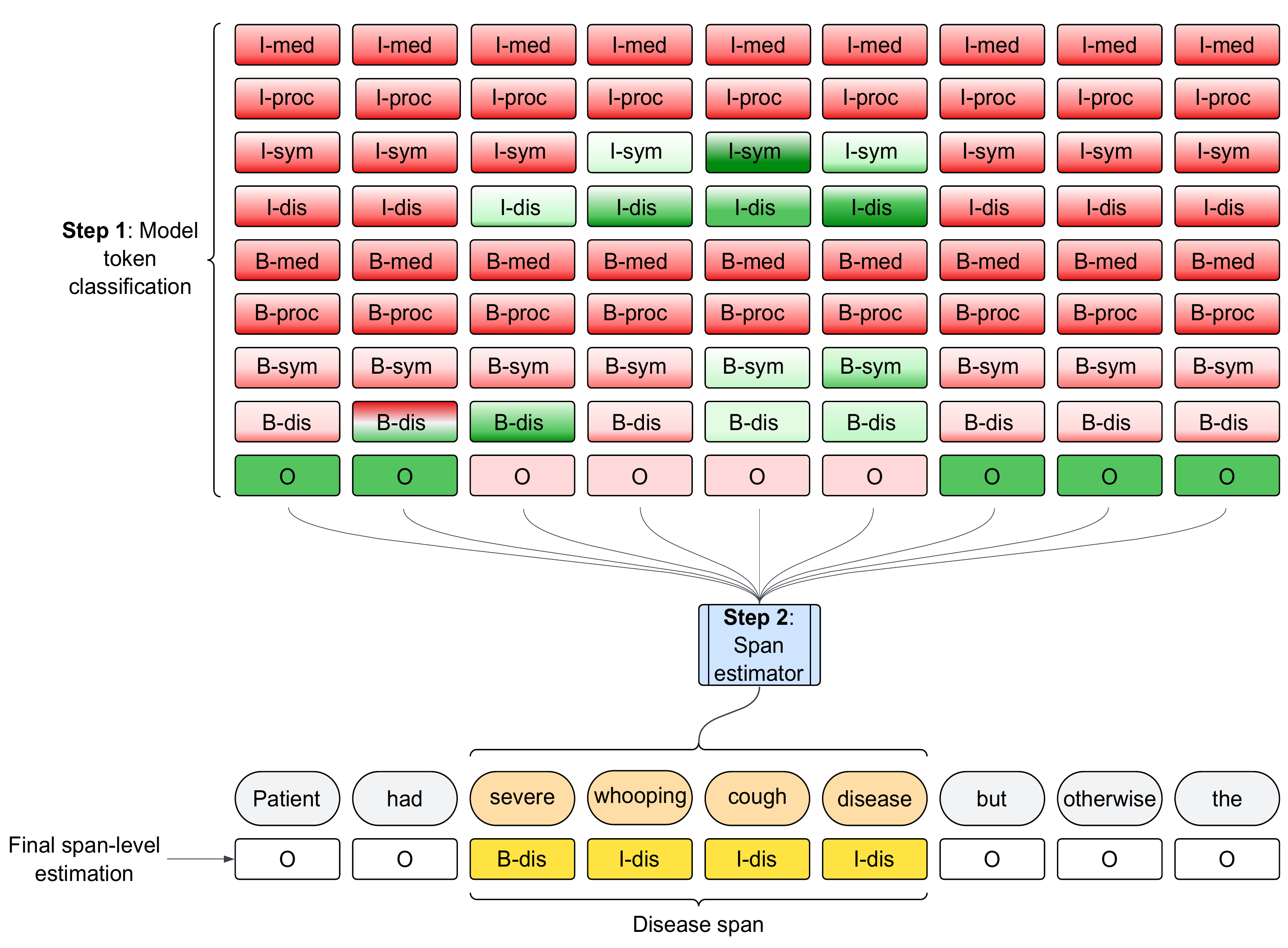}
  \caption{Two-stage token-to-span decoding, comprising (1) token-level BIO prediction and (2) span estimation and post-processing.}
  \label{fig:pipeline}
\end{figure}

The NER pipeline performs token-to-span decoding in two stages, as shown in Figure~\ref{fig:pipeline}. The first stage is token-level prediction: the encoder processes the input text and the classification head assigns BIO label probabilities to each sub-word token. The second stage is span estimation and post-processing: token-level predictions are aggregated into entity spans through a combination of heuristics, including probability thresholds for accepting token classes, resolution of inconsistent tag sequences (e.g., converting orphan I-tags to B-tags), and special character handling. The resulting spans are converted from BIO format back to character‑offset annotations in BRAT format for evaluation.

The pipeline operates on document-level annotations and chunks long clinical reports into paragraph-level segments before encoding, using a cascade strategy: each text is divided into paragraphs and, where a segment still exceeds the 256-token context window, successively into lines and then sentences, so that at each level the coarsest unit that fits within the window is retained. This preserves as much surrounding context as possible in each input sequence. Each retained segment is further split into word-level tokens, while preserving character offsets with respect to the original report, and labeled in BIO format from the available BRAT annotations. These word-level tokens are subsequently tokenized into model-specific sub-word tokens such as BPE (Byte-Pair Encoding), WordPiece, or SentencePiece (Unigram), depending on the base model, with the word-level BIO labels propagated to the corresponding sub-word tokens. 

Our fine-tuning strategy performs a full model-weight update, including the classification head. Across languages, models share a common training setup: 10 epochs, a batch size of 16, an initial learning rate of $2\times10^{-5}$ with a linear decay schedule and no warm-up, and a weight decay of $1\times10^{-4}$. The only per-language adjustment was for Czech, which required a higher initial learning rate of $7\times 10^{-5}$ for convergence. Training used ten-fold cross-validation over 90\% of the data, with the remaining 10\% held out as a fixed test set that was never seen during training. Folds were assigned at the document level over the 508 clinical cases and kept identical across all seven languages, ensuring that all translations of a given clinical case remained in the same fold and preventing cross-lingual information leakage in the jointly trained multilingual model. Each of the ten fold-level models was evaluated separately on the held-out test set, and Section~\ref{ner_performance} reports the mean performance across all folds. For distribution, we release a single model per configuration, obtained by averaging the ten fold-level checkpoints weight by weight. Because this merged checkpoint was not evaluated separately, the reported scores characterise the fine-tuning procedure rather than the released model itself, whose performance may slightly differ.

NER systems fine-tuning was performed on a range of GPU hardware, depending on the compute available for each language: the Czech and Romanian models were trained on an NVIDIA RTX A5000 (24 GB); the Dutch and Swedish models were trained across two machines, one with an NVIDIA RTX 4000 Ada and one with an RTX 2080 Super; the Spanish models on a single NVIDIA H100 (64 GB) on the BSC MareNostrum 5 cluster; the English models on a single Apple Silicon node with a 14-core integrated GPU and 16 GB of unified memory; and the Italian models on an NVIDIA RTX 3090 GPU.

We refrained from hyperparameter optimisation, leaving this to the NLP community who can exploit our openly released datasets, code\fref{https://github.com/DataTools4Heart/CardioLM}, and model weights\fref{https://huggingface.co/collections/DT4H/cardioberta-family}\textsuperscript{,}\fref{https://huggingface.co/collections/DT4H/cardioner}.

\subsubsection{NER inference}
We implemented a configurable inference framework for clinical NER that supports two backends: a standard Hugging Face token-classification pipeline with multiple aggregation strategies (simple, average, first, max), and a custom pipeline with BIO\mbox{-}based entity aggregation, confidence-aware decoding, and rule-based post\mbox{-}processing. All reported results in this paper used the custom pipeline, which we describe below. 

For long documents, input text is split into non-overlapping chunks of at most 125 tokens using a cascading text splitter that respects natural boundaries (paragraphs, sentences, punctuation) while staying within model context-length constraints. The preference for discourse-boundary segmentation reduces the risk of splitting an entity mention across chunks. The splitter also supports configurable overlap, but we use zero overlap in all reported experiments. Character-level offsets are preserved throughout processing via offset projection, enabling accurate mapping of predicted mentions back to the original document text.

Within each chunk, the model produces token-level logits, which are decoded uniformly via softmax followed by argmax, yielding exactly one predicted label per token. We applied this single fixed decoding protocol to every configuration, rather than customising decoding per training objective, so that results are comparable across \mbox{multilabel} and multiclass settings at the level of label selection. For models trained with a multilabel objective, this effectively selects the single most activated label, discarding the capacity for overlapping entity types per token. Because argmax is invariant to softmax versus sigmoid normalisation (both are monotonic in the \mbox{logits}), the top\mbox{-}ranked label a multilabel model would select under its native decoding is unchanged; the two differ only in the confidence values subsequently used for thresholding, since softmax redistributes probability mass across co-activated labels. A fixed threshold is therefore stricter for multilabel models, whose independently-trained \mbox{logits} yield flatter softmax distributions, and entities are filtered most aggressively where several labels are active. To quantify this effect, we report a sensitivity analysis in Section~\ref{ner_performance} in which span\mbox{-}level results were recomputed with confidences on the models' native scale, holding label selection and all other decoding steps fixed.

To boost recall at a controlled cost to precision, an O-class fallback mechanism reassigns tokens predicted as ``O'' (no entity) to the highest-scoring entity label, when the confidence for ``O'' falls below 0.70 and that label scores at least 0.50. Token-level predictions are then aggregated into contiguous entity spans using BIO tag sequences. Orphan I-tags (occurring without a preceding B-tag) are promoted to entity starts rather than discarded. Adjacent same-type spans separated by at most whitespace are merged, and isolated single-token label intrusions within an otherwise contiguous span are bridged. Entity acceptance is governed by a dual-threshold scheme: the median per-token confidence across the span must reach at least 0.65, and the minimum per\mbox{-}token confidence must exceed a floor of 0.10. All decoding thresholds were set as fixed defaults and were not optimized.

Finally, rule-based post-processing heuristics are applied to improve precision and reduce spurious extractions. These include: (i) boundary normalisation, which trims leading and trailing whitespace and punctuation and resolves unmatched parentheses; (ii) filtering of low-information spans such as numeric-only or special-character-only sequences; and (iii) language-specific trimming of trailing function words, which removes dangling prepositions and conjunctions from entity boundaries (e.g., Dutch 'van', 'met'; English 'of', 'with'; Romanian 'de', 'cu'). The word lists were manually curated for each of the seven target languages. 

\subsection{Evaluation}
\subsubsection{Language model evaluation (pseudo-perplexity)}
\label{subsec:ppl}

In this work, we use pseudo-perplexity (pseudo-PPL) to assess how well our \mbox{CardioBERTa} language models capture cardiology-specific language in ParaCliTe~\cite{pratesi2026multilingualartificialdatasetsaidriven} across the seven target languages. 

Perplexity (PPL) is commonly used to assess how well a language model captures the distribution of a given text. It can be interpreted as a measure of how ``surprised'' a model is when reconstructing an existing sample from a distribution of interest, with lower values indicating that the model assigns higher probability to the actual tokens given their available context and therefore finds the text more predictable.

For autoregressive language models, standard PPL is computed from the probability of each token conditioned on its preceding context. Masked language models, however, do not define a left-to-right probability distribution over a sequence and instead use pseudo-PPL, which is derived from pseudo\mbox{-}log\mbox{-}likelihood (PLL). PLL is obtained by masking each token in turn and estimating its probability from the remaining bidirectional context \cite{Salazar2020}.

\begin{align}
    \text{Causal language models:}\quad &\text{PPL}(\mathbf{x}) = \exp\left[-\frac{1}{N}\sum_{i=1}^{N} \log p_{\theta}(x_i\mid x_{<i})\right] \\
    \text{Masked language models:}\quad &\text{PLL}(\mathbf{x}) = \sum_{i=1}^{N} \log p_{\theta}(x_i\mid x_{\neq i}) \\[0.8em]
    \quad &\text{pseudo-PPL}(\mathbf{x}) = \exp\left[-\frac{1}{N}\operatorname{PLL}(\mathbf{x})\right]
\end{align}

For each target language, we score every ParaCliTe document in its version in that language  --- the original text for the cases authored in that language, and a translation for the remaining cases --- as full running text. Each document is scored with two independently implemented and complementary pseudo\mbox{-}log\mbox{-}likelihood methods, which differ in how tokens are masked: our own implementation\fref{https://github.com/DataTools4Heart/CardioLM} of the masked\mbox{-}token scheme of Salazar et al.~\cite{Salazar2020}, in which each token is masked in turn and the model's ability to recover it is measured, and the minicons\fref{https://github.com/kanishkamisra/minicons} library developed by Misra~\cite{Misra2022}, which provides a comparable but independently developed pseudo\mbox{-}log\mbox{-}likelihood estimator using the masking approach proposed by Kauf and Ivanova~\cite{Kauf2023}. Because clinical documents frequently exceed a model's context window, both methods score long documents using overlapping sliding windows whose per-window scores are then aggregated.  

Perplexity has some caveats; an important one is that a generic language model might have a highly skewed probability distribution towards the most likely token with few similar alternatives, and therefore assign a high probability to the matching token. In contrast, a domain-specific model might have a much smoother distribution where the masked token is just one of several tokens with a relatively high probability. As a result, (pseudo-)perplexity scoring might spuriously favor general-purpose over domain-specific models. 

One possible way to mitigate this limitation would be to create a perplexity-change score that considers a top-$N$ model selection of tokens and to measure the change in perplexity of the surrounding words when any of the top-$N$ tokens is selected to replace the initially masked token. This is, however, beyond the scope of this work.

Another way to offset this bias, and the approach we apply in this work, is to distinguish between medical and generic terms. We expect (larger) general-purpose models to be less ``surprised'' by nonmedical words than by specialised medical terms. Since most words in medical letters are in fact nonmedical, a single overall perplexity score would be dominated by these words and could mask differences on the medical content. We therefore compute pseudo-PPL at two granularities: the \textit{general} score, measured over the whole document, and the \textit{medical-span} score, restricted to medical entity spans. Because ParaCliTe carries no gold entity annotations, we first obtain the medical spans automatically by running the multilingual multilabel XLM\mbox{-}RoBERTa\mbox{-}large NER model\fref{https://huggingface.co/DT4H/cardio-ner-multilingual-xlm-roberta-large-multilabel} trained in this work over the target-language documents. We use this model because it achieves the best overall performance across the seven target languages and jointly predicts all four entity types (Section~\ref{multilabel_systems}). We then compute pseudo-perplexity by masking and scoring only the tokens within the predicted spans, using the full document as context. Medical-span PPL is therefore a silver-standard measure: it depends on the accuracy of the predicted spans and inherits any boundary errors, false positives, or missed mentions produced by the NER model, a dependency that should be kept in mind when interpreting these results.

A second caveat concerns differences in tokenizer coverage. When a tokenizer is applied to an underrepresented domain, words are more likely to be fragmented into multiple subword tokens. Each subword is then easier to predict once the other subwords are visible in the context, which inflates its probability and artificially lowers perplexity. To offset this, the minicons estimator applies a word-oriented perplexity scoring, developed by Kauf and Ivanova~\cite{Kauf2023}, in which the subsequent subwords of the same word are also masked. Our own implementation retains the original token-level masking of Salazar et al.~\cite{Salazar2020}.

At both granularities, we report overall results as well as per-language results. For the latter, we group the resulting per-document scores by each document's source language, isolating the effect of translation provenance on predictability. We report the geometric mean and the median of the per-document scores, deliberately omitting the arithmetic mean, since the geometric mean and median are more robust to the heavy right tail characteristic of perplexity distributions.

\subsubsection{NER evaluation metrics}
The NER systems were evaluated using standard information extraction metrics: \textit{\mbox{precision}, recall, and F1-score}. These metrics were computed along two independent dimensions. The first dimension is the matching criterion, which determines when a predicted entity counts as correct, and comprises two types of evaluation: (1) \textit{strict} evaluation, in which a prediction is considered correct \textit{only if} both the entity type and the exact character offsets match the gold-standard annotation, and (2) \textit{relaxed} (character-overlap) evaluation, in which a prediction is considered correct if the entity type is correct and there is any overlap between the predicted and gold-standard spans. The first one  is the most demanding evaluation setting, penalizing any boundary detection errors, while the latter captures cases where the model correctly identifies the entity type and approximate location but fails to match exact boundaries.

The second dimension is the aggregation method, which determines how scores are combined across entity types: (1) \textit{micro-averaging} (metrics are computed globally across all entity instances, giving more weight to frequent entity types), and (2) \textit{macro‑averaging} (metrics are computed per entity type and then averaged, giving equal weight to each type regardless of frequency). Both dimensions are reported in combination to provide a comprehensive view of performance, particularly in the presence of class imbalance across entity types.

\subsubsection{External clinical validation}
\label{sub:ext_clin_validation}
Beyond the intrinsic evaluation of the language models and the automatic NER evaluation against gold-standard CardioCCC annotations, we assessed the entity extraction outputs through a clinician-centred validation. The evaluation was performed on a multilingual sample combining synthetic hospital-like records from ParaCliTe~\cite{pratesi2026multilingualartificialdatasetsaidriven} with anonymised cardiology discharge letters from ELCardioCC~\cite{dimitriadis2025overview}, originally written in Greek and machine-translated into the seven target languages using Salamandra\mbox{-}7B\mbox{-}Instruct \cite{gonzalez2025salamandra}. The objective of this evaluation was to assess whether the Transformer-based approach produced annotations that were not only quantitatively accurate according to standard NER metrics, but also clinically meaningful and potentially useful for downstream hospital-oriented data structuring workflows. 

The validation followed a comparative design in which clinicians assessed the outputs of two systems: (i) a rule- and dictionary-based baseline extractor that matched text mentions against a predefined list of variables of interest, using exact string matching followed by approximate matching based on normalised Levenshtein similarity \cite{lcvenshtcin1966binary} with a threshold of 0.8; and (ii) the Transformer-based NER system developed in this work. Both systems were evaluated with respect to the four target semantic categories: diseases, medications, procedures, and symptoms. 

A structured qualitative questionnaire was completed by clinical evaluators for each of the seven languages, with one assessment per language, resulting in a balanced multilingual validation sample of seven evaluations. This design ensured that the assessment was not dominated by a single linguistic or institutional context, although the resulting sample size means that the findings should be interpreted descriptively rather than as a formal statistical comparison between systems. Clinicians rated the perceived performance of each system on a 0--10 scale for each entity type and for the output as a whole. We report the per-entity ratings and their mean across entity types, expressed as percentages of the maximum score. These scores reflect a subjective assessment of multiple aspects, including entity detection, label assignment, and span boundaries, rather than an exact-match accuracy measure.

\section{Results}

\subsection{Pseudo-perplexity of base and domain-adapted models}
\label{ppl_results}

\begin{table}[h]
    \centering
    \caption{Pseudo-perplexity aggregation over \textbf{all spans}.}
    \label{tab:perplexity_generic}
    \resizebox{\textwidth}{!}{
    \begin{tabular}{l|l|cc}
    Language     & Model & PPL$_{minicons}$ median(geometric mean) & PPL$_{ours}$ median(geometric mean) \\
         \hline
\hline
    \multirow{2}{*}{Czech} & robeczech-base & 56.49 (56.17) & 22.68 (23.77) \\
    & CardioBERTa.cs & \textbf{6.94} (\textbf{6.96}) & \textbf{3.22} (\textbf{3.53}) \\
    \hline
    \multirow{3}{*}{Dutch} & RobBERT2023-base & 12.70 (12.50) & \textbf{6.40} (\textbf{6.20}) \\
    & MedRoBERTa.nl & 44.70 (45.40) & 30.70 (30.60) \\
    & CardioBERTa.nl & \textbf{9.80} (\textbf{9.40}) & 6.80 (6.80) \\
    \hline
    \multirow{2}{*}{English} & biomed\_roberta\_base & 29.76 (29.88) & 16.13 (16.69) \\
    & CardioBERTa.en & \textbf{4.32} (\textbf{4.07}) & \textbf{2.77} (\textbf{2.72}) \\
    \hline
    \multirow{2}{*}{Italian} & IVN/bioBIT & \textbf{11.15} (\textbf{11.09}) & \textbf{4.87} (\textbf{4.92}) \\
    & CardioBERTa.it & 11.62 (11.50) & 5.48 (5.41) \\
    \hline
    \multirow{2}{*}{Romanian} & xlm-roberta-base & 15.41 (14.67) & 5.42 (5.31) \\
    & CardioBERTa.ro & \textbf{7.55} (\textbf{7.16}) & \textbf{3.05} (\textbf{2.92}) \\
    \hline
    \multirow{2}{*}{Spanish} & \makecell[l]{PlanTL-GOB-ES/roberta-\\base-biomedical-clinical-es} & 27.92 (36.66) & 33.72 (46.62) \\
    & CardioBERTa.es & \textbf{4.00} (\textbf{4.02}) & \textbf{5.51} (\textbf{5.48}) \\
    \hline
    \multirow{2}{*}{Swedish} & bert-base-swedish-cased & 49.20 (48.10) & 31.30 (30.90) \\
    & CardioBERTa.sv & \textbf{10.10} (\textbf{11.10}) & \textbf{4.50} (\textbf{5.00}) \\
    \hline
    \end{tabular}
    }
\end{table}

\begin{table}[h]
    \centering
    \caption{Pseudo-perplexity aggregation over \textbf{all medical spans}.}
    \label{tab:perplexity_medical}
    \resizebox{\textwidth}{!}{
    \begin{tabular}{l|l|cc}
    Language     & Model & PPL$_{minicons}$ median(geometric mean) & PPL$_{ours}$ median(geometric mean) \\
         \hline
\hline
    \multirow{2}{*}{Czech} & robeczech-base & 38.88 (49.41) & 4.90 (10.42) \\
    & CardioBERTa.cs & \textbf{9.09} (\textbf{14.37}) & \textbf{1.71} (\textbf{4.25}) \\
    \hline
    \multirow{3}{*}{Dutch} & RobBERT2023-base & 55.4 (62.1) & 9.8 (17.4) \\
    & MedRoBERTa.nl & 65.2 (87.5) & 24.9 (43.6) \\
    & CardioBERTa.nl & \textbf{17.6} (\textbf{30.3}) & \textbf{8.0} (\textbf{17.2}) \\
    \hline
    \multirow{2}{*}{English} & biomed\_roberta\_base & 12.77 (21.64) & 2.43 (6.69) \\
    & CardioBERTa.en & \textbf{3.54} (\textbf{6.28}) & \textbf{1.26} (\textbf{2.84}) \\
    \hline
    \multirow{2}{*}{Italian} & IVN/bioBIT & 15.52 (\textbf{25.40}) & \textbf{2.51} (\textbf{6.60}) \\
    & CardioBERTa.it & \textbf{15.44} (26.93) & 2.81 (7.61) \\
    \hline
    \multirow{2}{*}{Romanian} & xlm-roberta-base & 40.11 (51.02) & 4.55 (10.55) \\
    & CardioBERTa.ro & \textbf{8.81} (\textbf{13.94}) & \textbf{1.36} (\textbf{3.43}) \\
    \hline
    \multirow{2}{*}{Spanish} & \makecell[l]{PlanTL-GOB-ES/roberta-\\base-biomedical-clinical-es} & 14.16 (75.06) & 6.72 (44.53) \\
    & CardioBERTa.es & \textbf{6.88} (\textbf{13.61}) & \textbf{3.43} (\textbf{8.33}) \\
    \hline
    \multirow{2}{*}{Swedish} & bert-base-swedish-cased & 209.2 (188) & 101 (96.4) \\
    & CardioBERTa.sv & \textbf{10.3} (\textbf{17.8}) & \textbf{1.7} (\textbf{4.6}) \\
    \hline
    \end{tabular}}
\end{table}

Tables~\ref{tab:perplexity_generic} and~\ref{tab:perplexity_medical} show that continued pretraining generally reduces pseudo-perplexity relative to the corresponding base models. For whole-document scoring, CardioBERTa obtains lower scores under both implementations in five of the seven languages; Dutch shows mixed results across implementations, while Italian is the main exception, with BioBIT obtaining slightly lower scores. The pattern is more consistent for medical spans, where CardioBERTa outperforms the corresponding base model in six of the seven languages under both implementations. Italian again shows only marginal differences, possibly because BioBIT was pretrained on translated PubMed abstracts, similar to the data used for the continued pretraining of CardioBERTa.it. For medical spans, the gap between the median and geometric mean is also generally larger than for whole documents. This suggests greater variability and right skew in pseudo-PPL across documents, likely because some medical content is substantially more difficult to predict. Although the two implementations produce different absolute pseudo-PPL values, they lead to consistent conclusions regarding the effect of continued pretraining.

\subsection{NER system performance}
\label{ner_performance}
We report NER performance primarily at the span level, since the units of interest for extraction are entity mentions rather than individual tokens. For completeness, token\mbox{-}level results aggregated across all entity types are reported in Appendix~\ref{app:appendix_results} (Table~\ref{tab:ner-token-level}). 

For the main results, we decode the outputs of every model with a single-label protocol (argmax over softmax-normalised scores), applied identically to the multiclass and multilabel models. Because the multilabel models are natively trained with a per-label sigmoid objective, we verified that scoring them under this single\mbox{-}label softmax protocol, rather than on their native sigmoid scale, does not affect the multiclass\mbox{--}multilabel comparison. Recomputing the multilabel predictions on their native sigmoid scale (Appendix~\ref{app:appendix_results}, Table~\ref{tab:span_multilabel_sigmoid_results}) leaves overall F1 essentially unchanged relative to the single-label softmax protocol (Table~\ref{tab:span_multilabel_softmax_results}): strict micro-F1 differs by at most 0.01 and relaxed micro-F1 by at most 0.03. The strict micro-F1 ranking within each language is preserved, and XLM\mbox{-}RoBERTa\mbox{-}large remains the best-performing model in every language under both protocols. The main difference lies instead in a precision\mbox{--}recall trade-off, with precision rising by up to 0.04 and recall falling by up to 0.06 for all models except the two Italian monolingual models, for which recall rises slightly. These shifts are largest for the monolingual models and negligible for XLM\mbox{-}RoBERTa\mbox{-}large. At the entity level, the precision gain is largest for symptoms and smallest for medications, while recall losses are of similar magnitude across entity types (Table~\ref{tab:span_multilabel_softmax_per_entity}; Appendix~\ref{app:appendix_results}, Table~\ref{tab:span_multilabel_sigmoid_per_entity}). We therefore report results under the single\mbox{-}label softmax decoding protocol throughout, noting that the decoding choice shifts precision and recall in opposite directions while leaving overall F1 nearly unchanged and the strict ranking of models intact.

\paragraph{Multilabel systems}
\label{multilabel_systems}
\begin{table}[t]
    \centering
    \caption{\textbf{Span-level NER} performance of \textbf{multilabel} models on the CardioCCC held-out test set, averaged across the 10 fold models. Predictions are decoded using a \textbf{single-label protocol} (argmax over \textbf{softmax-normalised logits}) and aggregated into entity spans. Micro- and macro-averages are computed over the four entity types, with strict (exact boundary) and relaxed (overlapping boundary) precision, recall, and F1 reported. XLM-RoBERTa-large is trained jointly on all seven languages and evaluated on the full multilingual test set and separately by language.
    }
    \resizebox{\textwidth}{!}{
    \begin{tabular}{l|l|ccc|ccc}
    \multicolumn{2}{c}{} & \multicolumn{3}{c}{Micro} & \multicolumn{3}{c}{Macro} \\
    \hhline{--|---|---}
    \multirow{2}{*}{Language} & \multirow{2}{*}{Model} & F1 & Recall & Precision & F1 & Recall & Precision \\
    & & strict (relaxed) & strict (relaxed) & strict (relaxed) & strict (relaxed) & strict (relaxed) & strict (relaxed) \\ 
    \hline
\hline  
    \multirow{3}{*}{Czech} & robeczech-base & 0.68 (0.84) & 0.67 (0.82) & 0.70 (0.85) & 0.72 (0.85) & 0.70 (0.83) & 0.73 (0.87) \\
    & CardioBERTa.cs & 0.70 (0.84) & 0.68 (0.83) & \textbf{0.71} (\textbf{0.86}) & 0.72 (0.86) & 0.71 (0.84) & \textbf{0.74} (\textbf{0.88}) \\
    & xlm-roberta-large & \textbf{0.71} (\textbf{0.86}) & \textbf{0.70} (\textbf{0.85}) & \textbf{0.71} (\textbf{0.86}) & \textbf{0.74} (\textbf{0.87}) & \textbf{0.73} (\textbf{0.86}) & \textbf{0.74} (\textbf{0.88}) \\
    \hline
    \multirow{3}{*}{Dutch} & RobBERT2023-base & 0.63 (0.81) & 0.64 (0.82) & 0.63 (0.80) & 0.68 (0.83) & 0.68 (0.83) & 0.68 (0.82) \\
    & CardioBERTa.nl & 0.65 (0.82) & 0.63 (0.80) & 0.67 (0.84) & 0.68 (0.83) & 0.67 (0.81) & 0.70 (0.85) \\
    & xlm-roberta-large & \textbf{0.71} (\textbf{0.86}) & \textbf{0.70} (\textbf{0.85}) & \textbf{0.71} (\textbf{0.87}) & \textbf{0.74} (\textbf{0.87}) & \textbf{0.73} (\textbf{0.85}) & \textbf{0.75} (\textbf{0.88}) \\
    \hline
    \multirow{3}{*}{English} & biomed\_roberta\_base & 0.64 (0.81) & 0.65 (0.82) & 0.63 (0.79) & 0.69 (0.83) & 0.69 (0.84) & 0.68 (0.82) \\
    & CardioBERTa.en & 0.65 (0.81) & 0.65 (0.81) & 0.65 (0.81) & 0.69 (0.84) & 0.69 (0.84) & 0.69 (0.84) \\
    & xlm-roberta-large & \textbf{0.71} (\textbf{0.85}) & \textbf{0.71} (\textbf{0.85}) & \textbf{0.71} (\textbf{0.85}) & \textbf{0.74} (\textbf{0.86}) & \textbf{0.74} (\textbf{0.86}) & \textbf{0.74} (\textbf{0.86}) \\
    \hline
    \multirow{3}{*}{Italian} & bioBIT & 0.63 (0.82) & 0.61 (0.79) & 0.66 (0.86) & 0.68 (0.84) & 0.65 (0.81) & 0.66 (0.86) \\
    & CardioBERTa.it & 0.64 (0.83) & 0.62 (0.80) & 0.67 (\textbf{0.87}) & 0.68 (0.85) & 0.66 (0.81) & 0.67 (0.87) \\
    & xlm-roberta-large & \textbf{0.70} (\textbf{0.86}) & \textbf{0.69} (\textbf{0.85}) & \textbf{0.71} (\textbf{0.87}) & \textbf{0.73} (\textbf{0.87}) & \textbf{0.73} (\textbf{0.86}) & \textbf{0.74} (\textbf{0.88}) \\
    \hline
    \multirow{3}{*}{Romanian} & xlm-roberta-base & 0.65 (0.82) & 0.65 (0.83) & 0.64 (0.81) & 0.69 (0.83) & 0.70 (0.84) & 0.69 (0.83) \\
    & CardioBERTa.ro & 0.66 (0.83) & 0.67 (\textbf{0.85}) & 0.65 (0.82) & 0.70 (0.85) & 0.72 (\textbf{0.86}) & 0.69 (0.83) \\
    & xlm-roberta-large & \textbf{0.70} (\textbf{0.85}) & \textbf{0.69} (0.84) & \textbf{0.71} (\textbf{0.86}) & \textbf{0.74} (\textbf{0.86}) & \textbf{0.73} (\textbf{0.86}) & \textbf{0.74} (\textbf{0.88}) \\
    \hline
    \multirow{3}{*}{Spanish} & roberta-biomedical-clinical-es & 0.73 (0.88) & 0.73 (0.87) & 0.74 (0.88) & 0.76 (\textbf{0.89}) & 0.75 (0.87) & 0.77 (0.90) \\
    & CardioBERTa.es & 0.73 (0.88) & 0.72 (0.87) & 0.74 (\textbf{0.89}) & 0.76 (0.88) & 0.74 (0.86) & \textbf{0.78} (\textbf{0.91}) \\
    & xlm-roberta-large & \textbf{0.75} (\textbf{0.89}) & \textbf{0.75} (\textbf{0.89}) & \textbf{0.75} (\textbf{0.89}) & \textbf{0.78} (\textbf{0.89}) & \textbf{0.77} (\textbf{0.89}) & \textbf{0.78} (0.90) \\
    \hline
    \multirow{3}{*}{Swedish} & bert-base-swedish-cased & 0.66 (0.82) & 0.66 (0.82) & 0.65 (0.81) & 0.70 (0.83) & 0.70 (0.84) & 0.69 (0.83) \\
    & CardioBERTa.sv & 0.69 (0.85) & \textbf{0.70} (\textbf{0.86}) & 0.68 (0.84) & 0.73 (\textbf{0.87}) & \textbf{0.74} (\textbf{0.88}) & 0.72 (0.86) \\
    & xlm-roberta-large & \textbf{0.71} (\textbf{0.86}) & \textbf{0.70} (0.85) & \textbf{0.71} (\textbf{0.86}) & \textbf{0.74} (\textbf{0.87}) & 0.73 (0.87) & \textbf{0.74} (\textbf{0.87}) \\
    \hline
    Multilingual & xlm-roberta-large & 0.72 (0.88) & 0.71 (0.88) & 0.72 (0.89) & 0.75 (0.89) & 0.75 (0.88) & 0.76 (0.90) \\
    \hline
    \end{tabular}
    }
\label{tab:span_multilabel_softmax_results}
\end{table}

Table~\ref{tab:span_multilabel_softmax_results} reports overall span-level performance for the multilabel NER systems. Strict micro-F1 ranges from 0.63 to 0.75 across languages and models, and XLM\mbox{-}RoBERTa\mbox{-}large is the strongest system in every language. Evaluated on the pooled multilingual test set, the jointly trained XLM\mbox{-}RoBERTa\mbox{-}large reaches 0.72 strict micro-F1, in line with its per-language scores (0.70--0.75); performance is thus balanced across the seven languages, with no single language dominating or lagging the pooled result. Table~\ref{tab:span_multilabel_softmax_per_entity} breaks
this down by entity type. 

The relative difficulty of the entity types is consistent across all seven languages: medication is recognized most reliably (strict F1 between 0.83 and 0.90); procedure and disease are intermediate (0.62--0.79 and 0.62--0.74, respectively); and symptom is the hardest category throughout (0.56--0.70). This ordering also holds across backbones and may be explained by the more standardized and lexically constrained naming of medications, compared with the greater variability and context dependence of the other three types.

Cardiology-specific adaptation yields only modest gains over the corresponding base models. For disease and procedure, improvements reach at most 0.04 strict F1 points and are typically smaller. Symptom performance improves substantially only in Dutch (+0.04) and Swedish (+0.03) and is otherwise largely unchanged. Medication is the notable exception, with adaptation producing inconsistent effects across languages: strict F1 decreases in Dutch (0.87 to 0.83), Spanish (0.87 to 0.84), and Czech (0.86 to 0.84), increases in Swedish (0.86 to 0.90), Italian (0.85 to 0.87), and English (0.87 to 0.88), and remains unchanged in Romanian (0.86). Overall, these results suggest that cardiology-specific adaptation provides limited and inconsistent benefits at the entity level and does not reliably outperform the corresponding base models.

XLM-RoBERTa-large achieves the highest strict F1 for almost every combination of language and entity type, the only exception being medication in Swedish, where CardioBERTa.sv leads (0.90 versus 0.86). Its advantage over the best monolingual model varies considerably by language. It is small in Czech and Spanish (at most 0.02 for any entity type), moderate in Romanian and Swedish (at most 0.04), and reaches 0.06--0.09 in Dutch, English, and Italian, peaking for procedure in Italian (0.72 versus 0.63) and symptom in English (0.67 versus 0.58). For medication, the advantage largely disappears: XLM-RoBERTa-large ties the best monolingual model in Czech, Dutch, and Italian and leads by at most 0.02 elsewhere.

The gap between strict and relaxed scores varies sharply by entity type. For medication it is small (0.02--0.06 F1 points), indicating that mentions are delimited almost exactly when they are found at all. For disease, procedure, and symptom it is far larger (0.11--0.23 F1 points), so errors on these types are dominated by boundary mismatches rather than by entity-type confusion. 

\begin{landscape}
\begin{table}[]
    \centering
    \caption{\textbf{Span-level NER} performance of \textbf{multilabel} models on the CardioCCC held-out test set, broken down by entity type and averaged across the 10 fold models. Predictions are decoded using a \textbf{single-label protocol} (argmax over \textbf{softmax-normalised logits}) and aggregated into entity spans. Strict (exact boundary) and relaxed (overlapping boundary) precision, recall, and F1 are reported for each of the four entity types. XLM\mbox{-}RoBERTa\mbox{-}large is trained jointly on all seven languages and evaluated on the full multilingual test set and separately by language.}
    \label{tab:span_multilabel_softmax_per_entity}
    \resizebox{\columnwidth}{!}{%
    \begin{tabular}{l|l|cccc|cccc|cccc}
     \multicolumn{1}{l}{} & \multicolumn{1}{l}{} & \multicolumn{4}{c}{F1 strict (relaxed)} & \multicolumn{4}{c}{Recall strict (relaxed)} & \multicolumn{4}{c}{Precision strict (relaxed)} \\
    \hline
    Language & Model & Dis & Med & Proc & Sym & Dis & Med & Proc & Sym & Dis & Med & Proc & Sym \\
    \hline
    \hline
    \multirow{3}{*}{Czech} & robeczech-base & 0.67 (0.86) & \textbf{0.86} (0.90) & 0.72 (0.83) & 0.62 (0.80) & 0.67 (0.85) & 0.83 (0.87) & 0.69 (0.80) & 0.61 (0.79) & 0.68 (0.86) & \textbf{0.89} (\textbf{0.94}) & 0.75 (0.87) & 0.62 (0.80) \\
    & CardioBERTa.cs & 0.70 (\textbf{0.87}) & 0.84 (0.90) & 0.73 (0.85) & 0.63 (0.81) & 0.70 (0.87) & 0.83 (0.88) & 0.69 (0.80) & 0.63 (0.80) & 0.70 (\textbf{0.87}) & 0.86 (0.92) & \textbf{0.77} (\textbf{0.89}) & \textbf{0.64} (0.81) \\
    & xlm-roberta-large & \textbf{0.71} (\textbf{0.87}) & \textbf{0.86} (\textbf{0.91}) & \textbf{0.74} (\textbf{0.86}) & \textbf{0.64} (\textbf{0.82}) & \textbf{0.71} (\textbf{0.88}) & \textbf{0.86} (\textbf{0.91}) & \textbf{0.72} (\textbf{0.84}) & \textbf{0.64} (\textbf{0.83}) & \textbf{0.71} (\textbf{0.87}) & 0.87 (0.92) & \textbf{0.77} (\textbf{0.89}) & 0.63 (\textbf{0.82}) \\
    \hline
    \multirow{3}{*}{Dutch} & RobBERT2023-base & 0.62 (0.82) & \textbf{0.87} (0.89) & 0.65 (0.81) & 0.57 (0.78) & 0.62 (0.83) & 0.85 (0.87) & 0.64 (0.80) & 0.61 (0.82) & 0.62 (0.82) & 0.88 (0.90) & 0.67 (0.83) & 0.54 (0.73) \\
     & CardioBERTa.nl & 0.64 (0.84) & 0.83 (0.87) & 0.67 (0.82) & 0.61 (0.79) & 0.65 (0.85) & 0.82 (0.86) & 0.61 (0.76) & 0.59 (0.78) & 0.63 (0.83) & 0.84 (0.88) & 0.72 (\textbf{0.90}) & 0.61 (0.80) \\
     & xlm-roberta-large & \textbf{0.70} (\textbf{0.88}) & \textbf{0.87} (\textbf{0.90}) & \textbf{0.71} (\textbf{0.86}) & \textbf{0.66} (\textbf{0.83}) & \textbf{0.71} (\textbf{0.88}) & \textbf{0.86} (\textbf{0.88}) & \textbf{0.68} (\textbf{0.82}) & \textbf{0.67} (\textbf{0.84}) & \textbf{0.70} (\textbf{0.87}) & \textbf{0.89} (\textbf{0.91}) & \textbf{0.74} (0.89) & \textbf{0.66} (\textbf{0.82}) \\
    \hline
    \multirow{3}{*}{English} & biomed\_roberta\_base & 0.62 (0.81) & 0.87 (0.93) & 0.68 (0.81) & 0.58 (0.78) & 0.62 (0.80) & 0.86 (0.93) & 0.67 (0.80) & 0.62 (0.83) & 0.63 (0.81) & 0.87 (\textbf{0.93}) & 0.69 (0.82) & 0.54 (0.73) \\
    & CardioBERTa.en & 0.62 (0.81) & 0.88 (\textbf{0.94}) & 0.69 (0.81) & 0.58 (0.79) & 0.63 (0.81) & 0.88 (\textbf{0.94}) & 0.67 (0.79) & 0.59 (0.80) & 0.62 (0.81) & \textbf{0.88} (\textbf{0.93}) & 0.71 (0.83) & 0.57 (0.77) \\
    & xlm-roberta-large & \textbf{0.70} (\textbf{0.85}) & \textbf{0.89} (0.91) & \textbf{0.72} (\textbf{0.84}) & \textbf{0.67} (\textbf{0.83}) & \textbf{0.70} (\textbf{0.85}) & \textbf{0.89} (0.92) & \textbf{0.71} (\textbf{0.82}) & \textbf{0.68} (\textbf{0.85}) & \textbf{0.70} (\textbf{0.85}) & \textbf{0.88} (0.91) & \textbf{0.74} (\textbf{0.86}) & \textbf{0.66} (\textbf{0.82}) \\
    \hline
    \multirow{3}{*}{Italian} & bioBIT & 0.64 (0.86) & 0.85 (\textbf{0.90}) & 0.62 (\textbf{0.85}) & 0.61 (0.81) & 0.64 (0.86) & 0.86 (\textbf{0.91}) & 0.60 (0.82) & 0.61 (0.82) & 0.64 (0.86) & 0.85 (\textbf{0.89}) & 0.64 (0.88) & 0.60 (0.81) \\
    & CardioBERTa.it & 0.65 (0.86) & \textbf{0.87} (\textbf{0.90}) & 0.63 (\textbf{0.85}) & 0.61 (0.82) & 0.65 (\textbf{0.87}) & 0.87 (0.90) & 0.60 (\textbf{0.83}) & 0.62 (0.83) & 0.65 (0.86) & \textbf{0.86} (\textbf{0.89}) & 0.64 (0.88) & 0.60 (0.81) \\
    & xlm-roberta-large & \textbf{0.71} (\textbf{0.88}) & \textbf{0.87} (\textbf{0.90}) & \textbf{0.72} (\textbf{0.85}) & \textbf{0.64} (\textbf{0.83}) & \textbf{0.69} (0.86) & \textbf{0.88} (\textbf{0.91}) & \textbf{0.69} (0.82) & \textbf{0.65} (\textbf{0.84}) & \textbf{0.72} (\textbf{0.89}) & \textbf{0.86} (\textbf{0.89}) & \textbf{0.75} (\textbf{0.89}) & \textbf{0.63} (\textbf{0.82}) \\
    \hline
    \multirow{3}{*}{Romanian} & xlm-roberta-base & 0.66 (0.85) & 0.86 (0.90) & 0.68 (0.84) & 0.56 (0.75) & 0.67 (0.87) & 0.87 (0.91) & 0.67 (0.82) & 0.57 (0.77) & 0.64 (0.83) & 0.85 (0.89) & 0.70 (0.85) & 0.55 (0.74) \\
    & CardioBERTa.ro & 0.68 (0.86) & 0.86 (0.90) & 0.70 (0.86) & 0.57 (0.77) & \textbf{0.70} (\textbf{0.88}) & \textbf{0.88} (\textbf{0.93}) & 0.70 (\textbf{0.85}) & 0.59 (\textbf{0.79}) & 0.66 (0.84) & 0.84 (0.89) & 0.70 (0.86) & 0.55 (0.74) \\
    & xlm-roberta-large & \textbf{0.71} (\textbf{0.88}) & \textbf{0.88} (\textbf{0.92}) & \textbf{0.74} (\textbf{0.87}) & \textbf{0.61} (\textbf{0.79}) & \textbf{0.70} (\textbf{0.88}) & \textbf{0.88} (0.91) & \textbf{0.72} (0.84) & \textbf{0.61} (\textbf{0.79}) & \textbf{0.72} (\textbf{0.89}) & \textbf{0.88} (\textbf{0.92}) & \textbf{0.76} (\textbf{0.89}) & \textbf{0.61} (\textbf{0.79}) \\
    \hline
    \multirow{3}{*}{Spanish} & roberta-biomedical-clinical-es & 0.72 (0.89) & 0.87 (\textbf{0.92}) & 0.76 (0.88) & 0.69 (0.86) & 0.71 (0.88) & 0.84 (0.88) & 0.74 (0.86) & \textbf{0.71} (\textbf{0.88}) & 0.73 (\textbf{0.90}) & \textbf{0.91} (0.95) & 0.78 (0.91) & 0.67 (0.84) \\
    & CardioBERTa.es & 0.73 (0.89) & 0.84 (0.88) & 0.77 (0.89) & 0.69 (0.86) & 0.73 (0.89) & 0.78 (0.82) & 0.75 (\textbf{0.87}) & 0.68 (0.85) & 0.73 (0.89) & \textbf{0.91} (\textbf{0.96}) & 0.78 (0.90) & \textbf{0.70} (\textbf{0.87}) \\
    & xlm-roberta-large & \textbf{0.74} (\textbf{0.90}) & \textbf{0.88} (0.90) & \textbf{0.79} (\textbf{0.90}) & \textbf{0.70} (\textbf{0.87}) & \textbf{0.74} (\textbf{0.91}) & \textbf{0.88} (\textbf{0.91}) & \textbf{0.77} (\textbf{0.87}) & \textbf{0.71} (\textbf{0.88}) & \textbf{0.74} (\textbf{0.90}) & 0.87 (0.90) & \textbf{0.81} (\textbf{0.92}) & 0.69 (0.86) \\
    \hline
    \multirow{3}{*}{Swedish} & bert-base-swedish-cased & 0.64 (0.83) & 0.86 (0.88) & 0.69 (0.83) & 0.59 (0.78) & 0.66 (0.85) & 0.87 (0.89) & 0.67 (0.81) & 0.61 (0.80) & 0.63 (0.81) & 0.85 (0.88) & 0.70 (0.85) & 0.58 (0.77) \\
     & CardioBERTa.sv & 0.68 (0.87) & \textbf{0.90} (\textbf{0.92}) & 0.70 (\textbf{0.86}) & 0.62 (0.81) & 0.70 (\textbf{0.89}) & \textbf{0.92} (\textbf{0.94}) & 0.69 (\textbf{0.85}) & 0.64 (0.83) & 0.67 (0.85) & \textbf{0.88} (\textbf{0.90}) & 0.71 (0.87) & 0.62 (0.80) \\
     & xlm-roberta-large & \textbf{0.70} (\textbf{0.88}) & 0.86 (0.90) & \textbf{0.72} (\textbf{0.86}) & \textbf{0.66} (\textbf{0.83}) & \textbf{0.71} (0.88) & 0.87 (0.91) & \textbf{0.70} (0.83) & \textbf{0.67} (\textbf{0.84}) & \textbf{0.70} (\textbf{0.88}) & 0.85 (0.89) & \textbf{0.74} (\textbf{0.89}) & \textbf{0.65} (\textbf{0.83}) \\
     \hline
     Multilingual & xlm-roberta-large & 0.72 (0.90) & 0.89 (0.91) & 0.74 (0.88) & 0.66 (0.87) & 0.72 (0.89) & 0.89 (0.91) & 0.72 (0.85) & 0.67 (0.87) & 0.72 (0.90) & 0.89 (0.91) & 0.76 (0.91) & 0.65 (0.86) \\
     \hline
    \end{tabular}
    }
\end{table}
\end{landscape}

\begin{landscape}
\begin{table}
    \centering
    \caption{\textbf{Span-level NER} performance of \textbf{multiclass} models on the CardioCCC held-out test set, averaged across the 10 fold models. A separate model is fine-tuned and evaluated for each entity type; thus, results for the four entity types come from four independently trained models rather than a single multi-entity model. Predictions are decoded using argmax over \textbf{softmax-normalised logits} and aggregated into entity spans. Strict (exact boundary) and relaxed (overlapping boundary) precision, recall, and F1 are reported for each entity type. XLM-RoBERTa-large is trained jointly on all seven languages as four separate entity-specific models and evaluated on the full multilingual test set and separately by language.}
    \label{tab:span_multiclass_per_entity}
    \resizebox{\columnwidth}{!}{%
    \begin{tabular}{l|l|cccc|cccc|cccc}
     \multicolumn{1}{l}{} & \multicolumn{1}{l}{} & \multicolumn{4}{c}{F1 strict (relaxed)} & \multicolumn{4}{c}{Recall strict (relaxed)} & \multicolumn{4}{c}{Precision strict (relaxed)} \\
    \hline
    Language & Model & Dis & Med & Proc & Sym & Dis & Med & Proc & Sym & Dis & Med & Proc & Sym \\
    \hline
\hline
    \multirow{3}{*}{Czech} & robeczech-base & 0.67 (0.85) & 0.86 (0.91) & 0.74 (0.86) & 0.61 (0.79) & 0.68 (0.87) & 0.85 (0.90) & 0.72 (0.83) & 0.61 (0.79) & 0.65 (0.84) & 0.87 (0.92) & 0.76 (0.88) & 0.61 (0.80) \\
    & CardioBERTa.cs & 0.69 (0.86) & 0.85 (\textbf{0.92}) & 0.75 (0.87) & 0.62 (0.80) & 0.72 (\textbf{0.89}) & 0.85 (\textbf{0.92}) & 0.73 (0.85) & 0.65 (0.83) & 0.67 (0.84) & 0.84 (0.91) & 0.77 (0.89) & 0.61 (0.77) \\
    & xlm-roberta-large & \textbf{0.72} (\textbf{0.88}) & \textbf{0.87} (\textbf{0.92}) & \textbf{0.77} (\textbf{0.88}) & \textbf{0.68} (\textbf{0.83}) & \textbf{0.73} (\textbf{0.89}) & \textbf{0.87} (\textbf{0.92}) & \textbf{0.75} (\textbf{0.86}) & \textbf{0.69} (\textbf{0.85}) & \textbf{0.72} (\textbf{0.87}) & \textbf{0.88} (\textbf{0.93}) & \textbf{0.78} (\textbf{0.90}) & \textbf{0.66} (\textbf{0.81}) \\
    \hline
    \multirow{3}{*}{Dutch} & RobBERT2023-base & 0.63 (0.85) & \textbf{0.88} (0.90) & 0.67 (0.82) & 0.62 (0.77) & 0.61 (0.82) & \textbf{0.87} (0.88) & 0.62 (0.76) & 0.63 (0.81) & 0.64 (0.85) & \textbf{0.90} (\textbf{0.92}) & 0.72 (0.88) & 0.60 (0.77) \\
    & CardioBERTa.nl & 0.66 (0.86) & 0.86 (0.90) & 0.68 (0.82) & 0.65 (0.80) & 0.64 (0.83) & 0.85 (0.89) & 0.62 (0.75) & 0.64 (0.79) & 0.68 (\textbf{0.89}) & 0.87 (0.91) & 0.74 (\textbf{0.90}) & 0.65 (0.81) \\
    & xlm-roberta-large & \textbf{0.72} (\textbf{0.88}) & \textbf{0.88} (\textbf{0.91}) & \textbf{0.74} (\textbf{0.88}) & \textbf{0.69} (\textbf{0.84}) & \textbf{0.73} (\textbf{0.89}) & \textbf{0.87} (\textbf{0.90}) & \textbf{0.72} (\textbf{0.86}) & \textbf{0.70} (\textbf{0.85}) & \textbf{0.71} (0.87) & 0.89 (\textbf{0.92}) & \textbf{0.76} (\textbf{0.90}) & \textbf{0.68} (\textbf{0.82}) \\
    \hline
    \multirow{3}{*}{English} & biomed\_roberta\_base & 0.64 (0.81) & \textbf{0.89} (\textbf{0.94}) & 0.67 (0.77) & 0.64 (0.80) & 0.62 (0.78) & 0.88 (0.93) & 0.62 (0.72) & 0.64 (0.81) & 0.68 (0.85) & \textbf{0.90} (\textbf{0.95}) & 0.73 (0.84) & 0.63 (0.79) \\
    & CardioBERTa.en & 0.64 (0.81) & 0.88 (\textbf{0.94}) & 0.67 (0.77) & 0.62 (0.80) & 0.61 (0.78) & 0.88 (\textbf{0.94}) & 0.61 (0.70) & 0.63 (0.80) & 0.66 (0.84) & 0.88 (0.93) & 0.74 (0.85) & 0.62 (0.79) \\
    & xlm-roberta-large & \textbf{0.71} (\textbf{0.84}) & \textbf{0.89} (0.92) & \textbf{0.75} (\textbf{0.86}) & \textbf{0.70} (\textbf{0.84}) & \textbf{0.70} (\textbf{0.83}) & \textbf{0.90} (\textbf{0.94}) & \textbf{0.74} (\textbf{0.84}) & \textbf{0.72} (\textbf{0.86}) & \textbf{0.72} (\textbf{0.86}) & 0.88 (0.91) & \textbf{0.77} (\textbf{0.87}) & \textbf{0.69} (\textbf{0.82}) \\
    \hline
    \multirow{3}{*}{Italian} & bioBIT & 0.63 (0.83) & 0.87 (0.91) & 0.60 (0.82) & 0.60 (0.78) & 0.62 (0.81) & 0.83 (0.87) & 0.55 (0.75) & 0.57 (0.73) & 0.65 (0.85) & \textbf{0.92} (0.96) & 0.65 (\textbf{0.90}) & 0.64 (0.83) \\
    & CardioBERTa.it & 0.64 (0.83) & \textbf{0.88} (\textbf{0.92}) & 0.59 (0.81) & 0.61 (0.79) & 0.61 (0.80) & 0.84 (0.88) & 0.54 (0.74) & 0.59 (0.74) & 0.67 (0.87) & \textbf{0.92} (\textbf{0.97}) & 0.66 (\textbf{0.90}) & 0.65 (\textbf{0.84}) \\
    & xlm-roberta-large & \textbf{0.71} (\textbf{0.88}) & 0.87 (0.90) & \textbf{0.73} (\textbf{0.87}) & \textbf{0.67} (\textbf{0.84}) & \textbf{0.70} (\textbf{0.87}) & \textbf{0.88} (\textbf{0.92}) & \textbf{0.71} (\textbf{0.85}) & \textbf{0.69} (\textbf{0.85}) & \textbf{0.71} (\textbf{0.89}) & 0.85 (0.89) & \textbf{0.75} (\textbf{0.90}) & \textbf{0.66} (0.83) \\
    \hline
    \multirow{3}{*}{Romanian} & xlm-roberta-base & 0.67 (0.86) & 0.87 (0.91) & 0.72 (0.86) & 0.60 (0.76) & 0.70 (0.89) & 0.87 (0.91) & 0.71 (0.85) & 0.60 (0.75) & 0.65 (0.83) & 0.88 (0.92) & 0.73 (0.87) & 0.60 (0.76) \\
    & CardioBERTa.ro & 0.70 (0.87) & \textbf{0.89} (0.92) & 0.73 (0.87) & 0.62 (0.78) & 0.72 (\textbf{0.90}) & \textbf{0.90} (\textbf{0.93}) & 0.73 (\textbf{0.87}) & \textbf{0.64} (\textbf{0.79}) & 0.68 (0.85) & 0.89 (0.92) & 0.73 (0.87) & 0.61 (0.76) \\
    & xlm-roberta-large & \textbf{0.73} (\textbf{0.89}) & \textbf{0.89} (\textbf{0.93}) & \textbf{0.76} (\textbf{0.88}) & \textbf{0.64} (\textbf{0.79}) & \textbf{0.73} (0.89) & 0.88 (0.92) & \textbf{0.74} (0.86) & \textbf{0.64} (0.78) & \textbf{0.73} (\textbf{0.89}) & \textbf{0.90} (\textbf{0.93}) & \textbf{0.77} (\textbf{0.89}) & \textbf{0.65} (\textbf{0.79}) \\
    \hline
    \multirow{3}{*}{Spanish} & biomedical-clinical-es & 0.72 (0.89) & 0.91 (0.94) & 0.79 (0.90) & 0.68 (0.84) & 0.72 (0.88) & 0.90 (0.94) & 0.77 (0.88) & 0.69 (0.84) & 0.73 (\textbf{0.90}) & \textbf{0.91} (\textbf{0.95}) & 0.81 (\textbf{0.92}) & 0.68 (0.84) \\
    & CardioBERTa.es & 0.73 (0.89) & \textbf{0.92} (\textbf{0.95}) & 0.77 (0.89) & 0.67 (0.82) & 0.75 (\textbf{0.92}) & \textbf{0.92} (\textbf{0.96}) & 0.77 (0.89) & 0.68 (0.85) & 0.72 (0.88) & \textbf{0.91} (\textbf{0.95}) & 0.78 (0.90) & 0.66 (0.83) \\
    & xlm-roberta-large & \textbf{0.76} (\textbf{0.91}) & 0.90 (0.93) & \textbf{0.82} (\textbf{0.92}) & \textbf{0.74} (\textbf{0.88}) & \textbf{0.77} (\textbf{0.92}) & \textbf{0.92} (0.94) & \textbf{0.82} (\textbf{0.92}) & \textbf{0.76} (\textbf{0.89}) & \textbf{0.74} (0.89) & 0.89 (0.91) & \textbf{0.82} (\textbf{0.92}) & \textbf{0.73} (\textbf{0.86}) \\
    \hline
    \multirow{3}{*}{Swedish} & bert-base-swedish-cased & 0.65 (0.84) & 0.88 (0.91) & 0.70 (0.85) & 0.57 (0.78) & 0.66 (0.85) & 0.90 (0.93) & 0.69 (0.83) & 0.56 (0.76) & 0.64 (0.83) & 0.87 (0.90) & 0.73 (0.88) & 0.58 (0.80) \\
    & CardioBERTa.sv & 0.69 (0.87) & \textbf{0.91} (\textbf{0.92}) & 0.73 (0.87) & 0.60 (0.81) & 0.70 (\textbf{0.89}) & \textbf{0.93} (\textbf{0.95}) & 0.73 (\textbf{0.87}) & 0.59 (0.79) & 0.68 (0.86) & \textbf{0.88} (0.90) & 0.73 (0.88) & 0.62 (\textbf{0.82}) \\
    & xlm-roberta-large & \textbf{0.72} (\textbf{0.88}) & 0.88 (\textbf{0.92}) & \textbf{0.75} (\textbf{0.89}) & \textbf{0.70} (\textbf{0.84}) & \textbf{0.72} (\textbf{0.89}) & 0.89 (0.93) & \textbf{0.74} (\textbf{0.87}) & \textbf{0.71} (\textbf{0.86}) & \textbf{0.71} (\textbf{0.88}) & 0.87 (\textbf{0.91}) & \textbf{0.77} (\textbf{0.90}) & \textbf{0.68} (\textbf{0.82}) \\
    \hline
    Multilingual & xlm-roberta-large & 0.73 (0.90) & 0.90 (0.92) & 0.77 (0.90) & 0.70 (0.87) & 0.73 (0.90) & 0.90 (0.93) & 0.76 (0.88) & 0.71 (0.89) & 0.73 (0.90) & 0.90 (0.92) & 0.78 (0.91) & 0.69 (0.86) \\
    \hline
    \end{tabular}
    }
\end{table}
\end{landscape}

\paragraph{Multiclass systems}
In the multiclass setting, each entity type is handled by a separately fine-tuned model, and the systems are evaluated per entity type only. Table~\ref{tab:span_multiclass_per_entity} reports their span-level performance. The difficulty ordering largely matches the multilabel setting, with medication easiest and symptom generally hardest, the only exception being Italian, where the monolingual models score procedure at or below symptom.

The effect of cardiology-specific adaptation is comparable to the multilabel setting, with strict-F1 changes ranging from $-0.02$ to $+0.04$. Declines are smaller than in the multilabel setting (at most $-0.02$ versus $-0.04$) but more frequent. Only Swedish and Romanian show consistent gains, with all four entity types improving (by 0.03--0.04 and 0.01--0.03, respectively). Czech, Italian, Dutch, and Spanish show mixed effects, with gains on some entity types and declines on others; the largest swings occur in Dutch, where disease and symptom improve by 0.03 but medication declines by 0.02. English shows no benefit, with no entity type improving and a slight decline for symptom. As in the multilabel setting, adaptation does not reliably outperform the general-domain base models.

The benefit of multilingual training is not uniform across entity types. For medication, the monolingual models remain competitive: the stronger of the two matches XLM-RoBERTa-large in three languages (Dutch, English, and Romanian) and outperforms it in three others (Italian 0.88 versus 0.87, Spanish 0.92 versus 0.90, Swedish 0.91 versus 0.88), trailing only in Czech (0.86 versus 0.87). For symptom, by contrast, the multilingual model leads in every language, by margins ranging from 0.02 in Romanian to 0.10 in Swedish. Disease and procedure generally fall between these extremes, although the largest gap occurs for procedure in Italian, where the two monolingual models score 0.13 and 0.14 strict F1 points, respectively, below XLM-RoBERTa-large.

The same asymmetry between strict and relaxed scores is observed as in the multilabel setting, and with comparable magnitudes: medication mentions are delimited almost exactly (0.01--0.07 F1 points difference), whereas disease, procedure, and symptom show differences of 0.10--0.22 F1 points. This pattern is unchanged across the two label formulations, indicating that models often identify the correct entity type and an overlapping span but fail to delimit the mention exactly.

\paragraph{Comparison of multilabel and multiclass settings}
Comparing the two label formulations, the multiclass systems outperform their \mbox{multilabel} counterparts in most language--entity combinations (64 of 88) by a small margin that is typically 0.01--0.03 strict F1 and averages about 0.02 overall. The magnitude of this advantage varies across entity types: gains reach at most 0.02 strict F1 for disease, 0.04 for procedure, and 0.06 for symptom, the hardest category. The largest symptom gains include English biomed\_roberta\_base (0.58 to 0.64), Dutch RobBERT2023-base (0.57 to 0.62), and Romanian CardioBERTa.ro (0.57 to 0.62). Medication gains are mostly modest, but they include the single largest improvement in the comparison: Spanish CardioBERTa.es rises from 0.84 to 0.92, reflecting the recovery of an adapted model that had underperformed in the multilabel setting. 

Reversals, where the multilabel formulation is ahead, occur in 14 of the 88 combinations and are small (at most 0.04 strict F1). All involve monolingual models; XLM\mbox{-}RoBERTa\mbox{-}large never performs worse in the multiclass setting. Half are for symptom, affecting all monolingual models in Czech, Swedish, and Spanish, as well as the bioBIT model in Italian (each by 0.01--0.02). Italian accounts for most of the remainder, with both monolingual models trailing on disease (by 0.01) and procedure (by 0.02 and 0.04). The last three reversals are for procedure (both English monolingual models) and disease (Czech CardioBERTa.cs). Medication shows no reversals in any language.

The multiclass advantage comes at a substantial computational and deployment cost, since the multiclass configuration requires four independently trained models instead of one, and thus approximately four times the parameters and inference cost, for an average gain of only 0.02 strict F1. Whether this trade-off is worthwhile therefore depends on deployment constraints rather than on accuracy alone. 

We release a publicly available NER demo\footnote{\url{https://huggingface.co/spaces/DT4H-IE/DT4H-NER-Demo}} that annotates cardiology text in any of the seven languages. It is built on the strongest overall configuration in our experiments: the multilingual XLM\mbox{-}RoBERTa\mbox{-}large trained in the multiclass setting, with one model per entity type. Because XLM\mbox{-}RoBERTa\mbox{-}large is multilingual, four entity-specific models cover all seven languages, which keeps deployment practical despite the multiclass overhead discussed above.

\subsection{External clinical validation}
\begin{figure}[t]
  \centering
    \includegraphics[width=\linewidth]{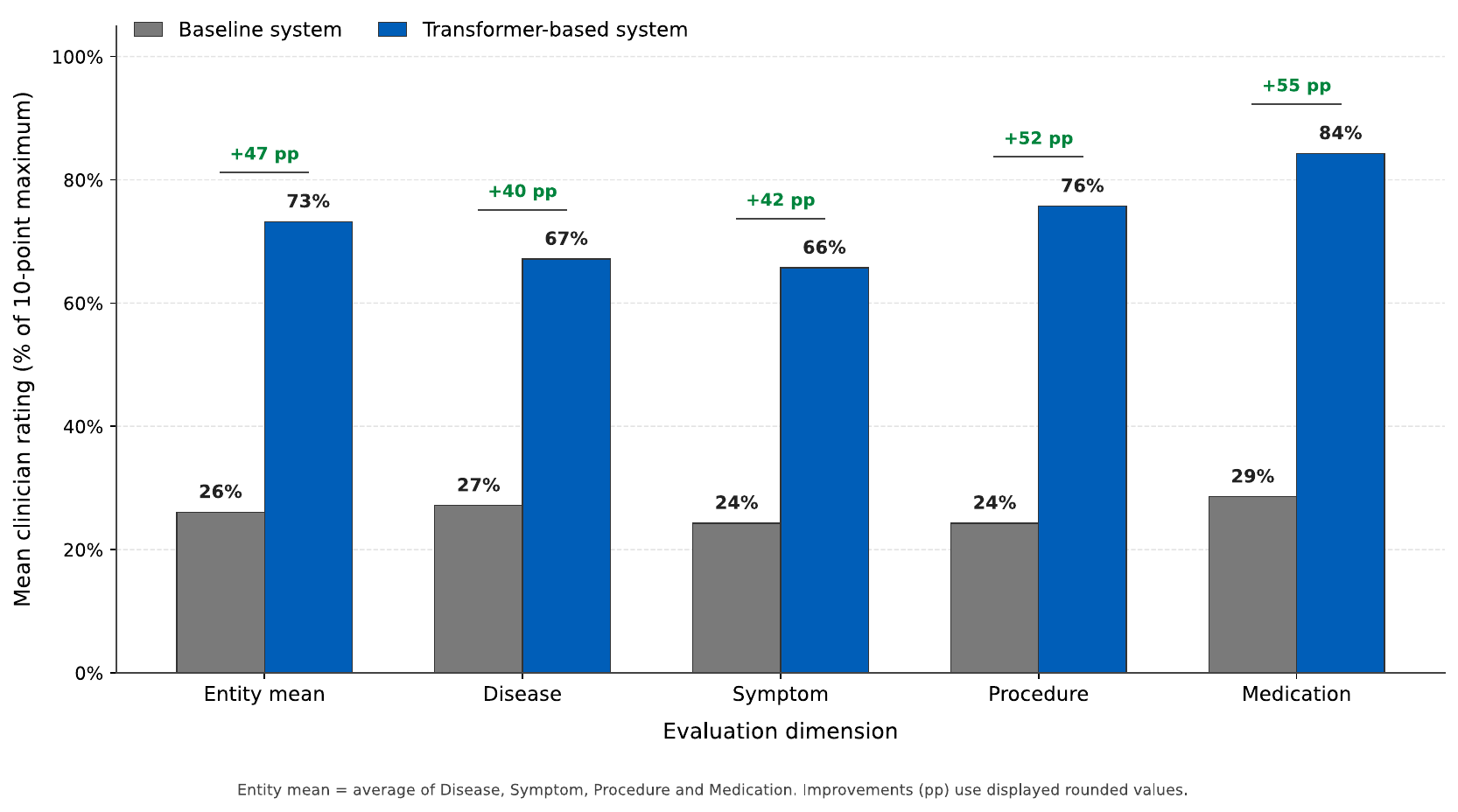}
    \caption{Mean clinician ratings for the baseline and Transformer-based NER systems, reported for each entity type (disease, symptom, procedure, and medication) and averaged across entity types.}
    \label{fig:clinical_evaluation}
\end{figure}

Clinical evaluators in all seven languages rated the outputs of a rule- and dictionary\mbox{-}based baseline and the Transformer-based NER system developed in this work (Section~\ref{sub:ext_clin_validation}). Figure~\ref{fig:clinical_evaluation} summarizes the mean clinician ratings for the two approaches. The Transformer-based system received higher ratings than the baseline both on average and for each entity category. The mean rating across entity types increased from 26\% for the baseline to 73\% for the Transformer-based system, an absolute improvement of 47 percentage points. Improvements were observed consistently across all four categories: diseases (27\% to 67\%), symptoms (24\% to 66\%), procedures (24\% to 76\%), and medications (29\% to 84\%). 

The largest absolute improvement was obtained for medications (+55 percentage points), followed by procedures (+52 percentage points). These categories also received the highest final ratings, suggesting that the system is particularly effective for entities with comparatively stable terminology and clearly delimited textual spans. Disease and symptom extraction obtained lower ratings, although both categories still showed substantial improvements over the baseline. This ranking mirrors the automatic NER evaluation, in which medication was recognised most reliably and symptom least, and is consistent with the greater semantic variability, contextual dependence, and boundary ambiguity typically associated with disease and symptom mentions in clinical narratives. 

\begin{figure}[t]
  \centering
    \includegraphics[width=\linewidth]{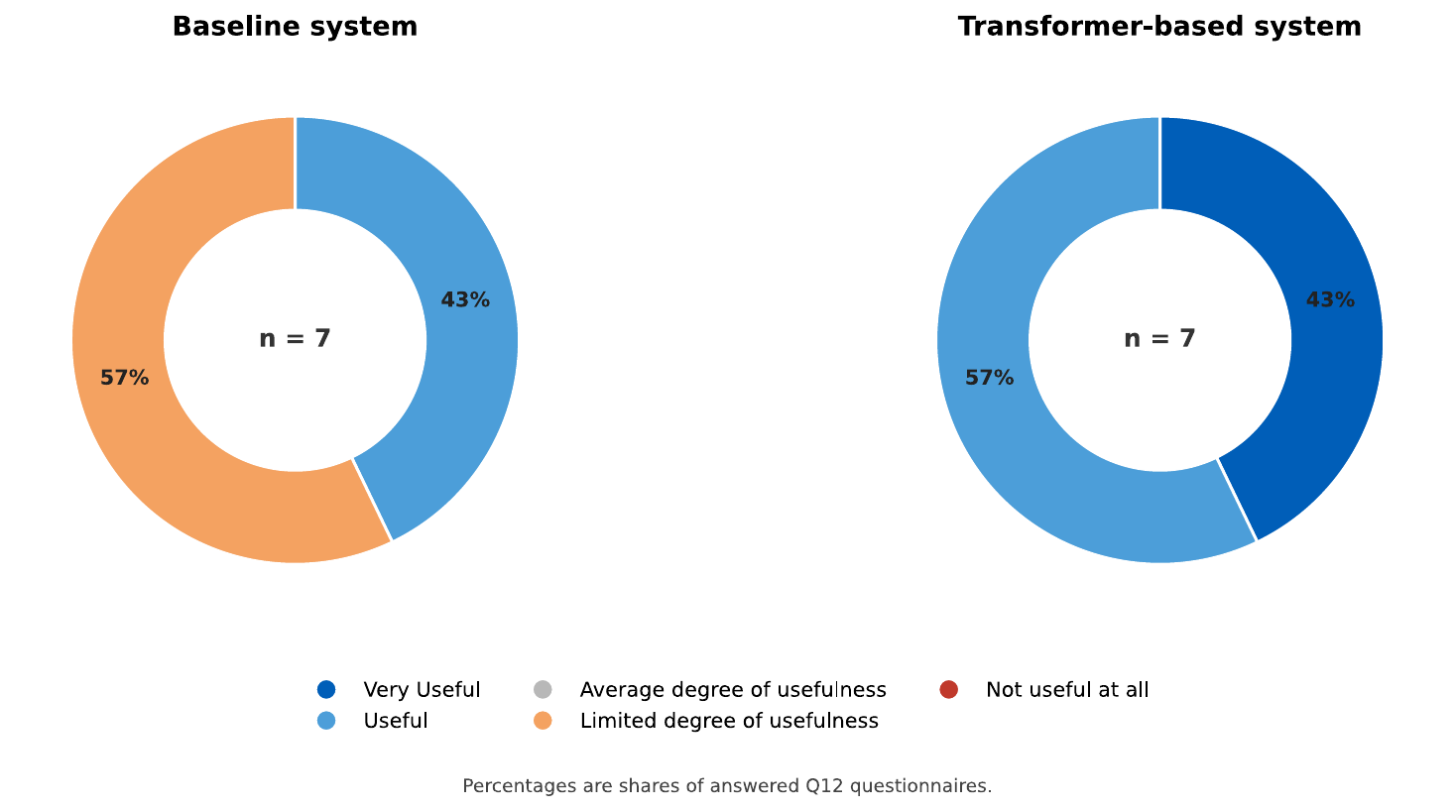}
    \caption{Clinician-reported usefulness of baseline and Transformer-based clinical
    entity extraction systems for potential hospital-oriented implementation.}
    \label{fig:clinical_usefulness}
\end{figure}

The perceived usefulness assessment showed a similarly clear separation between the two systems (see Figure~\ref{fig:clinical_usefulness}). For the baseline, 57\% of the evaluations (4/7) rated the output as having only a limited degree of usefulness, while 43\% (3/7) considered it useful. In contrast, all evaluations rated the Transformer-based system as at least useful: 57\% (4/7) rated it as useful and 43\% (3/7) as very useful. These results suggest that the improvements observed in the automatic evaluation extend beyond it and translate into output that clinicians perceive as useful, supporting the potential integration of the system into hospital-based NLP workflows.

Finally, the questionnaire also collected qualitative feedback on remaining error patterns. The main limitations identified by clinicians concerned semantic disambiguation, entity-boundary precision, and contextual interpretation. In particular, evaluators reported confusion between semantically related categories, especially diseases and symptoms; partial or over-extended entity spans; missed mentions; and the extraction of negated or context-dependent mentions as positive entities. These observations are consistent with the automatic evaluation, in which boundary mismatches accounted for a large share of errors and disease and symptom were the most difficult categories. Since the current annotation scheme does not encode negation or other contextual attributes, the latter error type reflects the scope of the task rather than a model failure. Rather than being treated solely as limitations, these observations will be incorporated into an iterative data and model-development cycle, informing future refinements of both the annotations and the models.

\section{Discussion}
This study aimed to address the two gaps identified in Section~\ref{subsec:gaps}: the lack of cardiology-specific encoders for European languages, and the absence of a benchmark that is simultaneously cardiology-specific, multilingual, and broad in entity coverage. The seven-language extension of CardioCCC closes the second gap and provides the means to evaluate empirically whether the cardiology-specific encoders developed in this work actually improve downstream performance. The results are informative precisely because they do not confirm the expected ordering: cardiology-specific pretraining substantially improves the encoders’ intrinsic language modelling performance, yet yields only small and uneven downstream gains, and is generally surpassed by joint multilingual fine-tuning.

Across both label formulations, cardiology-specific continued pretraining changes downstream recognition by $-0.04$ to $+0.04$ strict F1 relative to the corresponding base models. Medication shows the least benefit and the largest declines: in the \mbox{multilabel} setting, adaptation reduces performance in Dutch (0.87 to 0.83), Spanish (0.87 to 0.84), and Czech (0.86 to 0.84), improves it in Swedish (0.86 to 0.90), Italian (0.85 to 0.87), and English (0.87 to 0.88), and leaves it unchanged in Romanian. A plausible explanation is that drug names are lexically distinctive and largely standardised, and are therefore already well covered by general-domain pretraining; further specialisation offers little to gain and some capacity to lose. The gains also vary across languages: they are clearest for Swedish and Romanian, where adaptation never reduces performance, modest for Dutch, Czech, and Italian, where small net gains are offset by occasional declines, and negligible for English and Spanish, where small gains and losses largely cancel out. This suggests that the value of domain-adaptive pretraining depends on how much in-domain knowledge the initial model already has. The initialisation \mbox{backbones} in Table~\ref{tab:base-models-overview} are consistent with this: the languages that benefit most (Swedish and Romanian) started from general-domain encoders, whereas English, Spanish, and Italian started from models already adapted to biomedical text and gain least. Under this interpretation, cardiology-specific continued pretraining is best understood not as a uniformly beneficial step but as a way of compensating for weak in-domain coverage in the initial model, which is precisely the situation of the clinically under-resourced languages targeted in this work.

These limited downstream gains contrast with the improvements in pseudo\mbox{-}perplexity, which are substantial for most adapted models (Section~\ref{ppl_results}). Intrinsic language-model quality is therefore a weak predictor of downstream extraction performance in this setting, and pseudo-perplexity alone should not be used to decide whether domain-adaptive pretraining is worthwhile for a given language.

Whereas cardiology-specific adaptation yields only limited gains, joint multilingual fine-tuning proves considerably more effective: XLM\mbox{-}RoBERTa\mbox{-}large model fine-tuned jointly on all seven languages achieves the highest strict F1 score for almost every language--entity combination, exceeding the best monolingual model by up to 0.13 strict F1 points. This is true even against the cardiology-adapted monolingual models, which were pretrained on in-domain text in the target language. For multi-country deployment, this challenges the assumption that language-specific models are preferable: a single multilingual system is not only simpler to maintain, but, in most settings, also more accurate. Medication is the main exception, with monolingual models matching or outperforming the multilingual model in several languages, suggesting that the benefit of cross-lingual transfer is smallest for entity types whose surface forms are already lexically distinctive. The comparison is not, however, controlled for model capacity or data: XLM\mbox{-}RoBERTa\mbox{-}large is substantially larger than most of the \mbox{monolingual} models and is fine-tuned on seven times more annotated data, so its advantage cannot be attributed to multilinguality alone.

Across all models, a large share of errors concerns boundary placement. For disease, procedure and symptom, relaxed scores exceed strict ones by 0.10--0.23 F1, whereas for medication the difference is only 0.01--0.07. The models therefore locate mentions considerably more often than they delimit them exactly, and this pattern is essentially unchanged across label formulations, indicating that it stems from the entity types themselves rather than from how the classification task is formulated. Clinicians independently reported partial and over-extended spans among the main errors. The practical implication depends on the application: for indexing, retrieval or cohort identification, where approximate localisation suffices, the systems are appreciably more usable than their strict scores suggest, whereas for applications requiring exact spans, boundary handling is where further effort would be most productive.

Finally, the choice between label formulations involves a clear trade\mbox{-}off between accuracy and cost. The multiclass formulation improves over the multilabel one by only 0.02 strict F1 on average, yet requires four entity-specific models instead of one, with correspondingly higher parameter count and inference cost. Where inference cost or maintenance is a constraint, the multilabel formulation offers nearly the same performance from a single model.

\section{Limitations}
\label{sec:limitations}

The training and evaluation data differ from routine clinical text in two respects. First, CardioCCC consists of published clinical case reports, which are edited for publication and therefore lack the telegraphic phrasing, template structure, dense abbreviations, and typographical noise of everyday clinical documentation. Reported scores should thus be read as an upper bound for real electronic health records. Second, only the Spanish corpus is natively authored. The other six language versions were produced by machine translation and annotation projection, followed by clinician review. Keeping content identical across languages makes the cross-lingual comparison clean, but it also means the non-Spanish versions inherit Spanish documentation conventions and may retain translation artefacts. Non-Spanish scores may therefore not transfer to natively written clinical text. 

Several comparisons are only partially controlled. First, the multilingual system uses a considerably larger backbone than any monolingual model and, because the corpora are parallel, sees each training document in seven languages. Its advantage may therefore reflect the added capacity provided by its larger size and its repeated exposure to the same documents rather than cross-lingual transfer as such. Second, the amount of continued pretraining data differed across languages, and the Dutch model received an additional adaptation stage. Differences in how much each language benefits from adaptation may therefore reflect not only properties of the languages but also unequal training effort. Third, we report the average over ten fold models evaluated on a single fixed held-out test set, without dispersion or significance testing. Differences of the magnitude we discuss, such as the 0.02 multiclass advantage, could therefore arise from random fold-level variation. Inter-annotator agreement is likewise not reported, leaving the annotation ceiling unknown. This matters most for entity boundaries, which account for a large share of disease, procedure, and symptom errors.

Further limitations stem from how the task is defined and how predictions are decoded. The annotation scheme does not encode assertion status, so negated, hypothetical, and family-attributed mentions are labelled and extracted in the same way as mentions that describe the patient’s current condition. Downstream use would therefore require an additional assertion-detection step. In addition, roughly 4\% of annotated mentions carry multiple labels, mostly disease--symptom; because our decoding assigns a single label to each token, these overlapping annotations cannot be fully recovered. Moreover, normalisation to controlled terminologies, which cross-site interoperability would ultimately require, is also not addressed in this paper.

Regarding decoding, all span-level results were obtained with a single protocol in which fixed confidence thresholds are applied to softmax-normalised scores. However, softmax is not the scale on which the multilabel models were trained. Recomputing confidences on their native sigmoid scale (Appendix~\ref{app:appendix_results}) leaves F1 and model rankings essentially unchanged but shifts the balance between precision and recall. F1 is thus a more reliable basis for comparison than precision or recall individually. Finally, post\mbox{-}processing relies on hand-crafted, language- and domain-specific rules that may not generalise to other domains or languages and can introduce systematic errors, such as filtering out valid mentions or leaving boundary inconsistencies uncorrected.

A further limitation concerns the medical-span pseudo-perplexity analysis. Medical spans were identified with the multilingual XLM\mbox{-}RoBERTa\mbox{-}large NER model, so mentions it failed to detect were excluded from the evaluation. Because this model was fine-tuned on cardiology annotations, the selected spans may be skewed towards terminology characteristic of this domain, potentially introducing a bias in favour of the cardiology-adapted models. However, the gains of the adapted models on medical spans are not consistently larger than across all spans, suggesting the effect is limited. In addition, the word-level scoring used by minicons mitigates but does not remove tokenizer--granularity effects, so absolute pseudo-perplexity values are not directly comparable across languages or tokenizers. We therefore focus on comparing each base model with its adapted version.

Finally, the clinical validation and the overall scope are limited. The clinician assessment is descriptive: it comprises one unblinded questionnaire per language, \mbox{without} inter-rater reliability assessment, and therefore indicates the direction but not the size of the difference between systems. Regarding scope, all seven languages are European and only one specialty was evaluated, so transfer to typologically distant languages or other clinical domains remains untested. We also report no generative baseline: encoder-based models were chosen for deployment cost, span fidelity, and data governance, and we make no claim of an accuracy advantage over generative approaches.

\section{Future work}

The limitations above point to four main directions for follow-up work: controlled experiments that would sharpen the interpretation of our results, improved modelling of entity boundaries, extensions of the task beyond span detection, and evaluation closer to deployment.

The main open question is why the multilingual XLM\mbox{-}RoBERTa\mbox{-}large models, trained jointly on all seven languages, generally outperform the language-specific models, since they differ from them in both model size and the amount of training data. The direct control is to fine-tune XLM\mbox{-}RoBERTa\mbox{-}large separately on each language: if the advantage of joint training persists, it reflects cross-lingual transfer; if it disappears, the gains reflect the added capacity provided by the larger model size, and the recommendation becomes to scale the backbone rather than train jointly. Two further experiments would extend this analysis: adapting XLM\mbox{-}RoBERTa\mbox{-}large to cardiology text, to test whether domain adaptation pays off at a larger scale; and varying the number of pooled languages, to distinguish genuine cross-lingual transfer from repeated exposure to parallel content. These comparisons should report fold-level dispersion and significance tests, which the present results lack.

Since boundary placement accounts for a large share of the remaining errors, approaches that model span extent directly, such as span-based classifiers, boundary\mbox{-}aware training objectives, or structured decoding, are a natural next step. The decoding procedure also warrants further investigation: replacing single-label argmax with a genuinely multilabel decoding rule would recover the overlapping disease--symptom annotations that the current protocol discards and, together with tuning per-type thresholds, would show whether the multiclass advantage persists. Measuring inter-annotator agreement on entity boundaries would indicate how much of the boundary error is recoverable at all.

For clinical use, span extraction alone is not sufficient and must be extended in two respects. First, mentions need to be qualified by assertion status (negated, hypothetical, historical, or attributed to a relative rather than the patient), since clinicians flagged these as a source of misleading output. This requires adding assertion attributes to the CardioCCC schema and reporting performance conditioned on them. Second, extracted mentions should be linked to terminologies such as SNOMED CT~\cite{snomed2026}, ICD\mbox{-}11~\cite{who2019icd11}, and the UMLS~\cite{bodenreider2004unified}, a step at which boundary errors matter most, since a partially delimited mention may map to the wrong concept.

Finally, this work should be extended both towards real clinical settings and towards additional specialties, languages, and model families. Because CardioCCC consists of published case reports, with all language versions except Spanish produced by machine translation, the most informative test would be evaluation on natively written hospital documentation, where even small test sets per language would suffice to disentangle translation artefacts from the actual ability of models to process each language. Since such records cannot be centralised, this evaluation must take place within the infrastructure of the participating hospitals, where deployment costs can also be measured. Extending CardioLM to further specialties and typologically distant languages would test whether our findings generalise beyond cardiology and European languages. In addition, our choice of encoder-based models rests on arguments of deployment cost, span fidelity, and data governance rather than on measured accuracy; a systematic comparison against prompted and fine-tuned generative models in terms of extraction quality, exact-span reliability, and on-premise cost would replace these arguments with evidence.

\section{Conclusions}
We presented CardioLM, a multilingual suite of small language models for cardiology covering seven European languages. It comprises CardioBERTa, a set of language\mbox{-}specific encoders obtained by continued pretraining on cardiology and biomedical text, and CardioNER, a set of monolingual and multilingual models fine\mbox{-}tuned to recognise diseases, medications, procedures, and symptoms. Alongside the models, we released a seven\mbox{-}language extension of the CardioCCC corpus with more than 400,000 clinician\mbox{-}validated annotations. It is the first benchmark that is simultaneously cardiology\mbox{-}specific, multilingual, and broad in entity coverage, and the first annotated cardiology resource for Czech, Romanian, and Swedish.

Our results do not support the assumption that specialty\mbox{-}specific pretraining is the decisive factor for downstream entity recognition. Continued pretraining substantially reduces pseudo-perplexity in most languages, yet changes strict span-level F1 by at most 0.04 in either direction, with the clearest gains for languages whose initial models lacked biomedical adaptation. It is therefore best understood as a way of compensating for weak in-domain coverage rather than as a uniformly beneficial step. In contrast, XLM\mbox{-}RoBERTa\mbox{-}large fine\mbox{-}tuned jointly on all seven languages achieves the best strict F1 for nearly every language and entity type, exceeding the strongest monolingual model by up to 0.13. For multi-country deployment, this suggests that a single multilingual system can be both easier to maintain and more accurate than seven language-specific ones.

Three further findings have direct practical implications. First, the models usually locate mentions correctly but often misplace their exact boundaries, particularly for disease, procedure, and symptom, while boundary errors are much less frequent for medication. The models are thus already useful for tasks that tolerate approximate spans, such as retrieval or cohort identification, whereas tasks that depend on exact spans, such as entity linking, will first need more precise delimitation of mentions. Second, the multiclass formulation, which trains a separate model for each entity type, improves strict F1 by only 0.02 on average over the multilabel formulation at roughly four times the training and inference cost, so the choice between them should depend on deployment constraints. Third, clinicians in all seven languages rated the Transformer-based system well above a rule- and dictionary-based baseline for every entity type and judged its output useful or very useful, indicating that the measured improvements are noticeable in practice.

These conclusions are limited by the training and evaluation data: CardioCCC consists of published case reports rather than routine hospital documentation, and only its Spanish version is natively written. Validation on native clinical records, together with assertion detection and normalisation to controlled terminologies, remains necessary before the systems can support clinical data structuring. All models, code, and data are openly released, together with a public NER demo, to support this work and to lower the barrier to clinical NLP in under-resourced European languages.

\section*{Declarations}

\subsection*{Ethics approval and consent to participate}
The \ac{umcu} quality assurance research officer confirmed under project number 22U-0292 that this study does not fall under the scope of the Dutch Medical Research Involving Human Subjects Act (WMO) and therefore does not require approval from an accredited medical ethics committee. The study was performed compliant with local legislation and regulations. All patient data were deidentified in compliance with the European Union General Data Protection Regulation, and as a result, written informed consent was not required by the \ac{umcu} ethical committee.

\subsection*{Availability of data and materials}
The seven-language extension of the CardioCCC corpus developed in this study, including its clinician-validated annotations, is publicly available at \mbox{\url{https://doi.org/10.5281/zenodo.23060840}}. The ParaCliTe synthetic \mbox{clinical} records used in the clinician-centred validation can be accessed at \mbox{\url{https://zenodo.org/records/22641301}}, while the ELCardioCC discharge letters are distributed by their original providers under their respective terms of use~\cite{dimitriadis2025overview}. The questionnaires, evaluator instructions, and system outputs from the clinical validation are available at \url{https://doi.org/10.5281/zenodo.20394665}.

All CardioLM models are publicly released: the CardioBERTa encoders at \mbox{\url{https://huggingface.co/collections/DT4H/cardioberta-family}} and the CardioNER models at \url{https://huggingface.co/collections/DT4H/cardioner}. The code used for pretraining, fine-tuning, inference, and evaluation is publicly available at \mbox{\url{https://github.com/DataTools4Heart/CardioLM}}, and an interactive demo of the NER system can be found at \url{https://huggingface.co/spaces/DT4H-IE/DT4H-NER-Demo}.

All resources released with this work are licensed under the Creative Commons Attribution 4.0 International (CC BY 4.0) licence, except for the Czech models, which inherit the Creative Commons Attribution-NonCommercial-ShareAlike 4.0 International (CC BY-NC-SA 4.0) licence of their base model, RobeCzech~\cite{straka2021robeczech}.

\subsection*{Funding}
The work received funding from the European Union's Horizon Europe research and innovation programme under Grant Agreement No. 101057849 (DataTools4Heart project) and No. 101080430 (AI4HF project). Fernando Gallego-Donoso and \mbox{Salvador} Lima-López received a fellowship within the “Generación D” initiative, Red.es, Ministerio para la Transformación Digital y de la Función Pública, for talent attraction (C005/24-ED CV1). Funded by the European Union NextGenerationEU funds, through PRTR.

\subsection*{Acknowledgements}

We thank Google for providing access to the Google TPU research cloud as part of the pretraining.

\bibliography{bibliography}

\clearpage

\appendix
\counterwithin{table}{section}

\AtBeginEnvironment{table}{\small}

\section{Additional results}
\label{app:appendix_results}

\begin{table}[ht]
\centering
\caption{\textbf{Token-level NER} performance of \textbf{multilabel} models on the CardioCCC held-out test set, averaged across the 10 fold models. Averages are computed over the eight B-/I- labels, with the O class excluded from both micro- and macro-averages. Labels are decoded independently using \textbf{sigmoid} ($>0.5$), allowing a token to receive multiple labels or none. XLM-RoBERTa-large is trained jointly on all seven languages and evaluated on the full multilingual test set and separately by language.}
\label{tab:ner-token-level}
\resizebox{0.93\textwidth}{!}{
\begin{tabular}{l|l|c|c|c}
\multirow{2}{*}{Language}
& \multirow{2}{*}{Model}
& F1
& Recall
& Precision \\
& & macro (micro) & macro (micro) & macro (micro) \\
\hline
\hline

\multirow{3}{*}{Czech}
& robeczech-base
& 0.84 (0.83)
& 0.83 (0.82)
& 0.85 (0.83) \\

& CardioBERTa.cs
& \textbf{0.85} (\textbf{0.84})
& 0.85 (0.84)
& \textbf{0.86} (\textbf{0.84}) \\

& xlm-roberta-large
& \textbf{0.85} (\textbf{0.84})
& \textbf{0.86} (\textbf{0.85})
& 0.85 (\textbf{0.84}) \\
\hline

\multirow{3}{*}{Dutch}
& RobBERT2023-base
& 0.80 (0.79)
& 0.81 (0.80)
& 0.80 (0.79) \\

& CardioBERTa.nl
& 0.79 (0.77)
& 0.78 (0.76)
& 0.80 (0.77) \\

& xlm-roberta-large
& \textbf{0.85} (\textbf{0.85})
& \textbf{0.86} (\textbf{0.86})
& \textbf{0.84} (\textbf{0.84}) \\
\hline

\multirow{3}{*}{English}
& biomed\_roberta\_base
& 0.82 (0.79)
& 0.84 (0.82)
& 0.81 (0.77)\\

& CardioBERTa.en
& 0.83 (0.80)
& 0.84 (0.83)
& 0.81 (0.77)\\

& xlm-roberta-large
& \textbf{0.85} (\textbf{0.83})
& \textbf{0.86} (\textbf{0.84})
& \textbf{0.83} (\textbf{0.81}) \\
\hline

\multirow{3}{*}{Italian}
& bioBIT
& 0.85 (0.84)
& 0.86 (0.84)
& 0.84 (0.83) \\

& CardioBERTa.it
& \textbf{0.86} (0.84)
& 0.87 (0.85)
& \textbf{0.85} (0.84) \\

& xlm-roberta-large
& \textbf{0.86} (\textbf{0.85})
& \textbf{0.88} (\textbf{0.86})
& \textbf{0.85} (\textbf{0.85}) \\
\hline

\multirow{3}{*}{Romanian}
& xlm-roberta-base
& 0.83 (0.81)
& 0.83 (0.81)
& 0.82 (0.81) \\

& CardioBERTa.ro
& 0.84 (0.82)
& 0.85 (0.83)
& 0.83 (0.82) \\

& xlm-roberta-large
& \textbf{0.85} (\textbf{0.83})
& \textbf{0.86} (\textbf{0.84})
& \textbf{0.84} (\textbf{0.83}) \\
\hline

\multirow{3}{*}{Spanish}
& roberta-biomedical-clinical-es
& 0.87 (0.84)
& 0.87 (0.85)
& 0.86 (0.83) \\

& CardioBERTa.es
& 0.87 (0.84)
& 0.87 (0.85)
& 0.86 (0.83) \\

& xlm-roberta-large
& \textbf{0.90} (\textbf{0.88})
& \textbf{0.92} (\textbf{0.90})
& \textbf{0.88} (\textbf{0.87}) \\
\hline

\multirow{3}{*}{Swedish}
& bert-base-swedish-cased
& 0.81 (0.80)
& 0.81 (0.80)
& 0.81 (0.80) \\

& CardioBERTa.sv
& 0.84 (0.83)
& 0.86 (0.84)
& 0.83 (0.82) \\

& xlm-roberta-large
& \textbf{0.86} (\textbf{0.85})
& \textbf{0.87} (\textbf{0.85})
& \textbf{0.85} (\textbf{0.84}) \\
\hline
Multilingual
& xlm-roberta-large
& 0.86 (0.85)
& 0.87 (0.86)
& 0.85 (0.84) \\
\hline
\end{tabular}
}
\end{table}

\begin{table}[]
\centering
\caption{\textbf{Span-level NER} performance of \textbf{multilabel} models on the CardioCCC held-out test set, averaged across the 10 fold models. Predictions are decoded using \textbf{per-label sigmoid confidence scores} for thresholding, matching the training objective, and then aggregated into entity spans. Because argmax label selection is invariant to sigmoid versus softmax scaling, each token receives a single label. Micro- and macro-averages are computed over the four entity types, with strict (exact boundary) and relaxed (overlapping boundary) precision, recall, and F1 reported. XLM\mbox{-}RoBERTa\mbox{-}large is trained jointly on all seven languages and evaluated on the full multilingual test set and separately by language.}

\resizebox{\textwidth}{!}{
\begin{tabular}{l|l|ccc|ccc}
\multicolumn{2}{c}{} & \multicolumn{3}{c}{Micro} & \multicolumn{3}{c}{Macro} \\
\hhline{--|---|---}
\multirow{2}{*}{Language} & \multirow{2}{*}{Model}
& F1 & Recall & Precision
& F1 & Recall & Precision \\
& & strict (relaxed) & strict (relaxed) & strict (relaxed)
  & strict (relaxed) & strict (relaxed) & strict (relaxed) \\
\hline
\hline
\multirow{3}{*}{Czech}
& robeczech-base
& 0.68 (0.83) & 0.66 (0.80) & 0.71 (0.87)
& 0.72 (0.85) & 0.69 (0.81) & 0.75 (0.88) \\

& CardioBERTa.cs
& 0.70 (0.84) & 0.67 (0.81) & \textbf{0.73} (\textbf{0.88})
& 0.73 (0.85) & 0.70 (0.82) & \textbf{0.76} (\textbf{0.89}) \\

& xlm-roberta-large
& \textbf{0.71} (\textbf{0.86}) & \textbf{0.70} (\textbf{0.84}) & 0.72 (0.87)
& \textbf{0.74} (\textbf{0.87}) & \textbf{0.73} (\textbf{0.85}) & 0.75 (0.88) \\
\hline

\multirow{3}{*}{Dutch}
& RobBERT2023-base
& 0.64 (0.81) & 0.63 (0.80) & 0.65 (0.82)
& 0.68 (0.83) & 0.67 (0.81) & 0.70 (0.84) \\

& CardioBERTa.nl
& 0.65 (0.80) & 0.60 (0.74) & 0.71 (\textbf{0.88})
& 0.68 (0.81) & 0.63 (0.75) & 0.74 (\textbf{0.88}) \\

& xlm-roberta-large
& \textbf{0.71} (\textbf{0.86}) & \textbf{0.70} (\textbf{0.84}) & \textbf{0.72} (0.87)
& \textbf{0.74} (\textbf{0.86}) & \textbf{0.73} (\textbf{0.85}) & \textbf{0.75} (\textbf{0.88}) \\
\hline

\multirow{3}{*}{English}
& biomed\_roberta\_base
& 0.65 (0.80) & 0.63 (0.78) & 0.67 (0.83) & 0.69 (0.82) & 0.67 (0.80) & 0.71 (0.85) \\

& CardioBERTa.en
& 0.65 (0.80) & 0.62 (0.76) & 0.69 (0.84) & 0.70 (0.82) & 0.67 (0.79) & 0.73 (0.86) \\

& xlm-roberta-large
& \textbf{0.71} (\textbf{0.85}) & \textbf{0.71} (\textbf{0.83}) & \textbf{0.73} (\textbf{0.86})
& \textbf{0.75} (\textbf{0.86}) & \textbf{0.74} (\textbf{0.85}) & \textbf{0.76} (\textbf{0.87}) \\
\hline

\multirow{3}{*}{Italian}
& bioBIT
& 0.64 (0.85) & 0.63 (0.82) & 0.66 (0.87)
& 0.68 (0.85) & 0.67 (0.84) & 0.70 (\textbf{0.88}) \\

& CardioBERTa.it
& 0.65 (0.85) & 0.64 (0.83) & 0.66 (0.87)
& 0.69 (0.86) & 0.68 (0.84) & 0.70 (\textbf{0.88}) \\

& xlm-roberta-large
& \textbf{0.70} (\textbf{0.86}) & \textbf{0.69} (\textbf{0.84}) & \textbf{0.72} (\textbf{0.88})
& \textbf{0.74} (\textbf{0.87}) & \textbf{0.72} (\textbf{0.85}) & \textbf{0.75} (\textbf{0.88}) \\
\hline

\multirow{3}{*}{Romanian}
& xlm-roberta-base
& 0.66 (0.82) & 0.64 (0.80) & 0.67 (0.83)
& 0.70 (0.84) & 0.69 (0.82) & 0.71 (0.85) \\

& CardioBERTa.ro
& 0.67 (0.83) & 0.66 (\textbf{0.83}) & 0.68 (0.84)
& 0.71 (0.85) & 0.70 (0.84) & 0.71 (0.85) \\

& xlm-roberta-large
& \textbf{0.70} (\textbf{0.85}) & \textbf{0.69} (\textbf{0.83}) & \textbf{0.72} (\textbf{0.87})
& \textbf{0.74} (\textbf{0.86}) & \textbf{0.72} (\textbf{0.85}) & \textbf{0.75} (\textbf{0.88}) \\
\hline

\multirow{3}{*}{Spanish}
& roberta-biomedical-clinical-es
& 0.74 (0.88) & 0.72 (0.86) & 0.75 (\textbf{0.90})
& 0.77 (0.89) & 0.75 (0.87) & \textbf{0.79} (0.91) \\

& CardioBERTa.es
& 0.74 (0.88) & 0.72 (0.86) & \textbf{0.76} (\textbf{0.90})
& 0.76 (0.88) & 0.74 (0.85) & \textbf{0.79} (\textbf{0.92}) \\

& xlm-roberta-large
& \textbf{0.75} (\textbf{0.89}) & \textbf{0.75} (\textbf{0.88}) & \textbf{0.76} (\textbf{0.90})
& \textbf{0.78} (\textbf{0.90}) & \textbf{0.77} (\textbf{0.89}) & 0.78 (0.90) \\
\hline

\multirow{3}{*}{Swedish}
& bert-base-swedish-cased
& 0.66 (0.81) & 0.64 (0.78) & 0.69 (0.85)
& 0.70 (0.83) & 0.68 (0.80) & 0.73 (0.86) \\

& CardioBERTa.sv
& 0.69 (0.85) & 0.68 (0.83) & 0.71 (\textbf{0.87})
& 0.73 (0.86) & 0.72 (0.85) & 0.74 (\textbf{0.88}) \\

& xlm-roberta-large
& \textbf{0.71} (\textbf{0.86}) & \textbf{0.70} (\textbf{0.85}) & \textbf{0.72} (\textbf{0.87})
& \textbf{0.74} (\textbf{0.87}) & \textbf{0.73} (\textbf{0.86}) & \textbf{0.75} (\textbf{0.88}) \\
\hline

Multilingual
& xlm-roberta-large
& 0.72 (0.88) & 0.71 (0.87) & 0.73 (0.90)
& 0.75 (0.89) & 0.74 (0.88) & 0.77 (0.90) \\
\hline
\end{tabular}
}
\label{tab:span_multilabel_sigmoid_results}
\end{table}

\begin{landscape}
\begin{table}
\centering
\caption{\textbf{Span-level NER} performance of \textbf{multilabel} models on the CardioCCC held-out test set, broken down by entity type and averaged across the 10 fold models. Predictions are decoded using \textbf{per-label sigmoid confidence scores} for thresholding, matching the training objective, and then aggregated into entity spans. Because argmax label selection is invariant to sigmoid versus softmax scaling, each token receives a single label. Strict (exact boundary) and relaxed (overlapping boundary) precision, recall, and F1 are reported for each of the four entity types. XLM\mbox{-}RoBERTa\mbox{-}large is trained jointly on all seven languages and evaluated on the full multilingual test set and separately by language.}
\label{tab:span_multilabel_sigmoid_per_entity}
\resizebox{\columnwidth}{!}{%
\begin{tabular}{l|l|cccc|cccc|cccc}
\multicolumn{1}{l}{} & \multicolumn{1}{l}{}
& \multicolumn{4}{c}{F1 strict (relaxed)}
& \multicolumn{4}{c}{Recall strict (relaxed)}
& \multicolumn{4}{c}{Precision strict (relaxed)} \\
\hline
Language & Model
& Dis & Med & Proc & Sym
& Dis & Med & Proc & Sym
& Dis & Med & Proc & Sym \\
\hline
\hline

\multirow{3}{*}{Czech}
& robeczech-base
& 0.68 (0.86) & 0.85 (0.89) & 0.72 (0.84) & 0.62 (0.79)
& 0.66 (0.83) & 0.80 (0.85) & 0.68 (0.79) & 0.60 (0.77)
& 0.69 (0.88) & \textbf{0.90} (\textbf{0.95}) & 0.77 (0.89) & 0.65 (0.82) \\

& CardioBERTa.cs
& 0.70 (\textbf{0.87}) & 0.84 (0.89) & 0.73 (0.85) & \textbf{0.64} (0.80)
& 0.69 (0.85) & 0.81 (0.86) & 0.68 (0.79) & 0.61 (0.76)
& \textbf{0.72} (\textbf{0.89}) & 0.87 (0.93) & \textbf{0.79} (\textbf{0.91}) & \textbf{0.67} (\textbf{0.83}) \\

& xlm-roberta-large
& \textbf{0.71} (\textbf{0.87}) & \textbf{0.87} (\textbf{0.91}) & \textbf{0.74} (\textbf{0.87}) & \textbf{0.64} (\textbf{0.82})
& \textbf{0.70} (\textbf{0.87}) & \textbf{0.86} (\textbf{0.90}) & \textbf{0.71} (\textbf{0.83}) & \textbf{0.64} (\textbf{0.82})
& \textbf{0.72} (0.88) & 0.88 (0.92) & 0.77 (0.90) & 0.65 (0.82) \\
\hline

\multirow{3}{*}{Dutch}
& RobBERT2023-base
& 0.63 (0.82) & 0.86 (0.89) & 0.66 (0.81) & 0.58 (0.78)
& 0.61 (0.80) & 0.84 (0.86) & 0.63 (0.77) & 0.59 (0.80)
& 0.64 (0.84) & \textbf{0.89} (\textbf{0.92}) & 0.70 (0.86) & 0.57 (0.76) \\

& CardioBERTa.nl
& 0.64 (0.83) & 0.81 (0.85) & 0.66 (0.80) & 0.61 (0.78)
& 0.61 (0.79) & 0.76 (0.80) & 0.58 (0.71) & 0.56 (0.72)
& 0.67 (0.87) & 0.86 (0.90) & \textbf{0.76} (\textbf{0.92}) & 0.66 (\textbf{0.84}) \\

& xlm-roberta-large
& \textbf{0.70} (\textbf{0.87}) & \textbf{0.87} (\textbf{0.90}) & \textbf{0.72} (\textbf{0.86}) & \textbf{0.67} (\textbf{0.83})
& \textbf{0.70} (\textbf{0.87}) & \textbf{0.86} (\textbf{0.88}) & \textbf{0.68} (\textbf{0.82}) & \textbf{0.66} (\textbf{0.82})
& \textbf{0.71} (\textbf{0.88}) & \textbf{0.89} (0.91) & 0.75 (0.90) & \textbf{0.67} (0.83) \\
\hline

\multirow{3}{*}{English}
& biomed\_roberta\_base
& 0.63 (0.80) & 0.86 (0.92) & 0.69 (0.80) & 0.59 (0.78)
& 0.60 (0.76) & 0.85 (0.91) & 0.65 (0.75) & 0.59 (0.79)
& 0.67 (0.85) & 0.87 (0.93) & 0.73 (0.85) & 0.58 (0.78) \\

& CardioBERTa.en
& 0.63 (0.80) & 0.88 (\textbf{0.93}) & 0.68 (0.79) & 0.60 (0.79)
& 0.59 (0.75) & 0.87 (\textbf{0.93}) & 0.64 (0.73) & 0.58 (0.76)
& 0.67 (0.85) & 0.88 (\textbf{0.94}) & 0.74 (0.85) & 0.62 (0.81) \\

& xlm-roberta-large
& \textbf{0.70} (\textbf{0.84}) & \textbf{0.89} (0.91) & \textbf{0.73} (\textbf{0.84}) & \textbf{0.67} (\textbf{0.83})
& \textbf{0.69} (\textbf{0.83}) & \textbf{0.89} (0.92) & \textbf{0.70} (\textbf{0.82}) & \textbf{0.67} (\textbf{0.83})
& \textbf{0.71} (\textbf{0.86}) & \textbf{0.89} (0.91) & \textbf{0.76} (\textbf{0.88}) & \textbf{0.67} (\textbf{0.83}) \\
\hline

\multirow{3}{*}{Italian}
& bioBIT
& 0.65 (0.86) & 0.86 (\textbf{0.90}) & 0.62 (0.85) & 0.61 (0.82)
& 0.63 (0.85) & 0.85 (0.89) & 0.59 (0.81) & 0.60 (0.80)
& 0.65 (0.88) & 0.86 (\textbf{0.90}) & 0.66 (0.89) & 0.63 (\textbf{0.83}) \\

& CardioBERTa.it
& 0.66 (0.86) & \textbf{0.87} (0.89) & 0.63 (\textbf{0.86}) & 0.62 (0.82)
& 0.64 (0.85) & 0.86 (0.89) & 0.60 (\textbf{0.82}) & 0.61 (0.81)
& 0.67 (0.88) & \textbf{0.87} (\textbf{0.90}) & 0.66 (\textbf{0.90}) & 0.62 (\textbf{0.83}) \\

& xlm-roberta-large
& \textbf{0.71} (\textbf{0.88}) & \textbf{0.87} (\textbf{0.90}) & \textbf{0.72} (\textbf{0.86}) & \textbf{0.64} (\textbf{0.83})
& \textbf{0.69} (\textbf{0.86}) & \textbf{0.87} (\textbf{0.90}) & \textbf{0.69} (\textbf{0.82}) & \textbf{0.64} (\textbf{0.83})
& \textbf{0.72} (\textbf{0.90}) & 0.86 (0.89) & \textbf{0.76} (\textbf{0.90}) & \textbf{0.64} (\textbf{0.83}) \\
\hline

\multirow{3}{*}{Romanian}
& xlm-roberta-base
& 0.66 (0.85) & 0.87 (0.90) & 0.69 (0.84) & 0.57 (0.75)
& 0.66 (0.85) & 0.86 (0.89) & 0.66 (0.80) & 0.56 (0.74)
& 0.67 (0.86) & 0.87 (0.90) & 0.72 (0.88) & 0.58 (0.77) \\

& CardioBERTa.ro
& 0.69 (0.86) & 0.86 (0.90) & 0.71 (0.85) & 0.58 (0.77)
& 0.69 (\textbf{0.87}) & 0.86 (0.90) & 0.69 (0.83) & 0.58 (\textbf{0.77})
& 0.68 (0.86) & 0.85 (0.89) & 0.73 (0.88) & 0.58 (0.77) \\

& xlm-roberta-large
& \textbf{0.71} (\textbf{0.88}) & \textbf{0.88} (\textbf{0.92}) & \textbf{0.74} (\textbf{0.87}) & \textbf{0.61} (\textbf{0.79})
& \textbf{0.70} (\textbf{0.87}) & \textbf{0.87} (\textbf{0.91}) & \textbf{0.72} (\textbf{0.84}) & \textbf{0.60} (\textbf{0.77})
& \textbf{0.72} (\textbf{0.90}) & \textbf{0.89} (\textbf{0.92}) & \textbf{0.77} (\textbf{0.90}) & \textbf{0.63} (\textbf{0.80}) \\
\hline

\multirow{3}{*}{Spanish}
& roberta-biomedical-clinical-es
& 0.72 (0.89) & 0.87 (\textbf{0.92}) & 0.77 (0.89) & 0.69 (0.86)
& 0.70 (0.86) & 0.84 (0.88) & 0.74 (0.86) & \textbf{0.70} (0.86)
& \textbf{0.75} (\textbf{0.91}) & \textbf{0.91} (0.95) & 0.80 (0.92) & 0.69 (0.86) \\

& CardioBERTa.es
& 0.73 (0.89) & 0.85 (0.90) & 0.77 (0.89) & 0.69 (0.86)
& 0.73 (0.88) & 0.80 (0.84) & 0.74 (0.86) & 0.67 (0.84)
& 0.74 (0.90) & \textbf{0.91} (\textbf{0.96}) & 0.80 (0.92) & \textbf{0.71} (\textbf{0.88}) \\

& xlm-roberta-large
& \textbf{0.74} (\textbf{0.90}) & \textbf{0.88} (0.90) & \textbf{0.79} (\textbf{0.90}) & \textbf{0.70} (\textbf{0.87})
& \textbf{0.74} (\textbf{0.90}) & \textbf{0.89} (\textbf{0.91}) & \textbf{0.77} (\textbf{0.88}) & \textbf{0.70} (\textbf{0.87})
& \textbf{0.75} (\textbf{0.91}) & 0.88 (0.90) & \textbf{0.81} (\textbf{0.93}) & 0.70 (0.87) \\
\hline

\multirow{3}{*}{Swedish}
& bert-base-swedish-cased
& 0.65 (0.83) & 0.86 (0.88) & 0.69 (0.83) & 0.60 (0.78)
& 0.64 (0.81) & 0.85 (0.86) & 0.64 (0.77) & 0.58 (0.74)
& 0.66 (0.85) & 0.87 (0.89) & \textbf{0.75} (0.89) & 0.63 (0.81) \\

& CardioBERTa.sv
& 0.69 (\textbf{0.87}) & \textbf{0.90} (\textbf{0.92}) & 0.71 (0.85) & 0.63 (0.81)
& 0.69 (\textbf{0.87}) & \textbf{0.91} (\textbf{0.93}) & 0.68 (0.82) & 0.62 (0.79)
& 0.69 (0.87) & \textbf{0.89} (\textbf{0.91}) & 0.74 (0.89) & 0.65 (\textbf{0.83}) \\

& xlm-roberta-large
& \textbf{0.71} (\textbf{0.87}) & 0.86 (0.90) & \textbf{0.72} (\textbf{0.86}) & \textbf{0.66} (\textbf{0.83})
& \textbf{0.70} (\textbf{0.87}) & 0.87 (0.90) & \textbf{0.70} (\textbf{0.83}) & \textbf{0.66} (\textbf{0.83})
& \textbf{0.71} (\textbf{0.88}) & 0.86 (0.90) & \textbf{0.75} (\textbf{0.90}) & \textbf{0.67} (\textbf{0.83}) \\
\hline

Multilingual
& xlm-roberta-large
& 0.72 (0.90) & 0.89 (0.91) & 0.74 (0.88) & 0.66 (0.87)
& 0.71 (0.88) & 0.89 (0.91) & 0.72 (0.85) & 0.66 (0.86)
& 0.73 (0.90) & 0.89 (0.91) & 0.77 (0.92) & 0.67 (0.87) \\
\hline
\end{tabular}
}
\end{table}
\end{landscape}

\begin{table}[]
\centering
\caption{\textbf{Span-level NER} performance of \textbf{multiclass} models on the CardioCCC held-out test set for \textbf{disease} recognition, averaged across the 10 fold models. Each model is trained only on the disease entity type. Predictions are decoded using argmax over \textbf{softmax-normalised logits} and aggregated into entity spans. Precision, recall, and F1 are reported under strict (exact boundary) and relaxed (overlapping boundary) matching. XLM-RoBERTa-large is trained jointly on all seven languages and evaluated on the full multilingual test set and separately by language.}
\label{tab:ner-disease}
\resizebox{0.93\textwidth}{!}{
\begin{tabular}{l|l|c|c|c}
    \multirow{2}{*}{Language} & \multirow{2}{*}{Model} & F1 & Recall & Precision \\
    & & strict (relaxed) & strict (relaxed) & strict (relaxed) \\
    \hline
    \hline
    \multirow{3}{*}{Czech} & robeczech-base & 0.67 (0.85) & 0.68 (0.87) & 0.65 (0.84) \\
     & CardioBERTa.cs & 0.69 (0.86) & 0.72 (\textbf{0.89}) & 0.67 (0.84) \\
     & xlm-roberta-large & \textbf{0.72} (\textbf{0.88}) & \textbf{0.73} (\textbf{0.89}) & \textbf{0.72} (\textbf{0.87}) \\
    \hline
    \multirow{3}{*}{Dutch} & RobBERT2023-base & 0.63 (0.85) & 0.61 (0.82) & 0.64 (0.85) \\
     & CardioBERTa.nl & 0.66 (0.86) & 0.64 (0.83) & 0.68 (\textbf{0.89}) \\
     & xlm-roberta-large & \textbf{0.72} (\textbf{0.88}) & \textbf{0.73} (\textbf{0.89}) & \textbf{0.71} (0.87) \\
    \hline
    \multirow{3}{*}{English} & biomed\_roberta\_base & 0.64 (0.81) & 0.62 (0.78) & 0.68 (0.85) \\
     & CardioBERTa.en & 0.64 (0.81) & 0.61 (0.78) & 0.66 (0.84) \\
     & xlm-roberta-large & \textbf{0.71} (\textbf{0.84}) & \textbf{0.70} (\textbf{0.83}) & \textbf{0.72} (\textbf{0.86}) \\
    \hline
    \multirow{3}{*}{Italian} & bioBIT & 0.63 (0.83) & 0.62 (0.81) & 0.65 (0.85) \\
     & CardioBERTa.it & 0.64 (0.83) & 0.61 (0.80) & 0.67 (0.87) \\
     & xlm-roberta-large & \textbf{0.71} (\textbf{0.88}) & \textbf{0.70} (\textbf{0.87}) & \textbf{0.71} (\textbf{0.89}) \\
    \hline
    \multirow{3}{*}{Romanian} & xlm-roberta-base & 0.67 (0.86) & 0.70 (0.89) & 0.65 (0.83) \\
     & CardioBERTa.ro & 0.70 (0.87) & 0.72 (\textbf{0.90}) & 0.68 (0.85) \\
     & xlm-roberta-large & \textbf{0.73} (\textbf{0.89}) & \textbf{0.73} (0.89) & \textbf{0.73} (\textbf{0.89}) \\
    \hline
    \multirow{3}{*}{Spanish} & roberta-biomedical-clinical-es & \textbf{0.76} (0.90) & 0.76 (0.90) & \textbf{0.76} (\textbf{0.90}) \\
     & CardioBERTa.es & \textbf{0.76} (0.90) & \textbf{0.78} (\textbf{0.92}) & 0.74 (0.89) \\
     & xlm-roberta-large & \textbf{0.76} (\textbf{0.91}) & 0.77 (\textbf{0.92}) & 0.74 (0.89) \\
    \hline
    \multirow{3}{*}{Swedish} & bert-base-swedish-cased & 0.65 (0.84) & 0.66 (0.85) & 0.64 (0.83) \\
     & CardioBERTa.sv & 0.69 (0.87) & 0.70 (\textbf{0.89}) & 0.68 (0.86) \\
     & xlm-roberta-large & \textbf{0.72} (\textbf{0.88}) & \textbf{0.72} (\textbf{0.89}) & \textbf{0.71} (\textbf{0.88}) \\
    \hline
    Multilingual & xlm-roberta-large & 0.73 (0.90) & 0.73 (0.90) & 0.73 (0.90) \\
    \hline
\end{tabular}
}
\end{table}

\begin{table}[]
\centering
\caption{\textbf{Span-level NER} performance of \textbf{multiclass} models on the CardioCCC held-out test set for \textbf{medication} recognition, averaged across the 10 fold models. Each model is trained only on the medication entity type. Predictions are decoded using argmax over \textbf{softmax-normalised logits} and aggregated into entity spans. Precision, recall, and F1 are reported under strict (exact boundary) and relaxed (overlapping boundary) matching. XLM-RoBERTa-large is trained jointly on all seven languages and evaluated on the full multilingual test set and separately by language.}
\label{tab:ner-medication}
\resizebox{0.93\textwidth}{!}{
\begin{tabular}{l|l|c|c|c}
    \multirow{2}{*}{Language} & \multirow{2}{*}{Model} & F1 & Recall & Precision \\
    & & strict (relaxed) & strict (relaxed) & strict (relaxed) \\
    \hline
    \hline
    \multirow{3}{*}{Czech} & robeczech-base & 0.86 (0.91) & 0.85 (0.90) & 0.87 (0.92) \\
     & CardioBERTa.cs & 0.85 (\textbf{0.92}) & 0.85 (\textbf{0.92}) & 0.84 (0.91) \\
     & xlm-roberta-large & \textbf{0.87} (\textbf{0.92}) & \textbf{0.87} (\textbf{0.92}) & \textbf{0.88} (\textbf{0.93}) \\
    \hline
    \multirow{3}{*}{Dutch} & RobBERT2023-base & \textbf{0.88} (0.90) & \textbf{0.87} (0.88) & \textbf{0.90} (\textbf{0.92}) \\
     & CardioBERTa.nl & 0.86 (0.90) & 0.85 (0.89) & 0.87 (0.91) \\
     & xlm-roberta-large & \textbf{0.88} (\textbf{0.91}) & \textbf{0.87} (\textbf{0.90}) & 0.89 (\textbf{0.92}) \\
    \hline
    \multirow{3}{*}{English} & biomed\_roberta\_base & \textbf{0.89} (\textbf{0.94}) & 0.88 (0.93) & \textbf{0.90} (\textbf{0.95}) \\
     & CardioBERTa.en & 0.88 (\textbf{0.94}) & 0.88 (\textbf{0.94}) & 0.88 (0.93) \\
     & xlm-roberta-large & \textbf{0.89} (0.92) & \textbf{0.90} (\textbf{0.94}) & 0.88 (0.91) \\
    \hline
    \multirow{3}{*}{Italian} & bioBIT & 0.87 (0.91) & 0.83 (0.87) & \textbf{0.92} (0.96) \\
     & CardioBERTa.it & \textbf{0.88} (\textbf{0.92}) & 0.84 (0.88) & \textbf{0.92} (\textbf{0.97}) \\
     & xlm-roberta-large & 0.87 (0.90) & \textbf{0.88} (\textbf{0.92}) & 0.85 (0.89) \\
    \hline
    \multirow{3}{*}{Romanian} & xlm-roberta-base & 0.87 (0.91) & 0.87 (0.91) & 0.88 (0.92) \\
     & CardioBERTa.ro & \textbf{0.89} (0.92) & \textbf{0.90} (\textbf{0.93}) & 0.89 (0.92) \\
     & xlm-roberta-large & \textbf{0.89} (\textbf{0.93}) & 0.88 (0.92) & \textbf{0.90} (\textbf{0.93}) \\
    \hline
    \multirow{3}{*}{Spanish} & roberta-biomedical-clinical-es & \textbf{0.92} (\textbf{0.95}) & 0.91 (0.94) & \textbf{0.93} (\textbf{0.96}) \\
     & CardioBERTa.es & \textbf{0.92} (\textbf{0.95}) & \textbf{0.92} (\textbf{0.96}) & 0.91 (0.95) \\
     & xlm-roberta-large & 0.90 (0.93) & \textbf{0.92} (0.94) & 0.89 (0.91) \\
    \hline
    \multirow{3}{*}{Swedish} & bert-base-swedish-cased & 0.88 (0.91) & 0.90 (0.93) & 0.87 (0.90) \\
     & CardioBERTa.sv & \textbf{0.91} (\textbf{0.92}) & \textbf{0.93} (\textbf{0.95}) & \textbf{0.88} (0.90) \\
     & xlm-roberta-large & 0.88 (\textbf{0.92}) & 0.89 (0.93) & 0.87 (\textbf{0.91}) \\
    \hline
    Multilingual & xlm-roberta-large & 0.90 (0.92) & 0.90 (0.93) & 0.90 (0.92) \\
    \hline
\end{tabular}
}
\end{table}

\begin{table}[]
\centering
\caption{\textbf{Span-level NER} performance of \textbf{multiclass} models on the CardioCCC held-out test set for \textbf{procedure} recognition, averaged across the 10 fold models. Each model is trained only on the procedure entity type. Predictions are decoded using argmax over \textbf{softmax-normalised logits} and aggregated into entity spans. Precision, recall, and F1 are reported under strict (exact boundary) and relaxed (overlapping boundary) matching. XLM-RoBERTa-large is trained jointly on all seven languages and evaluated on the full multilingual test set and separately by language.}
\label{tab:ner-procedure}
\resizebox{0.93\textwidth}{!}{
\begin{tabular}{l|l|c|c|c}
    \multirow{2}{*}{Language} & \multirow{2}{*}{Model} & F1 & Recall & Precision \\
    & & strict (relaxed) & strict (relaxed) & strict (relaxed) \\
    \hline
    \hline
    \multirow{3}{*}{Czech} & robeczech-base & 0.74 (0.86) & 0.72 (0.83) & 0.76 (0.88) \\
     & CardioBERTa.cs & 0.75 (0.87) & 0.73 (0.85) & 0.77 (0.89) \\
     & xlm-roberta-large & \textbf{0.77} (\textbf{0.88}) & \textbf{0.75} (\textbf{0.86}) & \textbf{0.78} (\textbf{0.90}) \\
    \hline
    \multirow{3}{*}{Dutch} & RobBERT2023-base & 0.67 (0.82) & 0.62 (0.76) & 0.72 (0.88) \\
     & CardioBERTa.nl & 0.68 (0.82) & 0.62 (0.75) & 0.74 (\textbf{0.90}) \\
     & xlm-roberta-large & \textbf{0.74} (\textbf{0.88}) & \textbf{0.72} (\textbf{0.86}) & \textbf{0.76} (\textbf{0.90}) \\
    \hline
    \multirow{3}{*}{English} & biomed\_roberta\_base & 0.67 (0.77) & 0.62 (0.72) & 0.73 (0.84) \\
     & CardioBERTa.en & 0.67 (0.77) & 0.61 (0.70) & 0.74 (0.85) \\
     & xlm-roberta-large & \textbf{0.75} (\textbf{0.86}) & \textbf{0.74} (\textbf{0.84}) & \textbf{0.77} (\textbf{0.87}) \\
    \hline
    \multirow{3}{*}{Italian} & bioBIT & 0.60 (0.82) & 0.55 (0.75) & 0.65 (\textbf{0.90}) \\
     & CardioBERTa.it & 0.59 (0.81) & 0.54 (0.74) & 0.66 (\textbf{0.90}) \\
     & xlm-roberta-large & \textbf{0.73} (\textbf{0.87}) & \textbf{0.71} (\textbf{0.85}) & \textbf{0.75} (\textbf{0.90}) \\
    \hline
    \multirow{3}{*}{Romanian} & xlm-roberta-base & 0.72 (0.86) & 0.71 (0.85) & 0.73 (0.87) \\
     & CardioBERTa.ro & 0.73 (0.87) & 0.73 (\textbf{0.87}) & 0.73 (0.87) \\
     & xlm-roberta-large & \textbf{0.76} (\textbf{0.88}) & \textbf{0.74} (0.86) & \textbf{0.77} (\textbf{0.89}) \\
    \hline
    \multirow{3}{*}{Spanish} & roberta-biomedical-clinical-es & 0.81 (\textbf{0.92}) & 0.80 (0.90) & \textbf{0.82} (\textbf{0.93}) \\
     & CardioBERTa.es & 0.81 (\textbf{0.92}) & \textbf{0.82} (\textbf{0.92}) & 0.81 (0.92) \\
     & xlm-roberta-large & \textbf{0.82} (\textbf{0.92}) & \textbf{0.82} (\textbf{0.92}) & \textbf{0.82} (0.92) \\
    \hline
    \multirow{3}{*}{Swedish} & bert-base-swedish-cased & 0.70 (0.85) & 0.69 (0.83) & 0.73 (0.88) \\
     & CardioBERTa.sv & 0.73 (0.87) & 0.73 (\textbf{0.87}) & 0.73 (0.88) \\
     & xlm-roberta-large & \textbf{0.75} (\textbf{0.89}) & \textbf{0.74} (\textbf{0.87}) & \textbf{0.77} (\textbf{0.90}) \\
    \hline
    Multilingual & xlm-roberta-large & 0.77 (0.90) & 0.76 (0.88) & 0.78 (0.91) \\
    \hline
\end{tabular}
}
\end{table}

\begin{table}[]
\centering
\caption{\textbf{Span-level NER} performance of \textbf{multiclass} models on the CardioCCC held-out test set for \textbf{symptom} recognition, averaged across the 10 fold models. Each model is trained only on the symptom entity type. Predictions are decoded using argmax over \textbf{softmax-normalised logits} and aggregated into entity spans. Precision, recall, and F1 are reported under strict (exact boundary) and relaxed (overlapping boundary) matching. XLM-RoBERTa-large is trained jointly on all seven languages and evaluated on the full multilingual test set and separately by language.}
\label{tab:ner-symptom}
\resizebox{0.93\textwidth}{!}{
\begin{tabular}{l|l|c|c|c}
    \multirow{2}{*}{Language} & \multirow{2}{*}{Model} & F1 & Recall & Precision \\
    & & strict (relaxed) & strict (relaxed) & strict (relaxed) \\
    \hline
    \hline
    \multirow{3}{*}{Czech} & robeczech-base & 0.61 (0.79) & 0.61 (0.79) & 0.61 (0.80) \\
     & CardioBERTa.cs & 0.62 (0.80) & 0.65 (0.83) & 0.61 (0.77) \\
     & xlm-roberta-large & \textbf{0.68} (\textbf{0.83}) & \textbf{0.69} (\textbf{0.85}) & \textbf{0.66} (\textbf{0.81}) \\
    \hline
    \multirow{3}{*}{Dutch} & RobBERT2023-base & 0.62 (0.77) & 0.63 (0.81) & 0.60 (0.77) \\
     & CardioBERTa.nl & 0.65 (0.80) & 0.64 (0.79) & 0.65 (0.81) \\
     & xlm-roberta-large & \textbf{0.69} (\textbf{0.84}) & \textbf{0.70} (\textbf{0.85}) & \textbf{0.68} (\textbf{0.82}) \\
    \hline
    \multirow{3}{*}{English} & biomed\_roberta\_base & 0.64 (0.80) & 0.64 (0.81) & 0.63 (0.79) \\
     & CardioBERTa.en & 0.62 (0.80) & 0.63 (0.80) & 0.62 (0.79) \\
     & xlm-roberta-large & \textbf{0.70} (\textbf{0.84}) & \textbf{0.72} (\textbf{0.86}) & \textbf{0.69} (\textbf{0.82}) \\
    \hline
    \multirow{3}{*}{Italian} & bioBIT & 0.60 (0.78) & 0.57 (0.73) & 0.64 (0.83) \\
     & CardioBERTa.it & 0.61 (0.79) & 0.59 (0.74) & 0.65 (\textbf{0.84}) \\
     & xlm-roberta-large & \textbf{0.67} (\textbf{0.84}) & \textbf{0.69} (\textbf{0.85}) & \textbf{0.66} (0.83) \\
    \hline
    \multirow{3}{*}{Romanian} & xlm-roberta-base & 0.60 (0.76) & 0.60 (0.75) & 0.60 (0.76) \\
     & CardioBERTa.ro & 0.62 (0.78) & \textbf{0.64} (\textbf{0.79}) & 0.61 (0.76) \\
     & xlm-roberta-large & \textbf{0.64} (\textbf{0.79}) & \textbf{0.64} (0.78) & \textbf{0.65} (\textbf{0.79}) \\
    \hline
    \multirow{3}{*}{Spanish} & roberta-biomedical-clinical-es & 0.73 (0.86) & 0.73 (0.86) & 0.73 (\textbf{0.87}) \\
     & CardioBERTa.es & 0.73 (0.87) & 0.73 (0.87) & \textbf{0.74} (\textbf{0.87}) \\
     & xlm-roberta-large & \textbf{0.74} (\textbf{0.88}) & \textbf{0.76} (\textbf{0.89}) & 0.73 (0.86) \\
    \hline
    \multirow{3}{*}{Swedish} & bert-base-swedish-cased & 0.57 (0.78) & 0.56 (0.76) & 0.58 (0.80) \\
     & CardioBERTa.sv & 0.60 (0.81) & 0.59 (0.79) & 0.62 (\textbf{0.82}) \\
     & xlm-roberta-large & \textbf{0.70} (\textbf{0.84}) & \textbf{0.71} (\textbf{0.86}) & \textbf{0.68} (\textbf{0.82}) \\
    \hline
    Multilingual & xlm-roberta-large & 0.70 (0.87) & 0.71 (0.89) & 0.69 (0.86) \\
    \hline
\end{tabular}
}
\end{table}

% CZECH

\begin{table}
\centering
\caption{\textbf{Full-text (medical-span)} pseudo-perplexity scores for \textbf{Czech} models evaluated on Czech ParaCliTe texts, broken down by document source language. Scores were computed following \textbf{Kauf and Ivanova}~\cite{Kauf2023} using the \textbf{minicons} library~\cite{Misra2022}, with the median and geometric mean aggregated across documents within each source-language group. Lower values indicate a better model fit to the text.}
\label{tab:Czech_perplexity_minicons}
\resizebox{0.85\textwidth}{!}{
\begin{tabular}{l|l|cc}
\multirow{2}{*}{Model} & \multirow{2}{*}{Source language} & \multicolumn{2}{c}{Full-text PPL (Medical-span PPL)} \\
& & Median & Geometric mean \\
\hline
\hline
\multirow[t]{7}{*}{ufal/robeczech-base} & Czech & 93.08 (135.79) & 90.76 (154.51) \\
 & Dutch & 69.16 (34.04) & 69.77 (44.00) \\
 & English & 43.09 (20.78) & 40.85 (26.07) \\
 & Italian & 36.07 (24.75) & 35.60 (31.30) \\
 & Romanian & 56.92 (58.68) & 58.11 (67.13) \\
 & Spanish & 41.37 (34.46) & 43.77 (46.60) \\
 & Swedish & 67.60 (32.91) & 67.40 (42.17) \\
\hline
\multirow[t]{7}{*}{DT4H/CardioBERTa.cs} & Czech & 20.65 (49.17) & 19.37 (63.18) \\
 & Dutch & 7.76 (9.46) & 7.10 (12.74) \\
 & English & 4.26 (4.36) & 4.36 (5.95) \\
 & Italian & 5.18 (5.89) & 5.21 (8.69) \\
 & Romanian & 9.61 (15.94) & 9.70 (22.08) \\
 & Spanish & 7.02 (9.00) & 7.29 (12.43) \\
 & Swedish & 7.43 (6.59) & 6.84 (11.31) \\
\hline
\end{tabular}}
\end{table}

\begin{table}
\centering
\caption{\textbf{Full-text (medical-span)} pseudo-perplexity scores for \textbf{Czech} models evaluated on Czech ParaCliTe texts, broken down by  document source language. Scores were computed following \textbf{Salazar et al.}~\cite{Salazar2020} using \textbf{our implementation}, with the median and geometric mean aggregated across documents within each source-language group. Lower values indicate a better model fit to the text.}
\label{tab:Czech_perplexity_ours}
\resizebox{0.85\textwidth}{!}{
\begin{tabular}{l|l|cc}
\multirow{2}{*}{Model} & \multirow{2}{*}{Source language} & \multicolumn{2}{c}{Full-text PPL (Medical-span PPL)} \\
& & Median & Geometric mean \\
\hline
\hline
\multirow[t]{7}{*}{ufal/robeczech-base} & Czech & 43.70 (19.27) & 42.96 (33.03) \\
 & Dutch & 30.20 (4.79) & 31.77 (9.65) \\
 & English & 18.39 (2.56) & 18.13 (5.96) \\
 & Italian & 15.01 (3.28) & 14.58 (7.08) \\
 & Romanian & 19.71 (6.71) & 19.31 (12.18) \\
 & Spanish & 15.13 (3.93) & 16.00 (8.56) \\
 & Swedish & 28.55 (3.96) & 28.99 (9.61) \\
\hline
 \multirow[t]{7}{*}{DT4H/CardioBERTa.cs} & Czech & 11.66 (7.28) & 11.48 (18.90) \\
 & Dutch & 3.77 (1.69) & 3.66 (3.58) \\
 & English & 2.29 (1.13) & 2.36 (1.96) \\
 & Italian & 2.66 (1.40) & 2.69 (2.98) \\
 & Romanian & 4.25 (3.09) & 4.16 (6.04) \\
 & Spanish & 3.15 (1.59) & 3.28 (3.30) \\
 & Swedish & 3.56 (1.25) & 3.36 (3.27) \\
\hline
\end{tabular}}
\end{table}

% DUTCH

\begin{table}
\caption{\textbf{Full-text (medical-span)} pseudo-perplexity scores for \textbf{Dutch} models evaluated on Dutch ParaCliTe texts, broken down by document source language. Scores were computed following \textbf{Kauf and Ivanova}~\cite{Kauf2023} using the \textbf{minicons} library~\cite{Misra2022}, with the median and geometric mean aggregated across documents within each source-language group. Lower values indicate a better model fit to the text.}
\label{tab:Dutch_perplexity_minicons}
\resizebox{\textwidth}{!}{
\begin{tabular}{l|l|cc}
\multirow{2}{*}{Model} & \multirow{2}{*}{Source language} & \multicolumn{2}{c}{Full-text PPL (Medical-span PPL)} \\
& & Median & Geometric mean \\
\hline
\hline
\multirow[t]{7}{*}{CLTL/MedRoBERTa.nl} & Czech  & 34.47 (74.7) & 34.65 (101) \\
 & Dutch   & 27.71 (8.11)  & 24.79 (16.2)  \\
 & English & 33.23 (29.3)  & 31.62 (41.6)  \\
 & Italian & 57.70 (207.7) & 56.27 (242.4) \\
 & Romanian& 65.18 (239.8) & 65.22 (259.8) \\
 & Spanish & 34.53 (77.4)  & 34.47 (85.2)  \\
 & Swedish & 93.25 (82.9)  & 93.79 (106.5) \\
\hline
\multirow[t]{7}{*}{DTAI-KULeuven/robbert-2023-dutch-base} & Czech  & 10.41 (51.6) & 9.89 (61.6) \\
 & Dutch    & 14.21 (61.1) & 13.87 (63.2) \\
 & English  & 9.36  (28.7) & 9.13 (32.8)  \\
 & Italian  & 12.76 (65.4) & 12.95 (78.5) \\
 & Romanian & 12.67 (66)   & 12.90 (78.1) \\
 & Spanish  & 13.75 (53.9) & 14.18 (61.8) \\
 & Swedish  & 14.04 (63.7) & 14.60 (69.3) \\
\hline
\multirow[t]{7}{*}{UMCU/CardioBERTa.nl} & Czech  & 9.33 (16.3) & 9.44 (28.5) \\
 & Dutch     & 13.95 (16.1) & 14.42 (32.4) \\
 & English   & 4.30 (6.4)   & 4.55 (11.5)  \\
 & Italian   & 12.14 (36.1) & 12.16 (57.7  \\
 & Romanian  & 13.11 (35.3) & 12.89 (56)   \\
 & Spanish   & 9.19  (17.1) & 8.99 (24.3)  \\
 & Swedish   & 8.93 (14.5)  & 9.09 (26.7)  \\
\hline
\end{tabular}}
\end{table}

\begin{table}
\caption{\textbf{Full-text (medical-span)} pseudo-perplexity scores for \textbf{Dutch} models evaluated on Dutch ParaCliTe texts, broken down by document source language. Scores were computed following \textbf{Salazar et al.}~\cite{Salazar2020} using \textbf{our implementation}, with the median and geometric mean aggregated across documents within each source-language group. Lower values indicate a better model fit to the text.}
\label{tab:Dutch_perplexity_ours}
\resizebox{\textwidth}{!}{
\begin{tabular}{l|l|cc}
\multirow{2}{*}{Model} & \multirow{2}{*}{Source language} & \multicolumn{2}{c}{Full-text PPL (Medical-span PPL)} \\
& & Median & Geometric mean \\
\hline
\hline
\multirow[t]{7}{*}{CLTL/MedRoBERTa.nl} & Czech & 25.93 (30.2) & 24.88 (51.7) \\
 & Dutch    & 20.55 (4.9)   & 19.79 (11)    \\
 & English  & 21.28 (14.8)  & 22.51 (25.7)  \\
 & Italian  & 37.76 (62.5)  & 37.12 (107.7) \\
 & Romanian & 40.96 (78.4)  & 39.56 (117.7) \\
 & Spanish  & 24.08 (29.1)  & 24.42 (48.6)  \\
 & Swedish  & 56.50 (20.3)  & 53.56 (34.9)  \\
\hline
\multirow[t]{7}{*}{DTAI-KULeuven/robbert-2023-dutch-base} & Czech  & 5.60 (9.3) & 5.44 (18) \\
 & Dutch    & 7.30 (12.3)  & 7.09 (18.21) \\
 & English  & 4.78 (5.5)   & 4.66 (9.15)  \\
 & Italian  & 6.23 (11.2)  & 6.40 (20.4)  \\
 & Romanian & 5.80 (14.61) & 5.99 (23.97) \\
 & Spanish  & 6.76 (9.5)   & 6.74 (16)    \\
 & Swedish  & 7.30 (9.63)  & 7.17 (19.13) \\
\hline
\multirow[t]{7}{*}{UMCU/CardioBERTa.nl} & Czech & 6.84 (8.16) & 6.99 (16.8) \\
 & Dutch    & 11.76 (10.4)      & 12.54 (23.8) \\
 & English  & 3.47 (3.36)       & 3.59 (7.9)   \\
 & Italian  & 8.27 (12.5)       & 8.03 (27.8)  \\
 & Romanian & 8.18 (12.3)       & 7.87 (26.4)  \\
 & Spanish  & 6.35 (8.36)       & 6.50 (14.7)  \\
 & Swedish  & 5.68 (5.61)       & 5.76 (12.6)  \\
\hline
\end{tabular}}
\end{table}

% ENGLISH
\begin{table}
\centering
\caption{\textbf{Full-text (medical-span)} pseudo-perplexity scores for \textbf{English} models evaluated on English ParaCliTe texts, broken down by document source language. Scores were computed following \textbf{Kauf and Ivanova}~\cite{Kauf2023} using the \textbf{minicons} library~\cite{Misra2022}, with the median and geometric mean aggregated across documents within each source-language group. Lower values indicate a better model fit to the text.}
\label{tab:English_perplexity_minicons}
\resizebox{0.85\textwidth}{!}{
\begin{tabular}{l|l|cc}
\multirow{2}{*}{Model} & \multirow{2}{*}{Source language} & \multicolumn{2}{c}{Full-text PPL (Medical-span PPL)} \\
& & Median & Geometric mean \\
\hline
\hline
\multirow[t]{7}{*}{allenai/biomed\_roberta\_base} & Czech & 25.83 (18.09) & 25.42 (34.45) \\
 & Dutch & 33.15 (17.21) & 36.89 (27.05) \\
 & English & 20.36 (6.28) & 20.58 (10.34) \\
 & Italian & 16.67 (7.47) & 16.52 (15.57) \\
 & Romanian & 27.29 (20.38) & 27.23 (38.58) \\
 & Spanish & 20.86 (12.81) & 22.47 (18.73) \\
 & Swedish & 54.68 (12.31) & 53.00 (18.80) \\
\hline
\multirow[t]{7}{*}{DT4H/CardioBERTa.en} & Czech & 5.10 (4.89) & 5.20 (10.45) \\
 & Dutch & 4.44 (4.44) & 4.26 (7.11) \\
 & English & 2.76 (2.08) & 2.96 (3.30) \\
 & Italian & 3.55 (2.92) & 3.60 (5.22) \\
 & Romanian & 4.74 (5.49) & 4.90 (9.40) \\
 & Spanish & 4.17 (3.51) & 4.14 (5.26) \\
 & Swedish & 4.73 (3.17) & 4.50 (5.73) \\
\hline
\end{tabular}}
\end{table}

\begin{table}
\centering
\caption{\textbf{Full-text (medical-span)} pseudo-perplexity scores for \textbf{English} models evaluated on English ParaCliTe texts, broken down by document source language. Scores were computed following \textbf{Salazar et al.}~\cite{Salazar2020} using \textbf{our implementation}, with the median and geometric mean aggregated across documents within each source-language group. Lower values indicate a better model fit to the text.}
\label{tab:English_perplexity_ours}
\resizebox{0.85\textwidth}{!}{
\begin{tabular}{l|l|cc}
\multirow{2}{*}{Model} & \multirow{2}{*}{Source language} & \multicolumn{2}{c}{Full-text PPL (Medical-span PPL)} \\
& & Median & Geometric mean \\
\hline
\hline
\multirow[t]{7}{*}{allenai/biomed\_roberta\_base} & Czech & 16.02 (4.14) & 15.34 (12.21) \\
 & Dutch & 18.10 (2.97) & 21.34 (7.33) \\
 & English & 13.31 (1.50) & 12.77 (3.41) \\
 & Italian & 9.22 (1.92) & 9.31 (5.33) \\
 & Romanian & 14.05 (4.11) & 13.57 (11.81) \\
 & Spanish & 10.46 (2.25) & 10.85 (5.35) \\
 & Swedish & 32.16 (1.99) & 28.20 (5.56) \\
\hline
 \multirow[t]{7}{*}{DT4H/CardioBERTa.en} & Czech & 3.50 (1.72) & 3.50 (4.61) \\
 & Dutch & 2.89 (1.30) & 2.96 (3.13) \\
 & English & 2.08 (1.08) & 2.17 (1.80) \\
 & Italian & 2.37 (1.17) & 2.42 (2.48) \\
 & Romanian & 2.78 (1.56) & 2.82 (3.66) \\
 & Spanish & 2.47 (1.19) & 2.51 (2.31) \\
 & Swedish & 3.03 (1.18) & 2.96 (2.70) \\
\hline
\end{tabular}}
\end{table}

% ITALIAN
\begin{table}
\centering
\caption{\textbf{Full-text (medical-span)} pseudo-perplexity scores for \textbf{Italian} models evaluated on Italian ParaCliTe texts, broken down by document source language. Scores were computed following \textbf{Kauf and Ivanova}~\cite{Kauf2023} using the \textbf{minicons} library~\cite{Misra2022}, with the median and geometric mean aggregated across documents within each source-language group. Lower values indicate a better model fit to the text.}
\label{tab:Italian_perplexity_minicons}
\resizebox{0.85\textwidth}{!}{
\begin{tabular}{l|l|cc}
\multirow{2}{*}{Model} & \multirow{2}{*}{Source language} & \multicolumn{2}{c}{Full-text PPL (Medical-span PPL)} \\
 & & Median & Geometric mean \\
\hline
\hline
\multirow[t]{7}{*}{IVN-RIN/bioBIT} & Czech & 6.30 (6.71) & 6.01 (11.74) \\
 & Dutch & 9.49 (9.21) & 9.18 (14.50) \\
 & English & 3.86 (2.83) & 3.89 (4.84) \\
 & Italian & 11.14 (15.52) & 11.08 (25.40) \\
 & Romanian & 8.88 (14.69) & 8.99 (21.03) \\
 & Spanish & 6.12 (6.40) & 6.26 (8.88) \\
 & Swedish & 8.55 (8.51) & 8.08 (13.62) \\
\hline
\multirow[t]{7}{*}{DT4H/CardioBERTa.it} & Czech & 6.29 (6.57) & 6.08 (11.76) \\
 & Dutch & 9.05 (8.93) & 9.26 (13.76) \\
 & English & 3.93 (2.84) & 3.89 (4.78) \\
 & Italian & 11.62 (15.44) & 11.50 (26.93) \\
 & Romanian & 9.12 (14.46) & 9.36 (21.67) \\
 & Spanish & 6.51 (6.34) & 6.63 (9.28) \\
 & Swedish & 8.29 (6.92) & 7.68 (11.23) \\
\hline
\end{tabular}}
\end{table}

\begin{table}
\centering
\caption{\textbf{Full-text (medical-span)} pseudo-perplexity scores for \textbf{Italian} models evaluated on Italian ParaCliTe texts, broken down by document source language. Scores were computed following \textbf{Salazar et al.}~\cite{Salazar2020} using \textbf{our implementation}, with the median and geometric mean aggregated across documents within each source-language group. Lower values indicate a better model fit to the text.}
\label{tab:Italian_perplexity_ours}
\resizebox{0.85\textwidth}{!}{
\begin{tabular}{l|l|cc}
\multirow{2}{*}{Model} & \multirow{2}{*}{Source language} & \multicolumn{2}{c}{Full-text PPL (Medical-span PPL)} \\
 & & Median & Geometric mean \\
\hline
\hline
\multirow[t]{7}{*}{IVN-RIN/bioBIT} & Czech & 3.34 (1.55) & 3.29 (3.63) \\
 & Dutch & 4.60 (1.61) & 4.62 (3.80) \\
 & English & 2.35 (1.11) & 2.28 (1.85) \\
 & Italian & 4.87 (2.51) & 4.92 (6.60) \\
 & Romanian & 4.15 (2.65) & 4.10 (5.91) \\
 & Spanish & 3.03 (1.40) & 3.03 (2.62) \\
 & Swedish & 4.05 (1.35) & 3.86 (3.20) \\
\hline
\multirow[t]{7}{*}{DT4H/CardioBERTa.it} & Czech & 3.63 (1.64) & 3.58 (4.03) \\
 & Dutch & 4.85 (1.76) & 4.92 (4.07) \\
 & English & 2.52 (1.09) & 2.38 (1.92) \\
 & Italian & 5.48 (2.81) & 5.41 (7.61) \\
 & Romanian & 4.58 (3.17) & 4.63 (6.94) \\
 & Spanish & 3.39 (1.54) & 3.34 (2.96) \\
 & Swedish & 4.49 (1.32) & 4.05 (3.27) \\
\hline
\end{tabular}}
\end{table}

% ROMANIAN

\begin{table}
\centering
\caption{\textbf{Full-text (medical-span)} pseudo-perplexity scores for \textbf{Romanian} models evaluated on Romanian ParaCliTe texts, broken down by document source language. Scores were computed following \textbf{Kauf and Ivanova}~\cite{Kauf2023} using the \textbf{minicons} library~\cite{Misra2022}, with the median and geometric mean aggregated across documents within each source-language group. Lower values indicate a better model fit to the text.}
\label{tab:Romanian_perplexity_minicons}
\resizebox{0.85\textwidth}{!}{
\begin{tabular}{l|l|cc}
\multirow{2}{*}{Model} & \multirow{2}{*}{Source language} & \multicolumn{2}{c}{Full-text PPL (Medical-span PPL)} \\
& & Median & Geometric mean \\
\hline
\hline
\multirow[t]{7}{*}{FacebookAI/xlm-roberta-base} & Czech & 16.97 (37.82) & 16.14 (52.58) \\
 & Dutch & 14.01 (32.38) & 14.30 (41.19) \\
 & English & 8.22 (16.19) & 7.89 (20.19) \\
 & Italian & 17.61 (62.95) & 17.91 (82.48) \\
 & Romanian & 30.17 (118.19) & 30.17 (131.46) \\
 & Spanish & 17.28 (36.54) & 16.58 (42.86) \\
 & Swedish & 15.30 (30.22) & 15.52 (44.01) \\
\hline
\multirow[t]{7}{*}{DT4H/CardioBERTa.ro} & Czech & 8.26 (9.24) & 7.69 (15.02) \\
 & Dutch & 7.55 (7.62) & 7.69 (11.77) \\
 & English & 3.42 (3.34) & 3.36 (4.60) \\
 & Italian & 8.76 (11.78) & 8.63 (21.12) \\
 & Romanian & 18.28 (38.38) & 18.56 (54.79) \\
 & Spanish & 6.95 (6.52) & 6.91 (8.78) \\
 & Swedish & 7.70 (7.63) & 7.68 (12.06) \\
\hline
\end{tabular}}
\end{table}

\begin{table}
\centering
\caption{\textbf{Full-text (medical-span)} pseudo-perplexity scores for \textbf{Romanian} models evaluated on Romanian ParaCliTe texts, broken down by document source language. Scores were computed following \textbf{Salazar et al.}~\cite{Salazar2020} using \textbf{our implementation}, with the median and geometric mean aggregated across documents within each source-language group. Lower values indicate a better model fit to the text.}
\label{tab:Romanian_perplexity_ours}
\resizebox{0.85\textwidth}{!}{
\begin{tabular}{l|l|cc}
\multirow{2}{*}{Model} & \multirow{2}{*}{Source language} & \multicolumn{2}{c}{Full-text PPL (Medical-span PPL)} \\
& & Median & Geometric mean \\
\hline
\hline
\multirow[t]{7}{*}{FacebookAI/xlm-roberta-base} & Czech & 6.07 (4.42) & 5.83 (11.60) \\
 & Dutch & 5.19 (3.68) & 5.37 (9.00) \\
 & English & 3.47 (2.52) & 3.33 (4.96) \\
 & Italian & 6.25 (6.27) & 6.22 (15.48) \\
 & Romanian & 9.84 (14.80) & 9.84 (26.21) \\
 & Spanish & 5.57 (4.22) & 5.49 (7.89) \\
 & Swedish & 5.35 (3.36) & 5.32 (8.80) \\
 \hline
\multirow[t]{7}{*}{DT4H/CardioBERTa.ro} & Czech & 3.26 (1.53) & 3.12 (3.96) \\
 & Dutch & 3.16 (1.25) & 3.22 (2.97) \\
 & English & 1.75 (1.05) & 1.75 (1.49) \\
 & Italian & 3.46 (1.59) & 3.36 (5.12) \\
 & Romanian & 6.13 (5.26) & 6.07 (10.64) \\
 & Spanish & 2.55 (1.19) & 2.54 (2.12) \\
 & Swedish & 3.04 (1.16) & 2.93 (2.79) \\
 \hline
\end{tabular}}
\end{table}

% SPANISH
\begin{table}
\centering
\caption{\textbf{Full-text (medical-span)} pseudo-perplexity scores for \textbf{Spanish} models evaluated on Spanish ParaCliTe texts, broken down by document source language. Scores were computed following \textbf{Kauf and Ivanova}~\cite{Kauf2023} using the \textbf{minicons} library~\cite{Misra2022}, with the median and geometric mean aggregated across documents within each source-language group. Lower values indicate a better model fit to the text.}
\label{tab:spanish_perplexity_minicons}
\resizebox{0.85\textwidth}{!}{
\begin{tabular}{l|l|cc}
\multirow{2}{*}{Model} & \multirow{2}{*}{Source language} & \multicolumn{2}{c}{Full-text PPL} \\
& & Median & Geometric mean \\
\hline
\hline
 PlanTL-GOB-ES/ & Czech   & 34.07 (32.66) & 32.87 (146.24) \\
 \multirow[t]{6}{*}
 {roberta-base-biomedical-clinical-es} 
 & Dutch   & 42.32 (25.87) & 51.74 (122.79) \\
 & English & 28.38 (5.69)  & 25.36 (23.39) \\
 & Italian & 23.17 (15.19) & 25.49 (68.51) \\
 & Romanian& 25.78 (21.68) & 24.49 (100.40) \\
 & Spanish & 14.39 (3.37)  & 15.37 (11.50) \\
 & Swedish & 82.49 (29.54) & 77.49 (272.73) \\
\hline
\multirow[t]{7}{*}{DT4H/CardioBERTa.es}
 & Czech   & 8.37 (18.03) & 8.06 (30.97) \\
 & Dutch   & 6.40 (9.12)  & 6.12 (16.22) \\
 & English & 3.11 (2.52)  & 3.10 (5.54) \\
 & Italian & 6.51 (11.28) & 6.16 (20.38) \\
 & Romanian& 6.11 (10.97) & 6.36 (19.81) \\
 & Spanish & 4.19 (3.30)  & 3.94 (6.41) \\
 & Swedish & 5.27 (5.97)  & 5.31 (11.20) \\
\hline
\end{tabular}}
\end{table}

\begin{table}
\centering
\caption{\textbf{Full-text (medical-span)} pseudo-perplexity scores for \textbf{Spanish} models evaluated on Spanish ParaCliTe texts, broken down by document source language. Scores were computed following \textbf{Salazar et al.}~\cite{Salazar2020} using \textbf{our implementation}, with the median and geometric mean aggregated across documents within each source-language group. Lower values indicate a better model fit to the text.}
\label{tab:spanish_perplexity_ours}
\resizebox{0.85\textwidth}{!}{
\begin{tabular}{l|l|cc}
\multirow{2}{*}{Model} & \multirow{2}{*}{Source language} & \multicolumn{2}{c}{Full-text PPL} \\
& & Median & Geometric mean \\
\hline
\hline
PlanTL-GOB-ES/ & Czech   & 26.81 (18.09) & 25.90 (86.57) \\
\multirow[t]{6}{*}{roberta-base-biomedical-clinical-es}
 & Dutch   & 32.90 (12.03) & 42.22 (67.18) \\
 & English & 23.79 (3.51)  & 21.55 (15.73) \\
 & Italian & 16.39 (5.71)  & 18.34 (37.18) \\
 & Romanian& 17.97 (9.53)  & 17.79 (58.55) \\
 & Spanish & 11.58( 1.94)  & 12.50 (7.11) \\
 & Swedish & 56.93 (18.20) & 60.27 (164.06) \\
\hline
\multirow[t]{7}{*}{DT4H/CardioBERTa.es}
 & Czech   & 6.57 (9.26) & 6.37 (19.03) \\
 & Dutch   & 4.53 (3.83) & 4.38 (9.31) \\
 & English & 2.54 (1.76) & 2.47 (3.83) \\
 & Italian & 4.26 (4.55) & 4.26 (11.13) \\
 & Romanian& 4.20 (5.03) & 4.37 (10.91) \\
 & Spanish & 3.21 (2.27) & 3.14 (4.92) \\
 & Swedish & 4.00 (2.53) & 3.84 (6.58) \\
\hline
\end{tabular}}
\end{table}

% SWEDISH
\begin{table}
\centering
\caption{\textbf{Full-text (medical-span)} pseudo-perplexity scores for \textbf{Swedish} models evaluated on Swedish ParaCliTe texts, broken down by document source language. Scores were computed following \textbf{Kauf and Ivanova}~\cite{Kauf2023} using the \textbf{minicons} library~\cite{Misra2022}, with the median and geometric mean aggregated across documents within each source-language group. Lower values indicate a better model fit to the text.}
\label{tab:swedish_perplexity_minicons}
\resizebox{0.85\textwidth}{!}{
\begin{tabular}{l|l|cc}
\multirow{2}{*}{Model} & \multirow{2}{*}{Source language} & \multicolumn{2}{c}{Full-text PPL (Medical-span PPL)} \\
& & Median & Geometric mean \\
\hline
\hline
\multirow[t]{7}{*}{KB/bert-base-swedish-cased} & Czech & 56.25 (256.15) & 53.23 (241.11) \\
 & Dutch   & 43.59 (156.39) & 39.40 (143.75)  \\
 & English & 36.39 (114.18) & 32.48 (102.8)   \\
 & Italian & 35.29 (178.09) & 35.05 (157.31)  \\
 & Romanian& 50.79 (229.7)  & 54.10 (239.2)   \\
 & Spanish & 52.79 (264.8)  & 51.91 (231.7)   \\
 & Swedish & 78.93 (279.5)  & 79.38 (256.28)  \\
\hline
\multirow[t]{7}{*}{DT4H/CardioBERTa.sv} & Czech & 17.81 (21.72) & 16.14 (40.72) \\
 & Dutch   & 10.46 (10.09) & 10.10 (15.66) \\
 & English & 5.58 (5.33)   & 5.85 (7.22)   \\
 & Italian & 9.63 (10.24)  & 9.06 (19.23)  \\
 & Romanian& 9.45 (10.86)  & 9.19 (19.12)  \\
 & Spanish & 7.78 (8.55)   & 7.85 (11.27)  \\
 & Swedish & 26.44 (16.17) & 22.67 (28.27) \\
\hline
\end{tabular}}
\end{table}

\begin{table}
\centering
\caption{\textbf{Full-text (medical-span)} pseudo-perplexity scores for \textbf{Swedish} models evaluated on Swedish ParaCliTe texts, broken down by document source language. Scores were computed following \textbf{Salazar et al.}~\cite{Salazar2020} using \textbf{our implementation}, with the median and geometric mean aggregated across documents within each source-language group. Lower values indicate a better model fit to the text.}
\label{tab:swedish_perplexity_ours}
\resizebox{0.85\textwidth}{!}{
\begin{tabular}{l|l|cc}
\multirow{2}{*}{Model} & \multirow{2}{*}{Source language} & \multicolumn{2}{c}{Full-text PPL (Medical-span PPL)} \\
& & Median & Geometric mean \\
\hline
\hline
\multirow[t]{7}{*}{KB/bert-base-swedish-cased} & Czech & 39.74 (144) & 38.21 (126) \\
 & Dutch   & 27.94 (62.29)  & 25.18 (67.46)  \\
 & English & 21.80 (43.91)  & 20.31 (47.91)  \\
 & Italian & 25.42 (92.24)  & 24.96 (87.25)  \\
 & Romanian& 34.10 (134.04) & 35.74 (135.88) \\
 & Spanish & 35.67 (152.78) & 35.19 (128.27) \\
 & Swedish & 48.89 (125.81) & 46.89 (120.09) \\
\hline
\multirow[t]{7}{*}{DT4H/CardioBERTa.sv} & Czech & 8.71 (3.72) & 8.14 (10.97) \\
 & Dutch    & 4.65 (1.65) & 4.61 (3.7) \\
 & English  & 2.86 (1.14) & 2.86 (2.06) \\
 & Italian  & 4.77 (2.09) & 4.60 (5.53) \\
 & Romanian & 4.04 (2.03) & 4.02 (5.66) \\
 & Spanish  & 3.40 (1.43) & 3.36 (2.83) \\
 & Swedish  & 9.93 (1.96) & 8.79 (6.20) \\
\hline
\end{tabular}
}
\end{table}

% Medical-span PPL broken down by entity type for each language (MINICONS)
\begin{landscape}
\begin{table}
    \centering
    \caption{\textbf{Medical-span} pseudo-perplexity on the ParaCliTe dataset,
    reported per entity type for each language and model. Scores were computed following \textbf{Kauf and Ivanova}~\cite{Kauf2023} using the \textbf{minicons} library~\cite{Misra2022}.}
    \label{tab:medical_ppl_by_entity_type_minicons}
    \resizebox{\linewidth}{!}{
        \begin{tabular}{l|l|cc|cc|cc|cc}
            \multicolumn{2}{c}{} &
            \multicolumn{2}{c}{DISEASE} &
            \multicolumn{2}{c}{MEDICATION} &
            \multicolumn{2}{c}{PROCEDURE} &
            \multicolumn{2}{c}{SYMPTOM} \\
            \hline
            Language & Model &
            Median & Geo.\ mean &
            Median & Geo.\ mean &
            Median & Geo.\ mean &
            Median & Geo.\ mean \\
            \hline
            \hline

            \multirow{2}{*}{Czech}
            & robeczech-base
            & 22.32 & 27.97
            & 187.03 & 191.19
            & 32.05 & 41.36
            & 45.45 & 58.37 \\

            & CardioBERTa.cs
            & 5.23 & 8.40
            & 39.69 & 51.40
            & 7.57 & 11.90
            & 11.39 & 17.12 \\
            \hline

            \multirow{3}{*}{Dutch}
            & RobBERT2023-base
            & 38.64  & 44.6
            & 180.78 & 165.6
            & 48.5   & 51.5
            & 59.1   & 66.6 \\

            & MedRoBERTa.nl
            & 66.01 & 87.4
            & 232.9 & 277.70
            & 58.8  & 85.1
            & 33.75 & 53.2 \\

            & CardioBERTa.nl\_clinical
            & 18.04 & 29.7
            & 71.9  & 111.9
            & 13.4  & 21.5
            & 12.5  & 22.6 \\
            \hline

            \multirow{2}{*}{English}
            & biomed\_roberta\_base
            & 4.51 & 7.69
            & 27.22 & 31.66
            & 5.49 & 13.13
            & 5.15 & 7.35 \\

            & CardioBERTa.en
            & 1.64 & 2.88
            & 4.83 & 6.71
            & 2.0 & 3.25
            & 1.94 & 2.82 \\
            \hline

            \multirow{2}{*}{Italian}
            & bioBIT
            & 9.24 & 14.69
            & 345.26 & 316.26
            & 8.85 & 14.30
            & 16.97 & 27.15 \\
            
            & CardioBERTa.it
            & 9.70 & 16.96
            & 274.56 & 260.53
            & 10.28 & 16.21
            & 16.44 & 27.22 \\
            \hline

            \multirow{2}{*}{Romanian}
            & xlm-roberta-base
            & 25.94 & 34.11
            & 308.38 & 285.87
            & 40.85 & 45.98
            & 32.31 & 41.51 \\

            & CardioBERTa.ro
            & 5.06 & 9.26
            & 51.42 & 77.79
            & 7.43 & 11.25
            & 9.68 & 12.49 \\
            \hline

            \multirow{2}{*}{Spanish} 
            & PlanTL-GOB-ES/roberta-base-biomedical-clinical-es & 3.105 & 7.450 & 18.590 & 85.530 & 8.822 & 22.700 & 1.758 & 7.745 \\ 
            & DT4H/CardioBERTa.es & 3.091 & 5.389 & 23.723 & 28.177 & 5.760 & 10.716 & 2.067 & 4.112 
            \\ 
            \hline
        
            \multirow{2}{*}{Swedish}
            & bert-base-swedish-cased
            & 172.7 & 145.3
            & 821.1 & 687.5
            & 125.3 & 124.1
            & 192.1 & 197.6 \\

            & CardioBERTa.sv
            & 7.3 & 10.88
            & 48 & 79
            & 7.9 & 11.9
            & 12.1 & 21.6 \\
            \hline
        \end{tabular}%
        }    
\end{table}
\end{landscape}

% Medical-span PPL broken down by entity type for each language (OUR)
   
\begin{landscape}
\begin{table}
    \centering
    \caption{\textbf{Medical-span} pseudo-perplexity on the ParaCliTe dataset,
    reported per entity type for each language and model. Scores were computed following \textbf{Salazar et al.}~\cite{Salazar2020} using \textbf{our implementation}.}
    \label{tab:medical_ppl_by_entity_type_our}
    \resizebox{\columnwidth}{!}{%
        \begin{tabular}{l|l|cc|cc|cc|cc}
            \multicolumn{2}{c}{} &
            \multicolumn{2}{c}{DISEASE} &
            \multicolumn{2}{c}{MEDICATION} &
            \multicolumn{2}{c}{PROCEDURE} &
            \multicolumn{2}{c}{SYMPTOM} \\
            \hline
            Language & Model &
            Median & Geo.\ mean &
            Median & Geo.\ mean &
            Median & Geo.\ mean &
            Median & Geo.\ mean \\
            \hline
            \hline

            \multirow{2}{*}{Czech}
            & robeczech-base
            & 2.63 & 5.68
            & 46.03 & 48.53
            & 3.53 & 7.54
            & 6.46 & 13.31 \\

            & CardioBERTa.cs
            & 1.27 & 2.68
            & 3.89 & 11.21
            & 1.39 & 3.18
            & 2.62 & 5.75 \\
            \hline

            \multirow{3}{*}{Dutch}
            & RobBERT2023-base
            & 5.379 & 10.471
            & 48.811 & 52.573
            & 7.924 & 13.004
            & 14.570 & 23.195 \\

            & MedRoBERTa.nl
            & 26.855 & 45.383
            & 57.573 & 70.015
            & 22.299 & 38.932
            & 21.042 & 36.950 \\

            & CardioBERTa.nl\_clinical
            & 7.962 & 16.798
            & 22.286 & 41.282
            & 5.851 & 12.115
            & 7.580 & 15.788 \\
            \hline

            \multirow{2}{*}{English}
            & biomed\_roberta\_base
            & 1.24 & 2.69
            & 1.37 & 2.58
            & 1.74 & 4.89
            & 1.92 & 3.84 \\

            & CardioBERTa.en
            & 1.04 & 1.55
            & 1.01 & 1.78
            & 1.08 & 2.09
            & 1.24 & 1.89 \\
            \hline

            \multirow{2}{*}{Italian}
            & bioBIT
            & 1.61 & 3.50
            & 162.66 & 108.33
            & 1.77 & 4.19
            & 3.50 & 6.20 \\

            & CardioBERTa.it
            & 1.70 & 4.21
            & 191.68 & 109.78
            & 2.09 & 4.82
            & 4.18 & 7.21 \\
            \hline

            \multirow{2}{*}{Romanian}
            & xlm-roberta-base
            & 2.78 & 6.78
            & 85.33 & 67.42
            & 4.48 & 8.32
            & 5.01 & 9.43 \\

            & CardioBERTa.ro
            & 1.16 & 2.67
            & 2.40 & 13.04
            & 1.32 & 2.70
            & 1.62 & 3.12 \\
            \hline

            \multirow{2}{*}{Spanish} 
            & PlanTL-GOB-ES/roberta-base-biomedical-clinical-es & 2.048 & 5.256 & 3.050 & 20.553 & 3.554 & 10.867 & 1.393 & 5.964 \\ 
            & DT4H/CardioBERTa.es & 2.295 & 4.195 & 4.009 & 8.000 & 3.219 & 6.272 & 1.953 & 4.505 \\ 
            \hline

            \multirow{2}{*}{Swedish}
            & bert-base-swedish-cased
            & 83.95 & 73.22
            & 598 & 406
            & 49.8 & 55.7
            & 104.6 & 107.7 \\

            & CardioBERTa.sv
            & 1.35 & 2.89
            & 2.1 & 14.7
            & 1.4 & 3.1
            & 2.7 & 6.3 \\
            \hline
        \end{tabular}%
    }
\end{table}
\end{landscape}

\clearpage
\section{Pretraining information}
\label{appendix_cpt}

\begin{table}[h]
\caption{Overview of collected corpora. Word counts are expressed in millions.}
\label{tab:corpus-overview}
\resizebox{\textwidth}{!}{
\begin{tabular}{@{}p{2.5cm}p{6cm}p{5.3cm}r@{}}
\toprule
Language & Corpus name & Corpus type & \multicolumn{1}{r}{Words (M)} \\
\midrule
\midrule
Czech & EMEA CS & Biomedical & 12.21 \\
Czech & Wikipedia Life Science Cs & Biomedical & 112.01 \\
Czech & Wikipedia Cardiology Cs & Cardiology & 0.10 \\
Czech & ABSTRACTS\_CS & General & 0.21 \\
Czech & CorEtVasa & Cardiology & 10.09 \\
Dutch & BioLORD definitions & Biomedical & 300.00 \\
Dutch & EPFL guidelines & General & 100.00 \\
Dutch & PMC Patients & Biomedical & 50.00 \\
Dutch & UMCU cardio & Cardiology & 800.00 \\
Dutch & NtvG articles (Dutch clinical magazine) & Biomedical & 33.00 \\
Dutch & Pubmed abstracts & Biomedical & 2,000.00 \\
Dutch & AGCT & Biomedical & 25.00 \\
Dutch & Wikipedia Cardiology Nl & Cardiology & 0.19 \\
Dutch & EMEA Dutch & Biomedical & 15.00 \\
Dutch & Dutch medical guidelines & Biomedical & 100.00 \\
Dutch & MIMIC IV discharge & Biomedical & 400.00 \\
Dutch & MIMIC IV radio reports & Biomedical & 500.00 \\
Dutch & Mimic III Dutch translations & Biomedical & 300.00 \\
English, Czech & UFAL Medical Corpus CS & General, Biomedical & 610.85 \\
English & MeSpEn\_Parallel\_Corpora & Biomedical & 6.20 \\
English & MIMIC IV Note & Biomedical & 922.66 \\
English & n2c2 2014: De-identification and Heart Disease Risk Factors Challenge & Cardiology & 0.81 \\
English & PubMed & Biomedical & 134.13 \\
English & Wikipedia Cardiology En & Cardiology & 2.04 \\
English, Czech & CzEng 1.6 & General & 981.47 \\
English, Spanish & UFAL Medical Corpus ES-EN & General, Biomedical & 610.05 \\
English, Swedish & Ufal Medical Corpus SV & General, Biomedical & 208.24 \\
English, Romanian & UFAL\_RO\_EN & General, Biomedical & 935.40 \\
English, Czech & CzEng16pre & General + 500 Medical sentences & 980.98 \\
Italian & Multilingual-Medical-Corpus IT & Biomedical & 149.04 \\
Italian & BioBERT Italian & Biomedical & 4,127.06 \\
Italian & PsyNIT & Psychiatry & 0.05 \\
Italian & EMEA IT & Biomedical & 14.50 \\
Italian & Wikipedia Cardiology It & Cardiology & 0.70 \\
Romanian & EMEA RO & Biomedical & 13.12 \\
Romanian & SiMoNERo & Cardiology, Diabetes, and Endocrinology & 1.62 \\
Romanian & BioRo & Biomedical & 0.12 \\
Romanian & BioRo\_Cardiology & Cardiology & 0.01 \\
Romanian & MoNERo & Biomedical & 0.37 \\
Romanian & MoNERo\_cardiology & Cardiology & 0.25 \\
Romanian & RONEC & General & 2.29 \\
Romanian & Romanian micro-blogging named entity recognition (MicroBloggingNERo) & General & 0.18 \\
Romanian & OSCAR & General & 6,302.60 \\
Romanian & Paracrawl & General & 416.44 \\
Romanian & CCMatrix & General & 1,590.34 \\
Spanish & CARES - A Corpus of Anonymised Radiological Evidences in Spanish & Clinical Cases & 0.64 \\
Spanish & ClinSpEn Corpus & Biomedical & 0.08 \\
Spanish & EMEA ES & Biomedical & 14.71 \\
Spanish & Wikipedia Life Science Es & Biomedical & 48.76 \\
Spanish & Wikipedia Cardiology Es & Cardiology & 0.88 \\
Swedish & Wikipedia Cardiology Sv & Cardiology & 0.23 \\
Swedish & OpenChart-SE & Biomedical & 0.01 \\
Swedish & SwedMedNER & Biomedical & 7.59 \\
\hline
\end{tabular}
}
\end{table}

\end{document}